\documentclass{article}

\PassOptionsToPackage{numbers, compress}{natbib}

 \usepackage[main, final]{neurips_2026}

\usepackage[utf8]{inputenc} 
\usepackage[T1]{fontenc}    
\usepackage{graphicx}
\usepackage{booktabs}
\usepackage{enumitem}
\usepackage[accsupp]{axessibility}  
\usepackage{multirow}
\usepackage[table]{xcolor}  
\usepackage{pifont}
\definecolor{mred}{RGB}{238, 34, 12}
\definecolor{mgreen}{RGB}{1, 127, 0}
\definecolor{mblue}{RGB}{0, 77, 158}
\newcommand{\mredbf}[1]{\textcolor{mred}{\textbf{#1}}}
\newcommand{\mbluebf}[1]{\textcolor{mblue}{\textbf{#1}}}
\usepackage{amsmath}
\usepackage{arydshln} 
\usepackage{makecell}
\usepackage{titlesec}
\titlespacing*{\section}{0pt}{2pt plus 0.5pt minus 0.5pt}{2pt plus 0.5pt minus 0.5pt} 
\titlespacing*{\subsection}{0pt}{2pt plus 0.5pt minus 0.5pt}{2pt plus 0.5pt minus 0.5pt} 
\titlespacing*{\paragraph}
{0pt}    
{0.5em}    
{0.5em}    
\usepackage{hyperref}
\newcommand{\MYhref}[3][blue]{\href{#2}{\color{#1}{#3}}} 
\usepackage{orcidlink}
\usepackage{makecell}

\title{ManipShield: A Unified Framework for Image Manipulation Detection, Localization and Explanation}

\author{%
  Zitong Xu \\
  Shanghai Jiao Tong University\\
  \texttt{xuzitong@sjtu.edu.cn} \\
  \And
  Huiyu Duan \thanks{Corresponding Author} \\
  Shanghai Jiao Tong University\\
  \texttt{huiyuduan@sjtu.edu.cn} \\
  \AND
  Xiaoyu Wang \\
  University of Electronic and Science Technology of China\\
  \texttt{2024300903005@std.uestc.edu.cn} \\
  \And
    Zhaolin Cai \\
  Shanghai Jiao Tong University\\
  \texttt{sebqaq21@gmail.com} \\
  \AND
  Kaiwei Zhang \\
  Shanghai Jiao Tong University\\
  \texttt{zhangkaiwei@sjtu.edu.cn} \\
  \And
  Xiongkuo Min \\
  Shanghai Jiao Tong University\\
  \texttt{minxiongkuo@sjtu.edu.cn} \\
  \And
  Guangtao Zhai \\
  Shanghai Jiao Tong University\\
  \texttt{zhaiguangtao@sjtu.edu.cn} \\
}

\begin{document}
\maketitle
\begin{abstract}
With the rapid advancement of generative models, powerful image editing methods now enable diverse and highly realistic image manipulations that far surpass traditional deepfake techniques, posing new challenges for manipulation detection. Existing image manipulation detection and localization (IMDL) benchmarks suffer from limited content diversity, narrow generative-model coverage, and insufficient interpretability, which hinders the generalization and explanation capabilities of current manipulation detection methods. To address these limitations, we introduce \textbf{ManipBench}, a large-scale benchmark for image manipulation detection and localization focusing on AI-edited images. ManipBench contains over 450K manipulated images produced by 25 state-of-the-art image editing models across 12 manipulation categories, among which 100K images are further annotated with bounding boxes, judgment cues, and textual explanations to support interpretable detection. Building upon ManipBench, we propose \textbf{ManipShield}, an all-in-one model based on a Multimodal Large Language Model (MLLM) that leverages contrastive LoRA fine-tuning and task-specific decoders to achieve unified image manipulation detection, localization, and explanation. Extensive experiments on ManipBench and several public datasets demonstrate that ManipShield achieves state-of-the-art performance and exhibits strong generality to unseen manipulation models. The project is available at \MYhref[magenta]{this https URL}{https://github.com/IntMeGroup/ManipShield}.
\end{abstract}

\section{Introduction}
\label{sec:intro}

Image manipulation detection and localization (IMDL) have long attracted significant attention due to their crucial role in preserving image integrity and ensuring security \cite{mvss,pscc,hifi,fakeshield}. With the rapid advancement of generative models \cite{FLUX,DiT}, numerous powerful artificial intelligence editing techniques have emerged, such as NanoBanana \cite{nanolink} and Qwen-Image-Edit \cite{qwenedit}. Unlike early deepfake techniques, which are typically limited to straightforward, malicious facial or identity manipulations \cite{mvss,hifi,pscc}, recent new editing methods enable a broader range of modifications, including background changes, gesture alterations, text manipulations, \textit{etc.}, potentially posing new security and privacy risks \cite{FragFake}. Furthermore, these editing methods are increasingly imperceptible, maintaining strong subject consistency while producing highly realistic images \cite{omnigen2,Flowedit,qwenedit}, which underscores the urgent need for more advanced image manipulation detection and localization approaches.
\begin{figure}
    \centering
    \includegraphics[width=1\linewidth]{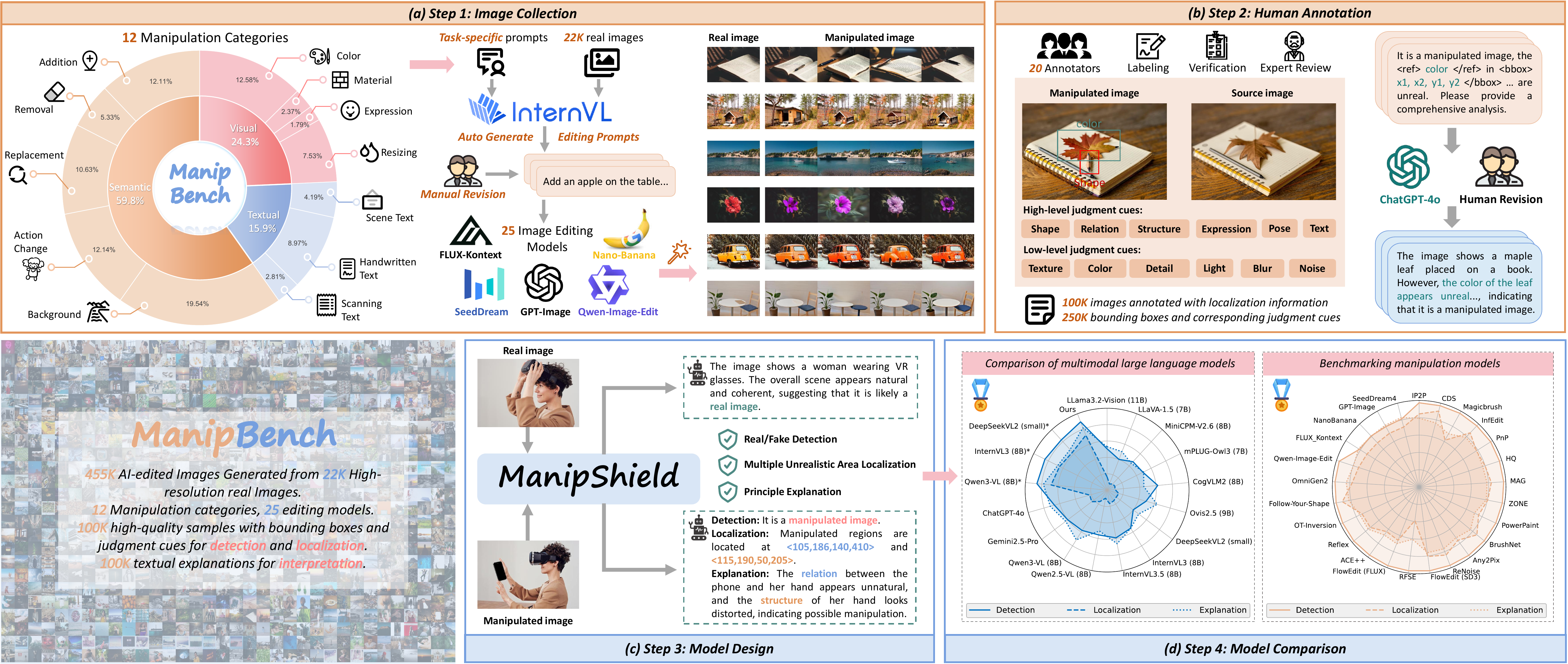}
    \caption{ An overview of the constructed image manipulation database and the proposed image manipulation detection model, termed ManipBench and ManipShield, respectively. (a) We collect 22K real images and generate editing prompts via a multimodal large language models across 12 manipulation categories, producing 455K manipulated images using 25 editing models. 
(b) 100K images are further annotated with localization and explanations. 
(c) We propose ManipShield for detection, localization, and explanation. 
(d) We perform model comparisons based on ManipBench.}
\label{teaser}
\end{figure}


As shown in Table~\ref{comparison}, existing IMDL datasets and benchmarks exhibit several critical limitations. \textbf{(i) Limited content diversity:} Existing datasets primarily focus on facial or identity forgeries \cite{casia,IMD2020,DDFT,DF40}, whereas modern editing techniques allow personalized and localized modifications guided by flexible instructions \cite{nanobanana,qwenedit,ACE}. The absence of such diverse editing categories limits the generalization of IMDL models on real-world manipulations. \textbf{(ii) Limited generative models:} most datasets rely on a small number of generative methods \cite{fakeshield,sid,gim,FragFake}, causing detection models to learn model-specific artifacts rather than generalizable features. Moreover, many of these methods are outdated, producing obvious artifacts, making detection easy but hindering the generalization of detection models to newer, more complex fakes. \textbf{(iii) Limited interpretability:} most benchmarks provide only binary real-or-fake labels \cite{IMD2020,DDFT,FakeBench,DF40}. While some improved benchmarks include annotations of manipulated regions \cite{casia,gim}, they still lack textual explanations to justify detection decisions, which limits the interpretability of model reasoning and constrains the development of more transparent manipulation detection methods. \textbf{(iv) Region-first pipeline:} Existing benchmarks predefine the manipulation region and conduct editing within a fixed mask \cite{fakeshield,sid,gim}, thereby restricting evaluation to mask-guided diffusion models and limiting their applicability to direct, instruction-driven editing scenarios where the manipulation region is not specified in advance. 

To address the limitations of existing datasets, we introduce ManipBench, as shown in Figure~\ref{teaser}, featuring diverse AI-edited images and fine-grained annotations, which has not been explored in prior benchmarks. \textbf{(i) Abundant content diversity:} ManipBench contains over 22K real source images covering diverse visual content. \textbf{(ii) Various manipulation tasks and numerous generative methods:} 21 state-of-the-art open-source editing models building on diverse diffusion backbones and 4 advanced closed-source editing methods are selected to produce manipulated images across 12 editing categories, resulting in over 450K manipulated images with diverse generative characteristics. \textbf{(iii) Fine-grained annotation:} ManipBench provides over 100K images meticulously annotated by trained annotators, using bounding boxes and corresponding judgment cues, along with textual explanations supporting the detection decisions. \textbf{(iv) Manipulation-first pipeline:} Unlike the region-first pipeline, we first apply the manipulations and then directly annotate the manipulated regions on the resulting images, benchmarking localization in more realistic, instruction-driven editing scenarios. As a multimodal dataset, ManipBench further enables comprehensive evaluation of MLLMs on detection, localization, and explanation. 

Based on ManipBench, we propose an all-in-one IMDL model, ManipShield, which is capable of (1) discriminating manipulated images, (2) localizing manipulated regions, (3) identifying unrealistic attributes within each region as evidential support, and (4) providing comprehensive analysis. Building upon a Multimodal Large Language Model (MLLM), ManipShield leverages contrastive low-rank adaptation (LoRA) fine-tuning to train the vision encoder, utilizes layer discrimination selection to identify the most informative large language model layer, and decodes the hidden state of this layer using task-specific decoders to generate the target outputs. Extensive experiments on our ManipBench and other IMDL datasets demonstrate that ManipShield achieves state-of-the-art performance. Moreover, we show that it achieves strong zero-shot performance on unseen generative models, highlighting its superior generalization ability. In summary, our main contributions are:
\begin{itemize}[left=12pt, labelsep=0.6em, labelwidth=0pt]
    \item We introduce ManipBench, a large-scale IMDL dataset focusing on AI-edited images, comprising over 450K manipulated images produced by 25 advanced editing models, with fine-grained annotations including bounding boxes, judgment cues, and textual explanations.
    \item We benchmark the image manipulation detection, localization, and explanation capabilities of MLLMs based on ManipBench.
    \item We propose ManipShield, an MLLM-based model that unifies manipulation detection, localization and explanation, offering a transparent manipulation detection method.
    \item Extensive experiments demonstrate that our method outperforms most existing approaches on our ManipBench and other image manipulation datasets. We also validate its strong zero-shot performance on unseen editing models.
\end{itemize}
\section{Related work}
\label{related work}
\subsection{AI editing}
With the advancement of generative models such as Stable Diffusion \cite{SD} and FLUX \cite{FLUX}, numerous editing methods have emerged \cite{ESurvey}. According to the type of editing prompts, these methods can be broadly categorized into description-based approaches (\textit{e.g.}, “dog” $\rightarrow$ “cat”) and instruction-based approaches (\textit{e.g.}, “change the dog to a cat”) \cite{ESurvey,lmm4edit}. Description-based methods primarily rely on inversion-based modification \cite{Flowedit, renoise, CDS, PnP, InfEdit}, which first performs image-to-noise inversion to obtain a latent representation, then conducts generation according to the new prompt. Instruction-based methods, such as InstructPix2Pix \cite{ip2p} and MagicBrush \cite{Magicbrush} generally utilize large-scale paired editing datasets to train editing models. Recent models \cite{qwenedit, omnigen2} further adopt the DiT \cite{DiT} architecture, enabling diverse and flexible editing instructions for generating highly realistic edited images.

\subsection{Image manipulation detection datasets}
As shown in Table~\ref{comparison}, early datasets~\cite{casia,mfc,coverage,IMD2020} mainly focus on manual manipulations such as splicing, removal, and copy-move. With the rise of generative models, datasets~\cite{DDFT,robustness,forgerynet,diffusionface,DF40} use GANs and diffusion models to produce deepfakes, but most are limited to facial content. Some datasets~\cite{FakeBench,forensics,wang2025dfbench} extend manipulations to general images, yet still lacking text-guided edited images that incorporate novel manipulations and distinctive visual features. Some recent image editing datasets~\cite{fakeshield,sid,gim,FragFake} have incorporated text-guided edited images, but typically cover only a limited range of models and manipulation types. Moreover, since they rely on region-first pipeline, most datasets only include mask-guided editing methods~\cite{fakeshield,sid,gim}, restricting to simple edits. Recent advances in MLLMs \cite{internvl3_5, qwen3, deepseekv2} have driven some benchmarks \cite{FakeBench,fakeshield,sid} to include textual explanations, offering interpretable insights into the manipulations, but the limited manipulation categories and models may restrict the generalization ability.

\subsection{Image manipulation detection and localization}
Early studies on IMDL mainly focus on localizing specific types of image manipulations, often limited to facial content \cite{mfcn,copymove,inpainting1,inpainting2,doagan}. However, since manipulation types vary widely in real-world scenarios, subsequent research has shifted toward universal manipulation localization methods \cite{mvss,pscc,hifi,diffforensics}, which aim to identify artifacts and inconsistencies across diverse manipulation types. There are also various advanced network architectures that can be adapted for localization tasks \cite{imlvit,su2025can,zhu2025mesoscopic,fastrcnn,detr,yang2026unipet,wen2024rethinking}.
With the rapid emergence of advanced editing models, recent methods \cite{sid,FragFake,fakeshield,gim} have focused on detecting AI-edited images. However, most methods are trained on only a few editing models, and because different models produce distinct editing artifacts, their generalization is limited. Some works \cite{sid,FakeBench,fakeshield} further leverage LLMs to explain the rationale behind their detection and localization decisions, providing human-interpretable reasoning. Nevertheless, these approaches are typically designed for simple editing scenarios involving single manipulated regions \cite{sid, fakeshield}. 
\begin{table*}[t]
\centering
\caption{An overview of image manipulation detection datasets.}
 \resizebox{1\textwidth}{!}{
\begin{tabular}{lcccccccc}
\toprule
\noalign{\vspace{-1.5pt}}
\multirow{2}{*}{Databases}&Image& Manipulation  & Manipulation & Text-guided  & Manipulated & Text-guided& Localization& Textual \\
&Content&  Categories & Methods & Editing Models & Images&Edited Images & Annotations &Explanations\\
\noalign{\vspace{-1pt}}
\midrule
\noalign{\vspace{-1pt}}
   IMD2020~\cite{IMD2020} & General &  3&3& \textcolor{red}{\ding{55}} & 35,000& \textcolor{red}{\ding{55}}  & \textcolor{red}{\ding{55}}& \textcolor{red}{\ding{55}} \\
   DDFT~\cite{DDFT} & Face &  4&7& \textcolor{red}{\ding{55}} & 240,336& \textcolor{red}{\ding{55}}  & \textcolor{red}{\ding{55}}& \textcolor{red}{\ding{55}} \\
   DF40~\cite{DF40} &Face&4&40&\textcolor{red}{\ding{55}}&1M+&\textcolor{red}{\ding{55}}&\textcolor{red}{\ding{55}}&\textcolor{red}{\ding{55}}\\

      FakeBench~\cite{FakeBench} &General&6&10&\textcolor{red}{\ding{55}}&3,000&\textcolor{red}{\ding{55}}&\textcolor{red}{\ding{55}}&3,000\\


         FakeShield~\cite{fakeshield} & General &  7&3& 
1& 196,123& 181,000 & 196,123& 34,000 \\
            SID-Set~\cite{sid} &General&2&1&1&200,000&200,000&200,000&3,000\\

                        GIM~\cite{gim} &General&2&3&3&320,000&320,000&320,000&\textcolor{red}{\ding{55}}\\

                        FragFake~\cite{FragFake} &General&2&4&4&20,000&20,000&\textcolor{red}{\ding{55}}&\textcolor{red}{\ding{55}}\\

\hline
\noalign{\vspace{0.5pt}}
\rowcolor{gray!20}  
 \textbf{ManipBench (Ours)} & \textbf{General} &  \textbf{12}&\textbf{25}& \textbf{
25}& \textbf{455,000}& \textbf{455,000} & \textbf{100,000}& \textbf{100,000} \\
 \noalign{\vspace{-2pt}}
\bottomrule
\label{comparison}
\end{tabular}}
\end{table*}

\section{ManipBench}
In this section, we introduce ManipBench, a large-scale IMDL dataset focusing on AI-edited images. The images are generated using multiple state-of-the-art editing methods and are annotated with bounding boxes, judgment cues, and detailed textual explanations.

\subsection{Image collection} 
We first collect 22K high-resolution real-world images from publicly available photography websites along with their corresponding textual descriptions. we define 12 manipulation categories, including add, remove, replace, color, background, size, action, material, expression, scene text, handwritten text, and scanning text. 

For each category, we employ the advanced MLLM, InternVL3.5 \cite{internvl3_5}, to generate suitable editing prompts based on the image content and the manipulation task, then perform manual examination and revision. The prompts contain both an instruction type and a description type to accommodate different editing models. 
We employ 21 state-of-the-art editing models to perform manipulation. 
The details of these editing models are provided in Section~\ref{editing_model} of the supplementary material. 
The whole generation process is shown in Figure~\ref{teaser}(a). Finally, a total of over 455K AI-edited images are produced, forming a rich set suitable for benchmarking image manipulation detection. 


\subsection{Metadata acquisition} 
\label{annotation}
To enable fine-grained localization analysis, we select 100K images for detailed annotation. Unlike the region-first pipeline, which can automatically obtain the manipulation region, the manipulation-first pipeline requires human annotation. Defining the manipulation region is therefore critical. Image editing often alters areas beyond the target region to maintain global consistency, resulting in widespread pixel-level changes. For this reason, we define the manipulated regions as those exhibiting perceptual semantic changes relative to the original image. Here, semantic changes also include alterations in texture, color, or other visual attributes that differ from the original identity, even if the object category remains the same. We emphasize perceptual changes because modifications that do not produce perceptible semantic differences are unlikely to be significant and generally do not pose safety or functional risks.

Since the annotated regions are defined at the semantic level rather than pixel-level, precise masks are unnecessary. Instead, we use bounding boxes, which provide a flexible and practical way to capture complex or ambiguous modifications. During annotation, annotators are presented with paired real and manipulated images, draw bounding boxes around each modified region, and record cues that indicate the region has been manipulated. These cues are categorized into high-level features (shape, structure, relation, text, pose, expression, \textit{etc.}) and low-level features (texture, blur, noise, light, detail, color, \textit{etc.}).

Annotations are performed using a custom Python-based interface. Before starting, annotators have to receive training and finish a test to ensure consistency. Annotators work under controlled conditions, with images shown in randomized order and sessions divided into short rounds to minimize fatigue. Reliability is ensured through a three-stage protocol of independent labeling, cross-verification, and expert voting. For explanatory analysis, we first use ChatGPT-4o \cite{chatgpt4o} to generate textual descriptions based on bounding boxes and cues, which are then manually reviewed and revised. The complete metadata annotation process is illustrated in Figure~\ref{teaser}(b). 
The explicit annotation process is described in Section~\ref{annotation_more} of supplementary material. 
Finally, we obtain 100K images with bounding-box annotations, judgment cues, and textual explanations.
\begin{figure*}[t]
   \centering
  \includegraphics[width=1\textwidth]{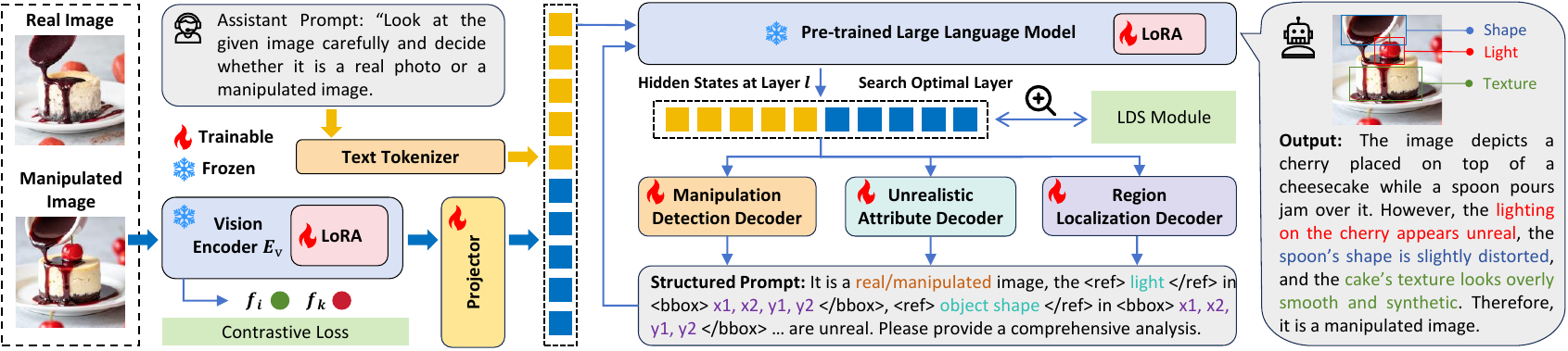}
  \caption{Overview of ManipShield. We first perform contrastive LoRA fine-tuning on paired real and manipulated images to train the vision encoder. The encoder is then frozen, and the projected image features, along with an assistant prompt, are fed into the LLM. The layer discrimination selection (LDS) module identifies the most discriminative layer, whose hidden state is used by three decoders for manipulation detection, unrealistic attribute extraction, and region localization. The outputs are combined into a structured prompt to generate explicit explanations.}
  \label{model}
\end{figure*}
\begin{figure}[t]
  \includegraphics[width=1\textwidth]{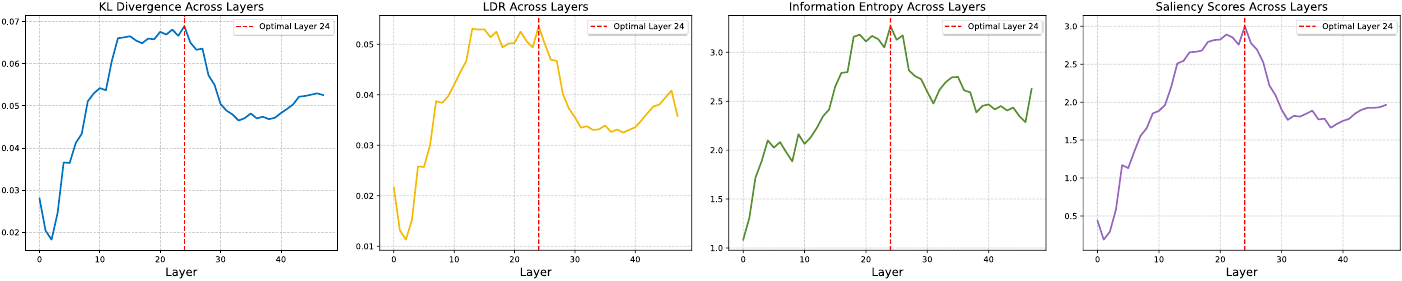}
  \caption{The plots of KL divergence, LDR, information entropy and saliency score across hidden states from different LLM layers.}
  \label{LDS}
\end{figure}
\section{ManipShield}
In this section, we present ManipShield, a unified model for image manipulation detection, localization and explanation.
\subsection{Overall architecture}
As shown in Figure~\ref{model}, ManipShield is built upon an MLLM architecture, \textit{i.e.}, InternVL3.5 \cite{internvl3_5}. First, we apply \textbf{contrastive LoRA fine-tuning} to the vision encoder of a MLLM. Then, we perform \textbf{layer discrimination selection} using a part of samples to identify the optimal layer in the LLM that best separates positive and negative samples based on statistical distributions. The hidden state from this layer is fed into three specialized decoders: a \textbf{manipulation detection decoder} for classification, an \textbf{unrealistic attribute decoder} for cues extraction, and a \textbf{region localization decoder} for bounding boxes prediction. The outputs of these three decoders are then combined into a \textbf{structured prompt}, which ManipShield uses, together with the image, to generate explicit explanatory analysis.

\subsection{Contrastive LoRA fine-tuning}
Table~\ref{acc1} shows that the zero-shot performance of MLLMs on image manipulation detection is limited. Many studies have demonstrated that fine-tuning can significantly improve the visual capabilities of MLLMs \cite{duan2025finevq,xu2025harmonyiqa,yang2026balancing,qi2026cacr,zheng2025fingpt}, motivating us to fine-tune the pre-trained model. We adopt InternVL3.5 \cite{internvl3_5} as our backbone and apply the LoRA technique \cite{lora}. LoRA models the weight update of each layer $W \in \mathbb{R}^{d \times k}$ as a low-rank decomposition $\Delta W = AB$, where $A \in \mathbb{R}^{d \times r}$ and $B \in \mathbb{R}^{r \times k}$ with $r \ll \min(d, k)$. The forward pass with LoRA is then given by
\begin{equation}
\small
    h = Wx + \Delta Wx = (W + AB)x.
\end{equation}

To enhance the model's ability to discriminate between real and manipulated images, we construct positive and negative image pairs and employ contrastive learning. 
Let the feature embeddings of a positive pair $(i, j)$ be denoted as $f_i$ and $f_j$. The feature embeddings of all other images in the batch are denoted as $f_k$ ($k \neq i,j$) and serve as negative samples. The contrastive loss is formulated as:
\begin{equation}
\small
\mathcal{L}_{\mathrm{contrastive}} = - \frac{1}{N} \sum_{i=1}^{N} 
\log \frac{\exp(\mathrm{sim}(f_i, f_j)/\tau)}
{\sum_{k \neq i, j} \exp(\mathrm{sim}(f_i, f_k)/\tau)},
\end{equation}
where $\mathrm{sim}$ denotes the cosine similarity, $\tau$ is a temperature hyperparameter, and $N$ is the batch size. 
All feature embeddings are $\ell_2$-normalized before computing the similarity. To enable parameter-efficient adaptation with minimal modifications to the MLLM, we only fine-tuning the vision encoder, due to its pivotal role in visual feature extraction.

\subsection{Layer discrimination selection}
Recent studies have shown that the middle layers of MLLMs can reflect reasoning processes and yield better performance on visual tasks \cite{hidden1,hidden2}. Motivated by this, we systematically investigate the hidden state representations from different layers of MLLMs to identify the most informative layer for image manipulation detection. Specifically, we randomly select $1\%$ of images from ManipBench and perform a single forward pass through the pre-trained MLLM, InternVL3.5 \cite{internvl3_5}, whose vision encoder has been fine-tuned using our contrastive LoRA method. From each layer $l$, we extract the hidden state $h_l$ of the first generation token. 
To quantify the information captured across layers statistically, we analyze properties of extracted hidden states $h_l$ with following metrics:

\paragraph{\textbf{Manipulation sensitivity via KL divergence:}} The Kullback–Leibler (KL) divergence quantifies the statistical distinguishability between features of real and manipulated images. 
For each feature dimension $d$ at layer $l$, we assume that the hidden states of positive samples ($\mathbf{h}_l^N$) and negative samples ($\mathbf{h}_l^A$) follow Gaussian distributions $\mathcal{N}(\mu_{l,d}^N, (\sigma_{l,d}^N)^2)$ and $\mathcal{N}(\mu_{l,d}^A, (\sigma_{l,d}^A)^2)$, respectively. 
The KL divergence is computed between these two Gaussian distributions for each feature dimension, and the overall manipulation sensitivity for layer $l$ is obtained by averaging across all feature dimensions $D$:
\begin{equation}
\small
    D_{\text{KL}}(l) = \frac{1}{D} \sum_{d=1}^{D} D_{\text{KL}}^{(d)}(l).
\end{equation}
A higher $D_{\text{KL}}(l)$ indicates a greater distributional difference between features of real and manipulated images at layer $l$.
\paragraph{\textbf{Class separability via local discriminant ratio:}} The Local Discriminant Ratio (LDR) measures the ability of features to linearly separate different classes. For each feature dimension $d$ at layer $l$, we calculate a LDR as the ratio of the squared difference between the means of normal ($\mu_{l,d}^N$) and anomalous ($\mu_{l,d}^A$) features to the sum of their variances $((\sigma_{l,d}^N)^2$ and $(\sigma_{l,d}^A)^2)$, with a small constant $\epsilon$ added for numerical stability:
    \begin{equation}
    \small
        \text{LDR}^{(d)}(l) = \frac{(\mu_{l,d}^N - \mu_{l,d}^A)^2}{(\sigma_{l,d}^N)^2 + (\sigma_{l,d}^A)^2 + \epsilon}.
    \end{equation}
    The overall class separability of layer~$l$ is the mean LDR across all~$D$ feature dimensions:
    \begin{equation}
    \small
        \text{LDR}(l) = \frac{1}{D} \sum_{d=1}^{D} \text{LDR}^{(d)}(l).
    \end{equation}
    A higher $\text{LDR}(l)$ suggests stronger linear separability between the real and manipulation classes, implying more discriminative features at layer~$l$.
\paragraph{\textbf{Information concentration via feature entropy:}} To assess the information concentration within the feature representations, for each feature dimension $d$ at layer $l$, we estimate the probability distribution by partitioning the feature values into a fixed number of $B$ bins with evenly spaced boundaries determined by the range of feature values across all samples. The entropy for the $d$-th dimension is then calculated as:
    \begin{equation}
    \small
        H^{(d)}(l) = -\sum_{j=1}^{B} p(\mathbf{h}_l[d] \in \text{bin}_j) \log_2 p(\mathbf{h}_l[d] \in \text{bin}_j),
    \end{equation}
    where $p(\mathbf{h}_l[d] \in \text{bin}_j)$ is the probability of the feature value falling into the $j$-th bin. The overall entropy for layer $l$ is the average entropy across all $D$ feature dimensions:
    \begin{equation}
    \small
        H(l) = \frac{1}{D} \sum_{d=1}^{D} H^{(d)}(l).
    \end{equation}
    Higher values indicate more uniform distribution of feature values across bins and capturing more diverse information.

To effectively combine these metrics, we apply Z-score normalization across all layers for KL divergence, LDR, and Entropy. For a metric $M \in \{D_{KL}(l), LDR(l), H(l)\} $, the normalized score $\hat M(l)=\frac{M(l)-\mu_M}{\sigma_M}$. The saliency score $S(l)$ for each layer is then calculated by:
\begin{equation}
\small
    S(l) = \hat D_{KL}(l)+\hat{LDR}(l)+\hat H(l).
\end{equation}
The optimal layer is determined as the one that maximizes the saliency score. As shown in Figure~\ref{LDS}, the saliency score peaks at layer 24, where the KL divergence, LDR, and information entropy also attain their peak values, indicating that layer 24 provides the most distinctive features between real and manipulated images. 
A more detailed discussion of these metrics is provided in Section~\ref{LDS} of supplementary material.
 \begin{table*}[t]
\belowrulesep=0pt
\aboverulesep=0pt
\centering
\setlength{\tabcolsep}{3pt} 
\renewcommand\arraystretch{1.2}  
\caption{Comparison results on different manipulation methods. $\spadesuit$ standard CNN/Transformer baselines, $\heartsuit$ image manipulation detection methods, $\clubsuit$ AI-generated image detection methods, $\diamondsuit$ multimodal large language models. The fine-tuned results are marked with \raisebox{0.25ex}{\tiny \ding{91}}. The best results are highlighted in \mredbf{red}, and the second-best results are highlighted in \mbluebf{blue}.}
\resizebox{1\textwidth}{!}{\begin{tabular}{l||cccccccccccc||cccccccc||cc}
\toprule
\noalign{\vspace{1pt}}
&\multicolumn{12}{c}{In-Distribution}&\multicolumn{8}{c}{Out-of-Distribution}\\
\cmidrule(lr){2-13}
\cmidrule(lr){14-21}
\noalign{\vspace{1pt}}
Testing Subset& \multicolumn{2}{c}{SD1.4} & \multicolumn{2}{c}{SD1.5} & \multicolumn{2}{c}{SD3} & \multicolumn{2}{c}{SDXL} & \multicolumn{2}{c}{FLUX} & \multicolumn{2}{c}{DiT} &\multicolumn{2}{c}{FLUX-Kontext} & \multicolumn{2}{c}{NanoBanana}  & \multicolumn{2}{c}{GPT-Image}& \multicolumn{2}{c}{SeedDream4}&\multicolumn{2}{c}{Overall} \\
\cmidrule(lr){2-3}
\cmidrule(lr){4-5}
\cmidrule(lr){6-7}
\cmidrule(lr){8-9}
\cmidrule(lr){10-11}
\cmidrule(lr){12-13}
\cmidrule(lr){14-15}
\cmidrule(lr){16-17}
\cmidrule(lr){18-19}
\cmidrule(lr){20-21}
\cmidrule(lr){22-23}
\noalign{\vspace{1pt}}
 Model/Metric&Acc$\uparrow$& F1$\uparrow$ &Acc$\uparrow$& F1$\uparrow$ &Acc$\uparrow$& F1$\uparrow$ &Acc$\uparrow$& F1$\uparrow$ &Acc$\uparrow$& F1$\uparrow$ &Acc$\uparrow$& F1$\uparrow$ &Acc$\uparrow$& F1$\uparrow$ &Acc$\uparrow$& F1$\uparrow$ &Acc$\uparrow$& F1$\uparrow$ &Acc$\uparrow$& F1$\uparrow$ &Acc$\uparrow$& F1$\uparrow$ \\
\midrule
\noalign{\vspace{0.5pt}}
Random Choice& 50.00 & 50.00 & 50.00 &50.00 & 50.00 & 50.00 & 50.00 & 50.00 & 50.00 & 50.00 & 50.00 & 50.00 & 50.00 & 50.00 & 50.00 & 50.00 & 50.00 & 50.00 & 50.00 & 50.00& 50.00 & 50.00\\
\hdashline
\noalign{\vspace{0.5pt}}
$\heartsuit$HifiNet \cite{hifi} & 63.75 & 64.72 & 60.35 & 62.54 & 61.29 & 63.09 & 57.26 & 60.74 & 58.02 & 61.21 & 57.39 & 60.85 & 53.49 & 58.71 & 56.72 & 60.44 & 55.65 & 59.85 & 53.76 & 10.27 & 59.45 & 60.15 \\
$\heartsuit$FakeShield \cite{fakeshield}& 73.11 & 74.43 & 67.81 & 70.66 & 67.34 & 70.33 & 60.48 & 66.21 & 61.34 & 66.82 & 60.21 & 66.11 & 54.57 & 63.02 & 54.84 & 63.16 & 54.84 & 63.16 & 53.40 & 11.73 & 64.45 & 66.81\\
\hdashline
\noalign{\vspace{0.5pt}}
$\clubsuit$Univ \cite{univ}& 72.63 & 73.78 & 66.87 & 69.61 & 64.92 & 68.38 & 58.06 & 64.38 & 59.41 & 65.25 & 57.39 & 64.05 & 51.08 & 60.78 & 55.65 & 63.09 & 52.96 & 61.71 & 52.02 & 11.22 & 62.96 & 65.55\\
$\clubsuit$AIDE \cite{AIDE}& 73.75 & 75.33 & 68.01 & 71.36 & 66.80 & 70.56 & 63.17 & 68.36 & 64.20 & 69.06 & 63.03 & 68.31 & 60.22 & 66.67 & 63.17 & 68.36 & 55.91 & 64.35 & 54.95 & 12.38 & 66.27 & 68.39\\

\hdashline
\noalign{\vspace{0.5pt}}
$\diamondsuit$DeepSeekVL2 (small) \cite{deepseekv2} & 53.05 & 66.47 & 51.81 & 65.87 & 51.88 & 65.91 & 52.96 & 66.41 & 52.06 & 65.99 & 50.27 & 65.16 & 49.73 & 64.92 & 48.12 & 64.19 & 48.66 & 64.43 & 53.36 & 17.63 & 50.20 & 63.52\\
$\diamondsuit$InternVL3.5 (8B) \cite{internvl3_5}& 69.89 & 71.32 & 64.78 & 67.67 & 66.80 & 68.95 & 57.53 & 63.43 & 60.08 & 64.96 & 67.61 & 69.62 & 62.10 & 66.02 & 64.25 & 67.32 & 54.84 & 61.99 & 55.96 & 11.80 & 64.10 & 65.44\\
$\diamondsuit$Qwen3-VL (8B) \cite{qwen3}& 69.53 & 70.68 & 64.78 & 67.18 & 67.20 & 68.73 & 56.18 & 62.18 & 58.83 & 63.79 & 68.28 & 69.68 & 59.95 & 64.27 & 61.83 & 65.37 & 53.76 & 60.91 & 56.06 & 11.20 & 63.52 & 64.65\\
$\diamondsuit$Gemini2.5-Pro \cite{nanobanana}& 68.10 & 65.24 & 64.31 & 62.39 & 65.72 & 63.31 & 58.33 & 58.67 & 57.57 & 58.33 & 61.97 & 60.90 & 55.38 & 56.99 & 62.37 & 61.11 & 57.26 & 58.05 & 65.59 & 12.09 & 62.62 & 59.45 \\
$\diamondsuit$ChatGPT-4o \cite{chatgpt4o}& 71.23 & 71.98 & 71.10 & 71.83 & 68.81 & 70.26 & 70.16 & 71.17 & 67.38 & 69.32 & 66.66 & 68.92 & 66.94 & 69.02 & 67.47 & 69.37 & 68.01 & 69.72 & 65.01 & 14.41& 68.99 & 68.29 \\
\hline
\noalign{\vspace{0.5pt}}
$\spadesuit$ResNet50\raisebox{0.25ex}{\tiny \ding{91}} \cite{resnet}& 88.80 & 80.11 & 88.98 & 80.15 & 88.84 & 80.12 & 88.71 & 80.10 & 88.62 & 80.08 & 86.56 & 79.70 & 65.32 & 74.77 & 58.06 & 72.48 & 56.99 & 72.11 & 68.01 & 75.52 & 84.33 & 79.05\\
$\spadesuit$Swin-T\raisebox{0.25ex}{\tiny \ding{91}} \cite{swin}& 90.59 & 82.80 & 90.52 & 82.79 & 90.59 & 82.80 & 89.78 & 82.67 & 90.10 & 82.72 & 89.52 & 82.63 & 62.63 & 76.90 & 66.13 & 77.85 & 62.37 & 76.82 & 66.94 & 78.06 & 86.17 & 81.90 \\
\hdashline
\noalign{\vspace{0.5pt}}
$\heartsuit$MVSSNet\raisebox{0.25ex}{\tiny \ding{91}} \cite{mvss}& 86.83 & 78.21 & 87.70 & 78.38 & 87.77 & 78.39 & 86.02 & 78.05 & 87.68 & 78.37 & 82.80 & 77.38 & 62.90 & 72.22 & 56.45 & 70.00 & 65.32 & 72.97 & 59.95 & 71.25& 82.78 & 77.16\\
$\heartsuit$PSCCNet\raisebox{0.25ex}{\tiny \ding{91}} \cite{pscc}& 90.99 & 83.68 & 91.06 & 83.69 & 90.99 & 83.68 & 91.13 & 83.70 & 91.08 & 83.70 & 89.11 & 83.39 & 66.40 & 78.91 & 59.41 & 77.00 & 57.26 & 76.34 & 63.71 & 78.22 & 86.19 & 82.70\\
$\heartsuit$HifiNet\raisebox{0.25ex}{\tiny \ding{91}} \cite{hifi} & 92.34 & 87.29 & 93.08 & 87.38 & 92.74 & 87.34 & 92.20 & 87.28 & 92.88 & 87.36 & 89.65 & 86.96 & 72.04 & 84.28 & 66.94 & 83.28 & 61.02 & 81.95 & 69.89 & 83.87 & 88.42 & 86.67 \\
$\heartsuit$FakeShield\raisebox{0.25ex}{\tiny \ding{91}} \cite{fakeshield}& 92.97 & 87.37 & 93.01 & 87.37 & 92.88 & 87.36 & 92.74 & 87.34 & 93.10 & 87.38 & 89.38 & 86.93 & 75.54 & 84.89 & 68.28 & 83.55 & 66.94 & 83.28 & 63.44 & 82.52 & 88.80 & 86.73\\
\hdashline
\noalign{\vspace{0.5pt}}
$\clubsuit$CNNSpot\raisebox{0.25ex}{\tiny \ding{91}} \cite{cnnspot}& 91.71 & 91.91 & 90.86 & 91.85 & 92.07 & 91.94 & 90.05 & 91.78 & 91.26 & 91.88 & 86.02 & 91.43 & 71.51 & 89.86 & 69.89 & 89.66 & 64.25 & 88.85 & 68.82 & 89.51 & 87.28 & 91.46\\
$\clubsuit$Lagrad\raisebox{0.25ex}{\tiny \ding{91}} \cite{lagrad}& 94.76 & 90.27 & 94.83 & 90.28 & 94.76 & 90.27 & 94.89 & 90.28 & 94.85 & 90.28 & 92.61 & 90.07 & 70.43 & 87.33 & 69.89 & 87.25 & 64.25 & 86.28 & 70.16 & 87.29 & 90.45 & 89.74\\
$\clubsuit$Univ\raisebox{0.25ex}{\tiny \ding{91}} \cite{univ}& 93.10 & 87.83 & 93.35 & 87.86 & 93.41 & 87.86 & 93.01 & 87.82 & 93.50 & 87.87 & 92.20 & 87.72 & 75.54 & 85.41 & 65.59 & 83.56 & 63.98 & 83.22 & 73.12 & 85.00 & 89.42 & 87.27 \\
$\clubsuit$AIDE\raisebox{0.25ex}{\tiny \ding{91}} \cite{AIDE}& \mbluebf{95.97} & \mbluebf{94.20} & \mbluebf{96.17} & 94.21 & \mbluebf{96.24} & 94.21 & 94.62 & 94.12 & \mbluebf{96.01} & 94.20 & \mbluebf{93.82} & \mbluebf{94.07}& \mbluebf{76.61} & \mbluebf{92.83} & 72.85 & 92.49 & 75.00 & \mbluebf{92.69} & 75.81 & 92.76 & 92.46 & \mbluebf{93.95} \\
\hdashline
\noalign{\vspace{0.5pt}}
$\diamondsuit$DeepSeekVL2 (small)\raisebox{0.25ex}{\tiny \ding{91}} \cite{deepseekv2} & 94.65 & 93.21 & 95.16 & 93.27 & 91.68 & 93.74 & 93.25 & 94.50 & 92.16 & 92.51 & 91.37 & 94.26 & 73.85 & 88.94 & 73.28 & 89.87 & 71.52 & 85.70 & 75.06 & 86.96 &88.54&89.30\\
$\diamondsuit$InternVL3 (8B)\raisebox{0.25ex}{\tiny \ding{91}} \cite{internvl3}& 93.26 & 93.50 & 95.17 & \mbluebf{94.33} & 92.39 & 94.45 & 95.47 & \mbluebf{94.62} & 92.17 & 93.53 & 94.15 & 94.38 & 74.84 & 89.06 & \mbluebf{77.17} & \mbluebf{92.62} & 72.10 & 85.58 & \mbluebf{76.49} & \mbluebf{93.62}&89.74&90.13 \\
$\diamondsuit$Qwen3-VL (8B)\raisebox{0.25ex}{\tiny \ding{91}} \cite{qwen3}& 95.14 & 93.26 & 94.17 & 94.29 & 95.27 & \mbluebf{94.75} & \mbluebf{96.42} & 92.38 & 94.53 & \mbluebf{95.17} & 93.28 & 93.51 & 76.24 & 91.18 & 76.68 & 91.88 & \mbluebf{79.37} & 90.13 & 76.48 & 92.25 &\mbluebf{92.58}&93.66 \\
\hdashline
\noalign{\vspace{0.5pt}}
\rowcolor{gray!20}  
ManipShield (Ours)\raisebox{0.25ex}{\tiny \ding{91}} & \mredbf{97.27} & \mredbf{95.26} & \mredbf{97.51} & \mredbf{95.27} & \mredbf{97.18} & \mredbf{95.26} & \mredbf{97.31} & \mredbf{95.26} & \mredbf{97.36} & \mredbf{95.27} & \mredbf{96.91} & \mredbf{95.24} & \mredbf{81.45} & \mredbf{94.39} & \mredbf{80.91} & \mredbf{94.36} & \mredbf{82.57} & \mredbf{94.27} & \mredbf{81.18} & \mredbf{94.38}& \mredbf{94.66} & \mredbf{95.12}\\
 \noalign{\vspace{-0.5pt}}
\bottomrule
\end{tabular}}
\label{acc1}
\end{table*}
 \subsection{Task-specific decoders}
After selecting the optimal layer, three specialized decoders are introduced to interpret the hidden state of the first generated token from that layer. The manipulation detection decoder focuses on distinguishing real and manipulated images. The unrealistic attribute decoder captures diverse cues that reflect unrealistic characteristic, while the region localization decoder predicts the corresponding spatial regions through bounding-box regression. All three decoders are based on a similar multi-layer perception (MLP) and are optimized with task-specific objectives. In this process, LoRA module is applied solely to the LLM. The outputs are aggregated into a structured prompt. This prompt is returned to ManipShield along with the image, allowing it to generate explicit analysis. With these task-specific decoders, ManipShield operates as a unified model for manipulation detection, localization, and explanation.
\begin{table*}[t]
\belowrulesep=0pt
\aboverulesep=0pt
\centering
\setlength{\tabcolsep}{8pt} 

\caption{Comparison results (loU$\uparrow$) of bounding box prediction. $\clubsuit$ object detection methods, $\diamondsuit$ multimodal large language models. The fine-tuned results are marked with \raisebox{0.25ex}{\tiny \ding{91}}. The best results are highlighted in \mredbf{red}, and the second-best results are highlighted in \mbluebf{blue}.}
\resizebox{1\textwidth}{!}{\begin{tabular}{l||cccccc||cccc||c}
\toprule
\noalign{\vspace{1pt}}
& \multicolumn{6}{c}{In-Distribution} & \multicolumn{4}{c}{Out-of-Distribution}\\
\cmidrule(lr){2-7}
\cmidrule(lr){8-11}
\noalign{\vspace{1.5pt}}
Model/Subset& SD1.4 & SD1.5 & SD3 & SDXL & FLUX & DiT & FLUX-Kontext & NanoBanana & GPT-Image &  SeedDream4&Overall \\
\midrule
\noalign{\vspace{0.5pt}}
$\diamondsuit$DeepSeekVL2 (small) \cite{deepseekv2} & 0.148 & 0.151 & 0.155 & 0.172 & 0.129 & 0.149 & 0.152 & 0.136 & 0.147 & 0.133 & 0.141 \\
$\diamondsuit$InternVL3.5 (8B) \cite{internvl3_5}& 0.201 & 0.197 & 0.184 & 0.209 & 0.137 & 0.182 & 0.188 & 0.141 & 0.154 & 0.160 & 0.168 \\
$\diamondsuit$Qwen3-VL (8B) \cite{qwen3}& 0.194 & 0.194 & 0.178 & 0.196 & 0.133 & 0.186 & 0.153 & 0.157 & 0.151 & 0.168 & 0.163 \\
$\diamondsuit$Gemini2.5-Pro \cite{nanobanana}& 0.257 & 0.287 & 0.264 & 0.272 & 0.198 & 0.252 & 0.234 & 0.228 & 0.213 & 0.231 & 0.234 \\
$\diamondsuit$ChatGPT-4o \cite{chatgpt4o}& 0.256 & 0.261 & 0.245 & 0.267 & 0.184 & 0.231 & 0.204 & 0.217 & 0.220 & 0.239 & 0.221 \\
\hline
\noalign{\vspace{0.5pt}}
MVSSNet \cite{mvss}& 0.327 & 0.306 & 0.312 & 0.337 & 0.295 & 0.308 & 0.312 & 0.286 & 0.293 & 0.315 & 0.311 \\
PSCCNet \cite{pscc}& 0.341&0.320&0.308&0.332&0.316&0.328&0.312&0.327&0.305&0.291&0.326  \\
HifiNet \cite{hifi}& 0.387&0.374&0.368&0.379&0.358&0.336&0.341&0.320&0.322&0.324 &0.361\\
FakeShield \cite{fakeshield}& 0.406&0.392&0.386&0.411&0.395&0.387&0.371&0.376&0.364&0.373&0.388  \\
\hline
\noalign{\vspace{0.5pt}}
$\clubsuit$Fast R-CNN\raisebox{0.25ex}{\tiny \ding{91}} \cite{fastrcnn} & 0.435 & 0.451 & 0.413 & 0.456 & 0.413 & 0.437 & 0.349 & 0.310 & 0.312 & 0.358 & 0.378 \\
$\clubsuit$DETR\raisebox{0.25ex}{\tiny \ding{91}} \cite{detr}& 0.661 & 0.696 & 0.633 & 0.645 & 0.574 & 0.638 & 0.521 & 0.491 & 0.478 & 0.510 & 0.573 \\
$\clubsuit$DeformableDETR\raisebox{0.25ex}{\tiny \ding{91}} \cite{deformabledetr}& 0.701 & 0.737 & 0.716 & 0.740 & \mbluebf{0.694} & 0.718 & 0.677 & 0.617 & 0.621 & 0.629 & 0.674 \\
$\clubsuit$GroundingDINO\raisebox{0.25ex}{\tiny \ding{91}} \cite{groundingdino}& \mbluebf{0.751} & \mbluebf{0.792} & \mbluebf{0.773} & \mbluebf{0.779} & 0.688 & 0.706 & 0.661 & 0.658 & 0.675 & \mbluebf{0.685} & 0.708 \\
\hdashline
\noalign{\vspace{0.5pt}}
$\diamondsuit$DeepSeekVL2 (small)\raisebox{0.25ex}{\tiny \ding{91}} \cite{deepseekv2} & 0.673 & 0.719 & 0.694 & 0.702 & 0.655 & 0.650 & 0.655 & \mbluebf{0.671} & 0.653 & 0.654 & 0.697 \\
$\diamondsuit$InternVL3 (8B)\raisebox{0.25ex}{\tiny \ding{91}} \cite{internvl3}& 0.703 & 0.726 & 0.708 & 0.730 & 0.675 & 0.696 & \mbluebf{0.683} & 0.642 & 0.657 & 0.670 & 0.704 \\
$\diamondsuit$Qwen3-VL (8B)\raisebox{0.25ex}{\tiny \ding{91}} \cite{qwen3}& 0.743 & 0.775 & 0.750 & 0.764 & 0.680 & \mbluebf{0.727}  & 0.682 & 0.651 & \mbluebf{0.676} & 0.684& \mbluebf{0.716} \\
\hdashline
\noalign{\vspace{0.5pt}}
\rowcolor{gray!20}  
ManipShield (Ours)\raisebox{0.25ex}{\tiny \ding{91}} & \mredbf{0.778} & \mredbf{0.798} & \mredbf{0.792} & \mredbf{0.787} & \mredbf{0.725} & \mredbf{0.753} & \mredbf{0.711} & \mredbf{0.701} & \mredbf{0.709} & \mredbf{0.716} & \mredbf{0.749} \\
 \noalign{\vspace{-0.5pt}}
\bottomrule
\end{tabular}}
\label{bbox}
\end{table*}
\begin{figure*}[t]
\centering
  \includegraphics[width=1\textwidth]{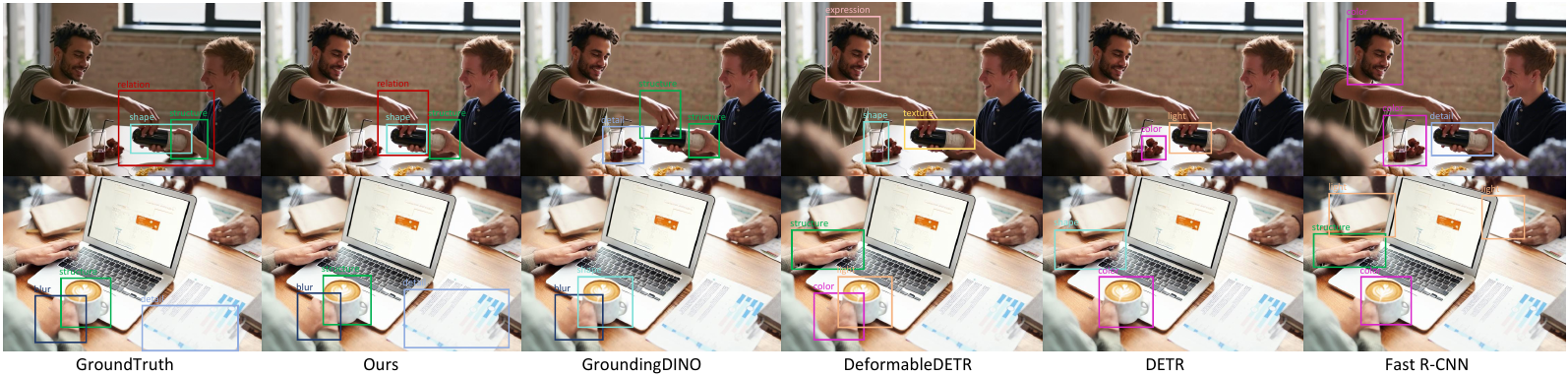}
  \caption{Examples of bounding box prediction results.}
  \label{example}
\end{figure*}
\begin{table*}[!tbp]
\belowrulesep=0pt
\aboverulesep=0pt
\centering
\setlength{\tabcolsep}{8pt} 
\caption{Comparison results (CSS$\uparrow$) of explanation capabilities. All embeddings are extracted using the InternVL3.5 \cite{internvl3_5}. The best results are highlighted in \mredbf{red}, and the second-best results are highlighted in \mbluebf{blue}.}
\resizebox{1\textwidth}{!}{\begin{tabular}{l||cccccccccc||c}
\toprule
\noalign{\vspace{1pt}}
& \multicolumn{10}{c}{Testing Subset}\\
\cmidrule(lr){2-11}
\noalign{\vspace{1.5pt}}
Model/Subset& SD1.4 & SD1.5 & SD3 & SDXL & FLUX & DiT & FLUX-Kontext & NanoBanana & GPT-Image &  SeedDream4&Overall \\
\midrule
\noalign{\vspace{0.5pt}}
LLaVA1.6 (7B) \cite{llava}&0.518&0.515&0.510&0.505&0.495&0.502&0.497&0.482&0.501&0.512&0.504\\
Ovis2.5 (9B) \cite{ovis25}&0.645&0.638&0.641&0.634&0.637&0.620&0.618&0.624&0.615&0.622&0.635\\
DeepSeekVL2 (small) \cite{deepseekv2}&0.512&0.508&0.506&0.471&0.477&0.482&0.482&0.478&0.455&0.464&0.484\\
Qwen3-VL (8B) \cite{qwen3}&0.718&0.720&0.706&\mbluebf{0.722}&0.713&\mbluebf{0.709}&0.697&0.702&\mbluebf{0.721}&0.715&0.714\\
InternVL3.5 (8B) \cite{internvl3_5}&0.685&0.672&0.682&0.653&0.659&0.662&0.672&0.661&0.653&0.676&0.665\\
ChatGPT-4o \cite{chatgpt4o}&\mbluebf{0.739}&\mbluebf{0.732}&\mbluebf{0.736}&0.721&\mbluebf{0.715}&0.702&\mbluebf{0.714}&\mbluebf{0.715}&0.711&\mbluebf{0.717}&\mbluebf{0.721}\\
Gemini2.5-Pro \cite{nanobanana}&0.663&0.658&0.592&0.671&0.624&0.604&0.629&0.625&0.631&0.637&0.632\\
\hdashline
\rowcolor{gray!20}  
ManipShield (Ours) &\mredbf{0.826}&\mredbf{0.814}&\mredbf{0.837}&\mredbf{0.831}&\mredbf{0.823}&\mredbf{0.818}&\mredbf{0.826}&\mredbf{0.806}&\mredbf{0.814}&\mredbf{0.829}&\mredbf{0.815}  \\
 \noalign{\vspace{-0.5pt}}
\bottomrule
\end{tabular}}
\label{explain}
\end{table*}

\begin{table}[h]
\belowrulesep=0pt
\aboverulesep=0pt
\centering
\caption{Explanation capabilities comparison based on human evaluation.}
\setlength{\tabcolsep}{8pt} 
\resizebox{1\textwidth}{!}{\begin{tabular}{l||cccccccccc||c}
\toprule
\noalign{\vspace{1pt}}
& \multicolumn{10}{c}{Testing Subset}\\
\cmidrule(lr){2-11}
\noalign{\vspace{1.5pt}}
Model/Subset& SD1.4 & SD1.5 & SD3 & SDXL & FLUX & DiT & FLUX-Kontext & NanoBanana & GPT-Image &  SeedDream4&Overall \\
\midrule
\noalign{\vspace{0.5pt}}
LLaVA1.6 (7B) \cite{llava}&28.8&26.4&25.1&28.4&26.3&27.1&25.7&26.4&26.7&26.2&26.8\\
Ovis2.5 (9B) \cite{ovis25}&39.4&38.6&42.4&38.2&41.5&38.0&40.2&38.3&39.2&38.0&39.5\\
DeepSeekVL2 (small) \cite{deepseekv2}&45.8&46.2&45.7&52.5&46.0&52.4&44.6&45.7&45.2&50.6&48.6\\
Qwen3-VL (8B) \cite{qwen3}&66.6&64.8&65.7&64.8&66.2&61.4&62.8&64.2&65.5&64.2&65.3\\
InternVL3.5 (8B) \cite{internvl3_5}&68.0&67.5&66.5&64.3&68.6&64.2&65.2&65.4&67.4&63.6&65.8\\
ChatGPT-4o \cite{chatgpt4o}&77.1&75.9&74.6&\mbluebf{74.5}&\mbluebf{77.3}&74.5&76.8&\mbluebf{78.4}&\mbluebf{76.3}&77.6&\mbluebf{77.5}\\
Gemini2.5-Pro \cite{nanobanana}&\mbluebf{78.6}&\mbluebf{76.3}&\mbluebf{74.9}&70.6&72.8&\mbluebf{76.4}&\mbluebf{79.6}&75.2&75.6&\mbluebf{79.5}&77.2\\
\hline
\rowcolor{gray!20}  
ManipShield (Ours)&\mredbf{84.7}&\mredbf{85.2}&\mredbf{84.5}&\mredbf{85.6}&\mredbf{82.9}&\mredbf{84.8}&\mredbf{83.2}&\mredbf{85.0}&\mredbf{84.3}&\mredbf{83.9}&\mredbf{84.6}  \\
 \noalign{\vspace{-0.5pt}}
\bottomrule
\end{tabular}}
\label{explain_human}
\end{table}
\begin{figure*}[t]
\centering
  \includegraphics[width=1\textwidth]{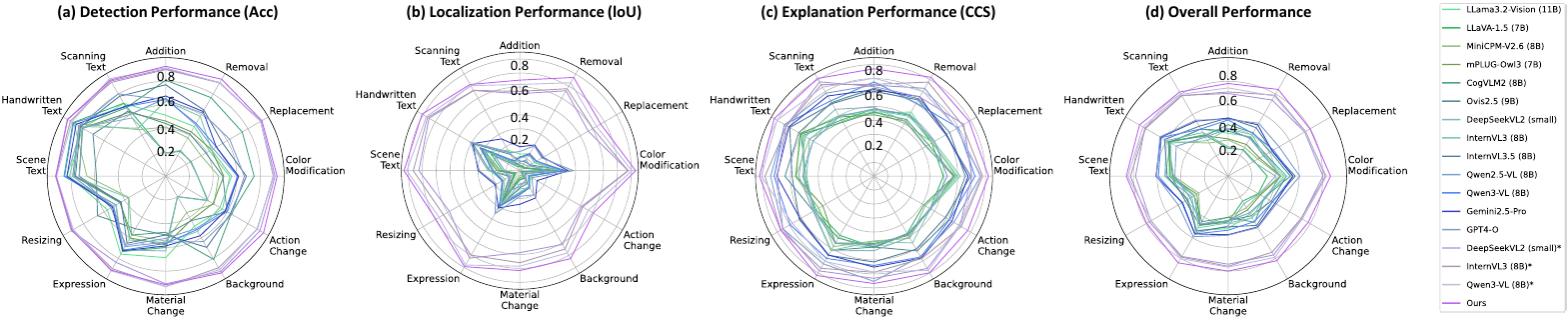}
  \caption{Comparison of performance of different MLLMs in terms of image manipulation detection, localization, and explanation, respectively.}
  \label{comparison_cat}
\end{figure*}
\begin{table}[!htbp]
\centering
\setlength{\tabcolsep}{22pt} 
\caption{Ablation study on the different backbones, decoders and LoRA tuning strategy.}
\label{ablation}
\belowrulesep=0pt
\aboverulesep=0pt
\centering
 \resizebox{1\textwidth}{!}{\begin{tabular}{@{\hspace{6pt}}lcccc|ccc}
\toprule
 \noalign{\vspace{1pt}}
\multicolumn{5}{c}{Backbone$\&$Strategy} & \multicolumn{3}{c}{Metrics} \\
\cmidrule(lr){1-5} 
\cmidrule(lr){6-8}
 \noalign{\vspace{0.5pt}}
Backbone&Decoders&  LoRA(vision) & LoRA(llm) & LDS &  Acc$\uparrow$ & loU$\uparrow$ & CSS$\uparrow$\\ 
\midrule
 \noalign{\vspace{0.5pt}}
InternVL3.5 \cite{internvl3_5} & & \checkmark &  \checkmark && 90.3 & 0.225 &0.347 \\
InternVL3.5 \cite{internvl3_5}&\checkmark &  & \checkmark & & 92.6 & 0.714 &0.852 \\
InternVL3.5 \cite{internvl3_5} & \checkmark& \checkmark &  & & 85.3 & 0.379 &0.231 \\
InternVL3.5 \cite{internvl3_5} & \checkmark& \checkmark &\checkmark  & & 93.5 & 0.732 &0.859 \\
\rowcolor{gray!20}  
InternVL3.5 \cite{internvl3_5}&\checkmark & \checkmark &  \checkmark &\checkmark & \textbf{94.7} & \textbf{0.765} &\textbf{0.892} \\
InternVL3 \cite{internvl3}&\checkmark & \checkmark &  \checkmark &\checkmark & 92.4 & 0.729 &0.846 \\
DeepSeekVL2 \cite{deepseekv2} &\checkmark & \checkmark &  \checkmark &\checkmark & 88.5 & 0.693 &0.812  \\
Qwen3-VL \cite{qwen3}&\checkmark & \checkmark &  \checkmark &\checkmark & 92.8 & 0.745 &0.861 \\
 \noalign{\vspace{-0.5pt}}
\bottomrule
\end{tabular}}
\end{table}

\section{Experiment}
\subsection{Experiment setup}
We report accuracy (Acc) and F1 for image manipulation detection, Intersection over Union (IoU) for localization, and Cosine Semantic Similarity (CSS) for explanation. A default threshold of 0.5 is applied for detection and localization. We employ a diverse set of methods for comparison. For detection, standard CNN and Transformer architectures are used as baselines, while image manipulation detection methods, AI-generated image detection methods, and MLLMs are also included in the evaluation. For localization, we compare our method with object detection approaches and MLLMs, while for explanation, we benchmark it against several advanced MLLMs. For all fine-tuned models, we adopt the same training, validation, and testing split (4:1:1), and fine-tune other MLLMs using the same strategy as our model. 
The detailed training configuration of our model is provided in Section~\ref{training_process} of supplementary material.
\subsection{Comparison results on ManipBench}


The \textbf{detection} results are shown in Table~\ref{acc1}, where in-distribution images are further grouped based on the backbones of the editing models. We observe that both image manipulation and AI-generated image detection methods perform poorly in zero-shot settings due to unseen feature distributions. Advanced MLLMs also show limited detection ability. After fine-tuning on our dataset, however, all methods improve significantly, underscoring the importance of ManipBench. ManipShield achieves the best results, while other fine-tuned MLLMs remain competitive, demonstrating the adaptability and effectiveness of our approach. Table~\ref{acc1} also presents the results for the out-of-distribution subset. Most methods show a performance drop, highlighting the challenge of generalizing to unseen editing methods, while ManipShield still demonstrates strong performance.


Table~\ref{bbox} shows the \textbf{localization} results on both the in-distribution and out-of-distribution subsets. For those manipulation methods that output masks, we infer the bounding boxes from the masks and calculate the loU results. We observe that the zero-shot performance of MLLMs is weak, indicating limited spatial reasoning ability and low sensitivity to manipulation regions. Although object detection methods perform well after training, they have difficulty predicting judgment cues as shown in Figure~\ref{example}. Furthermore, the results on the out-of-distribution subset reveal poor generalization for these methods. In contrast, our method achieves the best performance across all subsets, demonstrating strong localization capability and robust generalization to unseen manipulation patterns.

To assess the quality of \textbf{textual explanation}, we compare the performance of pre-trained MLLMs with our ManipShield. We not only compare the CSS with the explanation texts in ManipBench, as shown in Table~\ref{explain}, but also conduct a subjective study to evaluate the generated explanations. Five experts are recruited to rate the explanations. After removing outliers, the final results were obtained by averaging the remaining scores, as reported in Table~\ref{explain_human}. 
The detailed procedure is described in Supplementary Section 7.4. 
Our approach consistently achieves the best performance across all subsets. While some MLLMs can identify manipulated regions, they often struggle to precisely specify the type of inconsistency. 

Figure~\ref{comparison_cat} further presents the image manipulation detection, localization, explanation, and overall performance of MLLMs, demonstrating that ManipBench also serves as a benchmark for evaluating MLLM capabilities. Without fine-tuning, MLLMs perform poorly on regional manipulations, such as addition, removal, replacement, and resizing, likely due to their limited sensitivity to local semantic features. After fine-tuning, their performance improves across all categories, with our ManipShield still achieving superior results.

\subsection{Ablation results}
To validate the effectiveness of each component in ManipShield, we perform comprehensive ablation studies, with the results summarized in Table~\ref{ablation}. Rows 1 and 4 demonstrate the task-specific decoders in our model can dramatically improve the performance. As shown in Rows 2–4, applying LoRA to both the vision encoder and the LLM leads to the best overall performance. The layer discrimination selection (LDS) module further enhances performance, as evidenced in Row 5. Finally, Rows 5–8 compare different MLLM backbones of similar parameter scales, where the InternVL3.5 backbone used in our model achieves the best results. 
\begin{table*}[!t]
\belowrulesep=0pt
\aboverulesep=0pt
\centering
\setlength{\tabcolsep}{8pt} 
\caption{ Accuracy results (Acc$\uparrow$) of cross-validation on different training and testing subsets using ResNet50 \cite{resnet} and our ManipShield. Proprietary subset includes images generated by FLUX-Kontext \cite{fluxkontext}, NanoBanana \cite{nanobanana}, GPT-Image \cite{gptimage} and SeedDream4 \cite{seedream4}.}
\resizebox{1\textwidth}{!}{\begin{tabular}{l||cccccc||c||cccccc||c}
\toprule
\noalign{\vspace{1pt}}
Model& \multicolumn{7}{c}{ResNet-50} & \multicolumn{7}{c}{ManipShield}\\
\cmidrule(lr){2-15}
\noalign{\vspace{1.5pt}}
Train/Test on& SD1.4 & SD1.5 & SD3 & SDXL & FLUX & DiT & Proprietary&SD1.4 & SD1.5 & SD3 & SDXL & FLUX & DiT & Proprietary\\
\midrule
\noalign{\vspace{0.5pt}}
SD1.4 & \textbf{93.19} & 85.75 & 72.04 & 86.69 & 68.77 & 62.50& 58.80 &\textbf{94.31} & 90.59 & 83.33 & 85.89 & 74.10 & 73.79 &67.74\\
SD1.5 & 88.84 & \textbf{92.11} & 66.40 & 70.16 & 70.39 & 64.25 & 60.28 &91.85 & \textbf{92.92} & 69.35 & 69.09 & 79.12 & 70.43 &64.31\\
SD3 & 76.39 & 72.31 & \textbf{91.67} & 71.10 & 62.99 & 61.02 &  59.81 &80.33 & 79.21 & \textbf{90.86} & 81.59 & 67.38 & 65.59 &69.42\\
SDXL & 77.24 & 68.28 & 60.22 & \textbf{92.47} & 63.98 & 63.31 & 58.87 &80.51 & 70.88 & 78.49 & \textbf{92.47} & 65.50 & 64.92 &62.70 \\
FLUX & 71.06 & 74.73 & 70.97 & 77.42 & \textbf{93.41} & 77.15 &  63.04 &85.39 & 85.75 & 72.85 & 83.47 & \textbf{92.07} & 79.03 &73.86 \\
DiT & 69.80 & 66.13 & 66.13 & 73.66 & 71.33 & \textbf{92.61} & 63.51 &76.48 & 70.34 & 75.54 & 76.88 & 81.32 & \textbf{95.70} &76.55\\

\hdashline
\noalign{\vspace{0.5pt}}
Full Dataset& 88.80 & 88.98 & 88.84 & 88.71 & 88.62 &  86.56  &62.10 & 88.80 & 88.98 & 88.84 & 88.71 & 88.62 &  86.56 &81.53 \\
\noalign{\vspace{-0.5pt}}
\bottomrule
\end{tabular}}
\label{resnet50}
\end{table*}
\begin{table*}[!t]
\belowrulesep=0pt
\aboverulesep=0pt
\centering
\setlength{\tabcolsep}{10pt} 
\caption{Accuracy (Acc$\uparrow$) on different testing subsets, averaged over models trained on \textit{different training subsets}. The best results are highlighted in \mredbf{red}, and the second-best results are highlighted in \mbluebf{blue}.}
\resizebox{1\textwidth}{!}{\begin{tabular}{@{\hspace{6pt}}l||cccccc||cccc||c}
\toprule
\noalign{\vspace{1pt}}
 & \multicolumn{10}{c}{Testing Subset} \\
\cmidrule(lr){2-11}
\noalign{\vspace{1.5pt}}
Model& SD1.4 & SD1.5 & SD3 & SDXL & FLUX & DiT & FLUX-Kontext & NanoBanana & GPT-Image &  SeedDream4& Avg \\
\midrule
 \noalign{\vspace{0.5pt}}
ResNet50 \cite{resnet}& 79.42 & 76.55 & 71.24 & 78.58 & 71.81 & 70.14 & 59.90 & 60.30 & 58.47 & 64.20 & 69.37 \\
Swin-T \cite{swin}& 78.00 & 79.02 & 71.51 & 77.42 & 69.85 & 70.83 & 58.15 & 62.19 & 58.38 & 63.04 & 68.84 \\
MVSSNet \cite{mvss} & 76.15 & 74.34 & 65.19 & 73.70 & 66.05 & 67.18 & 59.59 & 56.90 & 57.30 & 61.29 & 65.77 \\
HiFiNet \cite{hifi}& 84.01 & 76.67 & 74.64 & 79.41 & 75.11 & 69.83 & 65.10 & 61.07 & 58.11 & 64.87 & 70.88 \\
CNNSpot \cite{cnnspot}& 76.40 & 72.25 & 71.33 & 75.16 & 71.28 & 71.82 & 57.84 & 56.50 & 57.89 & 64.34 & 67.48 \\
AIDE \cite{AIDE}& \mbluebf{82.06} & \mbluebf{77.17} & \mbluebf{76.34} & \mbluebf{80.73} & \mbluebf{75.07} & \mbluebf{71.10} & \mbluebf{65.50} & \mbluebf{64.56} & \mbluebf{64.47} & \mbluebf{68.86} & \mbluebf{72.59} \\

\hdashline
\noalign{\vspace{0.5pt}}
\rowcolor{gray!20}  
ManipShield (Ours) & \mredbf{84.81} & \mredbf{81.62} & \mredbf{78.41} & \mredbf{81.56} & \mredbf{76.58} & \mredbf{74.91} & \mredbf{71.01} & \mredbf{66.85} & \mredbf{68.32} & \mredbf{70.21} & \mredbf{75.43} \\
\noalign{\vspace{-0.5pt}}
\bottomrule
\end{tabular}}
\label{generalization_compare}
\end{table*}

\begin{table}[!t]
\belowrulesep=0pt
\aboverulesep=0pt
\centering
\setlength{\tabcolsep}{24pt} 
\caption{Comparison of performance (Acc$\uparrow$) on other image manipulation datasets. The best results are highlighted in \mredbf{red}, and the second-best results are highlighted in \mbluebf{blue}.}
\resizebox{1\textwidth}{!}{\begin{tabular}{@{\hspace{6pt}}l||cccccccc}
\toprule
\noalign{\vspace{1pt}}
Dataset& \multicolumn{2}{c}{CASIA2 \cite{casia}}& \multicolumn{2}{c}{IMD2020 \cite{IMD2020}} & \multicolumn{2}{c}{DDFT \cite{DDFT}}& \multicolumn{2}{c}{DF40 \cite{DF40}}\\
\cmidrule(lr){2-3}
\cmidrule(lr){4-5}
\cmidrule(lr){6-7}
\cmidrule(lr){8-9}
\noalign{\vspace{1pt}}
 Model/Metric&Acc$\uparrow$& F1$\uparrow$ &Acc$\uparrow$& F1$\uparrow$ &Acc$\uparrow$& F1$\uparrow$ &Acc$\uparrow$& F1$\uparrow$ \\
\midrule
\noalign{\vspace{0.5pt}}
MVSSNet \cite{mvss}&82.53&83.06 &85.25&86.24&79.46&80.26&82.71&82.54\\
PSCCNet \cite{pscc}& 89.41&87.28&80.60&82.10&85.64&86.05&86.90&87.32\\
HifiNet \cite{hifi}& 87.16&86.50&86.38&87.15&84.32&85.21&85.25&86.00\\
FakeShield \cite{fakeshield}& \mbluebf{91.65}&\mbluebf{92.11}&\mbluebf{87.43}&\mbluebf{86.25}&86.29&87.05&85.65&87.10\\
\hdashline
\noalign{\vspace{0.5pt}}
Univ \cite{univ}&86.32&87.23&84.34&86.20&\mbluebf{89.22}&\mbluebf{90.15}&85.24&85.87\\
AIDE \cite{AIDE}&88.47&90.05&86.76&87.56&85.18&88.03&\mbluebf{88.37}&\mbluebf{89.17}\\
\hdashline
\noalign{\vspace{0.5pt}}
\rowcolor{gray!20}  
ManipShield (Ours) &  \mredbf{93.71}&  \mredbf{93.08}&\mredbf{92.53}&  \mredbf{92.88}&\mredbf{91.62}&  \mredbf{93.14}&\mredbf{92.46}&\mredbf{93.25}\\
\noalign{\vspace{-0.5pt}}
\bottomrule
\end{tabular}}
\label{other}
\end{table}

\subsection{Generalization ability}
\label{generalization}
From the experimental results on ManipBench, we observe that existing image manipulation detection methods suffer from a clear performance drop when evaluated on images generated by unseen models. To further analyze this issue, we conduct experiments to assess their generalization ability across different generation backbones. Specifically, each model is trained on images from one backbone and tested on others.

As shown in Table~\ref{resnet50}, although most methods perform well on test data from the same backbone, their accuracy decreases significantly when applied to unseen backbones. This suggests that these methods tend to overfit to backbone-specific artifacts rather than learning general manipulation cues. In contrast, ManipShield captures more high-level inconsistency features and generalizes well across different generation models. Table~\ref{generalization_compare} further confirms this robustness, showing consistently strong performance across training-testing backbone shifts.

\subsection{Comparison results on other datasets}

We further evaluate the performance of ManipShield on several widely used manipulation datasets to assess its general applicability. CASIA2~\cite{casia} and IMD2020~\cite{IMD2020} are well-known manually manipulated image datasets, while DDFT~\cite{DDFT} and DF40~\cite{DF40} are representative deepfake datasets focusing on identity alteration. We adopt the same training and testing split (4:1) for all datasets to ensure fair comparison. As shown in Table~\ref{other}, ManipShield achieves best results, demonstrating its strong generalization ability and potential as a unified solution for general image manipulation detection.

\section{Conclusion}
In this paper, we introduce ManipBench, a large-scale benchmark for image manipulation detection focusing on AI-edited images, containing 450K+ manipulated samples and 100K+ fine-grained annotations. Based on ManipBench, we propose ManipShield, a unified model for image manipulation detection, localization, and explanation. Extensive experiments demonstrate that ManipShield overcomes existing methods, showing state-of-the-art performance and strong generalization ability.

\section{Acknowledgement}
This work was supported in part by the National Natural Science Foundation of China under Grants 62401365, 62225112, 62522116, 62322302, 62273011, 62271312, 62132006, and U24A20220, and in part by the China Postdoctoral Science Foundation under Grants BX20250411 and 2025M773473, and in part by the National Key Research and Development Program of China under Grant 2024YFE0100500.

\newpage
\clearpage
\bibliographystyle{splncs04}
\bibliography{main}
\newpage
\clearpage
\clearpage
\begin{center}
\Large\textbf{Supplementary Material}
\end{center}
\maketitle
\section{Overview}
\label{sec:rationale}
This supplementary material provides additional information on the data collection, methodology, experiments, and results discussed in the main paper. Section~\ref{related_work} presents the related work. Section~\ref{image collection} describes the image collection process, including the 12 distinct tasks for prompt and 25 image manipulation models for generation. Section~\ref{annotation_more} elaborates on the manual annotation process, including the annotation standards, interface design, and management strategy. Section 5 provides an in-depth analysis of the ManipBench dataset. Section 6 details the hyper-parameters and loss functions used in training the ManipShield model. Finally, Section 7 presents additional results and comparisons between models.

\section{Image collection}
\label{image collection}
\subsection{Manipulation tasks define}

In this study, we conducted a comprehensive investigation into various categories of image modification, which we systematically divided into three main dimensions: semantic, visual, and textual. Across these dimensions, a total of 12 distinct manipulation types were defined to capture diverse editing behaviors at different levels of abstraction.

Within the semantic dimension, modifications primarily involve changes to image content and meaning, including addition, removal, replacement, and action change, which reflect variations in object presence, relationships, and contextual semantics.

The visual dimension focuses on perceptual and appearance-level edits such as background, color, material, expression, and resize, encompassing adjustments to aesthetic or structural aspects of the image.

Finally, the textual dimension addresses manipulations related to embedded text within images, encompassing scene text, handwritten text, and scanning text, which involve recognition, alteration, or regeneration of text in different modalities.
This structured categorization provides a fine-grained framework for analyzing and benchmarking image-editing models, enabling a deeper understanding of model performance across diverse manipulation types.

\begin{itemize}
    \item \textbf{Addition}: This task involves inserting new objects or regions into an existing image. Detecting such modifications often relies on identifying inconsistent object boundaries, lighting disparities, or unnatural blending between the newly added region and its surrounding context.

    \item \textbf{Removal}: This task removes existing objects or regions from an image. Detection methods typically focus on recognizing unnatural background reconstruction patterns, texture repetitions, or inpainting artifacts in the removed area.

    \item \textbf{Replacement}: This task substitutes one object or region with another. Identifying replacements often involves detecting semantic mismatches between the object and its environment, such as inconsistent shading, geometry, or contextual incongruence.

    \item \textbf{Action Change}: This task alters the pose or activity of a subject (e.g., changing a standing person to a sitting one). Detection commonly depends on analyzing spatial or anatomical inconsistencies, unnatural body part alignments, or motion artifacts in regions of change.

    \item \textbf{Background}: This task replaces or edits the scene background while preserving the main subjects. Detection methods can utilize boundary consistency checks, depth discontinuities, or background–foreground semantic mismatches to localize the modification.

    \item \textbf{Color}: This task changes the color properties of objects or regions. Such modifications can be detected through abnormal color histograms, inconsistent illumination cues, or deviations from material-specific color distributions.

    \item \textbf{Material}: This task modifies an object’s surface material (e.g., wood to metal). Detection focuses on identifying inconsistencies in reflectance, glossiness, or fine texture gradients that contradict the expected material properties.

    \item \textbf{Expression}: This task alters facial expressions while retaining identity. Detection often involves subtle analysis of facial muscle movements, local geometric deformations, and texture discontinuities in high-frequency facial regions.

    \item \textbf{Resizing}: This task resizes objects or regions in the image. Detection can rely on geometric distortion analysis, perspective inconsistency, or mismatched scaling ratios among related scene elements.

    \item \textbf{Scene Text}: This task edits or replaces text in natural scenes. Detection generally targets inconsistencies in font alignment, lighting adaptation, or perspective transformation of the rendered text relative to its surface.

    \item \textbf{Handwritten Text}: This task modifies handwritten content. Such edits can be detected through irregularities in stroke continuity, pen pressure distribution, or inconsistent texture blending between the handwriting and background.

    \item \textbf{Scanning Text}: This task edits text in scanned or printed documents. Detection methods focus on recognizing pixel-level noise patterns, font irregularities, misalignments, or scanning artifacts indicative of post-processing.
\end{itemize}

%
%

\subsection{Details of manipulation methods}
\label{editing_model}
\begin{table*}[t]
\caption{An overview of AI-editing models in our ManipBench}
\centering
\makebox[\textwidth][c]{%
\small
\resizebox{1\textwidth}{!}{\begin{tabular}{lcccc}
\toprule
\textbf{Models} & \textbf{Time} & \textbf{Prompt Type} & \textbf{Method} & \textbf{URL} \\
\midrule
IP2P \cite{ip2p} & 2023.04 & Instruction & SD1.4 \cite{SD} & \url{https://github.com/timothybrooks/instruct-pix2pix} \\
CDS \cite{CDS}& 2023.11 & Description & SD1.4 \cite{SD} & \url{https://github.com/HyelinNAM/ContrastiveDenoisingScore} \\
Magicbrush \cite{Magicbrush} & 2023.06 & Instruction & SD1.4 & \url{https://github.com/OSU-NLP-Group/MagicBrush} \\
PnP \cite{PnP} & 2023.10 & Description & SD1.5 \cite{SD}& \url{https://github.com/MichalGeyer/plug-and-play} \\
Any2Pix \cite{instructany2pix}& 2023.12 & Instruction & SDXL \cite{sdxl}& \url{https://github.com/jacklishufan/InstructAny2Pix} \\
InfEdit \cite{InfEdit} & 2023.12 & Description & SD1.4 \cite{SD}& \url{https://github.com/sled-group/InfEdit} \\
MAG \cite{MAG}& 2023.12 & Description & SD1.4 \cite{SD}& \url{https://github.com/HelenMao/MAG-Edit} \\
ZONE \cite{zone} & 2023.12 & Instruction & SD1.5 \cite{SD}& \url{https://github.com/lsl001006/ZONE} \\
PowerPaint \cite{PPT} & 2023.12 & Description & SD1.5 \cite{SD} & \url{https://github.com/open-mmlab/PowerPaint} \\
ReNoise \cite{renoise}& 2024.03 & Description & SDXL \cite{sdxl} & \url{https://github.com/garibida/ReNoise-Inversion} \\
BrushNet \cite{brushnet}& 2024.03 & Description & SD1.5 \cite{SD}& \url{https://github.com/TencentARC/BrushNet} \\
HQEdit \cite{HQ} & 2024.04 & Instruction & SD1.4 \cite{SD}  & \url{https://github.com/UCSC-VLAA/HQ-Edit} \\
RFSE \cite{RFSE} & 2024.11 & Description & FLUX \cite{FLUX}& \url{https://github.com/wangjiangshan0725/RF-Solver-Edit} \\
FlowEdit (SD3) \cite{Flowedit} & 2024.12 & Description & SD3 \cite{SD}& \url{https://github.com/fallenshock/FlowEdit} \\
FlowEdit (FLUX) \cite{Flowedit} & 2024.12 & Description & FLUX \cite{FLUX}& \url{https://github.com/fallenshock/FlowEdit} \\
ACE++ \cite{ACE} & 2025.01 & Instruction & FLUX \cite{FLUX}& \url{https://github.com/ali-vilab/ACE_plus} \\
OmniGen2 \cite{omnigen2}& 2025.01 & Instruction & DiT \cite{DiT}& \url{https://github.com/VectorSpaceLab/OmniGen2} \\
Reflex \cite{reflex} & 2025.07 & Description & FLUX \cite{FLUX}& \url{https://github.com/wlaud1001/ReFlex} \\
OT-Inversion \cite{OT}& 2025.08 & Description & FLUX \cite{FLUX}& \url{https://github.com/marianlupascu/OT-Inversion} \\
Follow-Your-Shape \cite{followyourshape}& 2025.08 & Description & FLUX \cite{FLUX}& \url{https://github.com/mayuelala/FollowYourShape} \\
Qwen-Image-Edit \cite{qwenedit}& 2025.08 & Instruction & DiT \cite{DiT} & \url{https://github.com/QwenLM/Qwen-Image} \\
GPT-Image \cite{gptimage} & 2025.04 & Instruction & N/A & \url{https://chatgpt.com/} \\
FLUX-Kontext \cite{fluxkontext} & 2025.06 & Instruction & FLUX \cite{FLUX}& \url{https://github.com/black-forest-labs/flux} \\
NanoBanana \cite{nanolink}& 2025.07 & Instruction & N/A & \url{https://ainanobanana.io/} \\
SeedDream4 \cite{seedream4}& 2025.09 & Instruction & N/A & \url{https://dreamina.capcut.com/} \\
\bottomrule
\end{tabular}}
}
\vspace{1em}
\end{table*}

\begin{itemize}

\item \textbf{IP2P \cite{ip2p}}
is a conditional diffusion model designed for image editing guided by human-written instructions. It is trained on a large synthetic dataset generated by combining GPT-3 and Stable Diffusion, enabling the model to generalize to real images and unseen prompts without per-example fine-tuning or inversion. By conditioning on both the input image and the editing instruction, IP2P enables fast, high-quality edits in seconds while preserving the original structure.

\item \textbf{CDS \cite{CDS}} 
introduces the Contrastive Denoising Score for text-guided latent diffusion image editing. Unlike Delta Denoising Score, CDS incorporates contrastive learning over intermediate self-attention features of latent diffusion models, preserving structural elements of the source image and enabling zero-shot, structure-consistent image-to-image translation.

\item \textbf{MagicBrush \cite{Magicbrush}} 
is the first large-scale, manually annotated dataset for instruction-guided real image editing. It provides diverse editing pairs under single-turn, multi-turn, mask-provided, and mask-free conditions. Fine-tuning models such as IP2P on MagicBrush improves human-rated edit quality and realism.

\item \textbf{PnP \cite{PnP}} 
is a description-based image editing system built upon Stable Diffusion 1.5, enabling flexible global image transformation from descriptive textual inputs. It maintains high realism and structural consistency across varied global domains.

\item \textbf{Any2Pix \cite{instructany2pix}} 
is a multi-modal instruction-following image editing framework supporting text, image, and audio inputs. It employs a unified multi-modal encoder, a diffusion decoder, and an LLM-based instruction interpreter to execute complex edits with high fidelity across global domains.

\item \textbf{InfEdit \cite{InfEdit}} 
is an inversion-free, text-guided diffusion editing method that overcomes the inefficiency of inversion-based pipelines. Using a Denoising Diffusion Consistent Model and unified attention control, it enables fast, faithful, and structure-preserving global edits without explicit inversion.

\item \textbf{MAG \cite{MAG}} 
is a diffusion-based inpainting model tailored for local edits guided by text descriptions. It ensures coherent blending between masked and unmasked regions, maintaining global context and high-quality restoration.

\item \textbf{ZONE\cite{zone}} 
is a zero-shot, instruction-guided local editing framework that translates textual instructions into editable regions. Through segmentation-based region extraction and FFT-based edge smoothing, it achieves seamless, localized edits while preserving non-edited content.

\item \textbf{PowerPaint\cite{PPT}} 
is a diffusion-based inpainting system emphasizing semantic consistency and fine-grained texture recovery. It performs text-driven completion for masked areas, maintaining seamless integration within the global image structure.

\item \textbf{ReNoise \cite{renoise}} 
is a description-driven global editing framework that improves robustness to noisy inputs through refined noise modeling and enhanced denoising strategies, achieving high-fidelity, structure-consistent edits under degraded conditions.

\item \textbf{BrushNet \cite{brushnet}} 
is a diffusion-based inpainting model with brush-like latent operations for masked region editing. It introduces hierarchical attention and spatially adaptive blending to maintain smooth transitions between edited and unedited areas.

\item \textbf{HQEdit \cite{HQ}} 
is an instruction-guided global editing model optimized for high-quality (HQ) outputs. It features enhanced instruction parsing and guided sampling strategies to produce professional-grade, photorealistic edits with strong structural fidelity.

\item \textbf{RFSE \cite{RFSE}} 
is a description-based editing framework leveraging the FLUX backbone. By exploiting its dynamic latent representation, RFSE enables robust global image transformations with high controllability and semantic preservation.

\item \textbf{FlowEdit (SD3)\cite{Flowedit}} 
is a flow-based description-driven editing model using Stable Diffusion 3. It performs global edits through latent flow interpolation and adaptive guidance, maintaining structural coherence and semantic alignment.

\item \textbf{FlowEdit (FLUX) \cite{FLUX}} 
is a flow-based description-driven editing model using FLUX.1, combining flow-based latent editing with FLUX’s expressive diffusion space for precise, high-fidelity global edits.

\item \textbf{ACE++ \cite{ACE}} 
is an instruction-guided editing framework built on the FLUX backbone. It integrates a refined instruction encoder and fast latent adaptation mechanism to achieve real-time, high-quality global edits with strong semantic consistency.

\item \textbf{OmniGen2 \cite{omnigen2}} 
is an instruction-guided image editing model using a Diffusion Transformer (DiT) backbone. It unifies semantic reasoning and generation, supporting complex global edits driven by natural language instructions.

\item \textbf{Reflex \cite{reflex}} 
is a description-based global editing model that captures stylistic or emotional cues in textual prompts. By conditioning latent style embeddings on descriptions, it produces coherent mood-driven transformations while preserving content fidelity.

\item \textbf{OT-Inversion \cite{OT}} 
introduces an optimal-transport-based inversion method for latent diffusion models. It computes latent flows between source and target semantics, enabling faithful reconstruction and description-guided global editing.

\item \textbf{Follow-Your-Shape \cite{followyourshape}} 
is a shape-conditional description-based editing model that integrates contour guidance with text-driven FLUX diffusion. It ensures shape alignment and structure preservation in globally consistent edits.

\item \textbf{Qwen-Image-Edit \cite{qwenedit}} 
is an instruction-based image editing system aligned with the Qwen multi-modal model family. Built on a DiT backbone, it interprets high-level natural language instructions for semantically consistent global edits.

\item \textbf{GPT-Image \cite{gptimage}} 
is an instruction-based image editing interface that connects a large language model with a diffusion editing pipeline. It translates user intentions into image transformations, abstracting backbone specifics while ensuring global semantic coherence.

\item \textbf{FLUX-Kontext \cite{fluxkontext}} 
is a context-aware instruction-guided editing model supporting multi-turn interactions. It tracks editing history and adapts latent states within the FLUX framework to maintain consistency across iterative global edits.

\item \textbf{NanoBanana \cite{nanobanana}} 
is a lightweight instruction-based diffusion model optimized for speed and interactivity. It performs fast global edits guided by simple textual commands while maintaining structural coherence.

\item \textbf{SeedDream4 \cite{seedream4}} 
is an instruction-guided global editing model emphasizing seed-based control for reproducible edits. By coupling user instructions with seed conditioning, it achieves semantically faithful and consistent transformations across runs.
\end{itemize}

\section{Details of manual annotation process}
\label{annotation_more}
\begin{figure}[t]
\centering
  \includegraphics[width=0.8\textwidth]{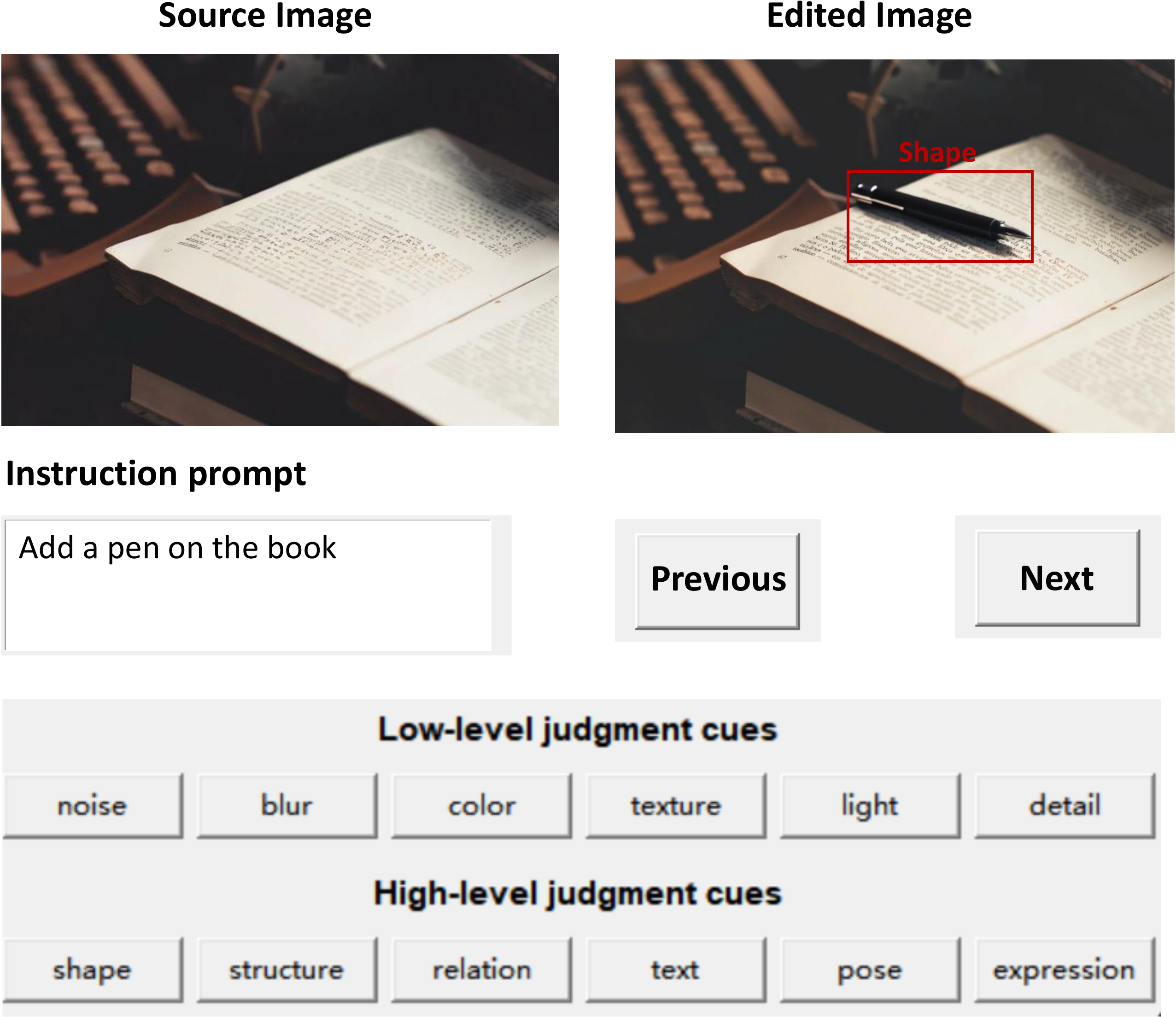}
  \caption{An example of the annotation interface.}
  \label{gui}
\end{figure}

\subsection{Localization annotation}
To ensure an efficient manual annotation, we developed a custom annotation interface, as illustrated in Figure~\ref{gui}. The manual evaluation platform was implemented using the Python Tkinter package, featuring an intuitive and lightweight design to facilitate large-scale manipulation localization and cue labeling. 

During the annotation process, human evaluators are presented with paired real and manipulated images and are instructed to draw bounding boxes around all perceptual semantically modified regions in the manipulated image. Here, semantic changes also include alterations in texture, color, or other visual attributes that differ from the original identity, even if the object category remains the same. For example, if a cat in the edited image is clearly not the same as the cat in the original image, it should still be considered a semantic change, even though it remains a cat. We emphasize perceptual changes because modifications that do not produce perceptible semantic differences are unlikely to be significant and generally do not pose safety or functional risks. The bounding box should encompass the entire semantic object and closely fit its boundaries. The overlap of different bounding boxes are allowed. For each annotated region, annotators also specify the judgment cues that can help identify the manipulation. These cues are categorized into two main groups: high-level cues and low-level cues.

\vspace{1em}
\noindent\textbf{High-level judgment cues} capture semantic and structural information of the scene, including object geometry, spatial relations, and human expressions. They reflect how AIGC models understand and reconstruct global semantics, and inconsistencies in these cues often reveal a lack of real-world reasoning or contextual coherence in edited images. Specifically, the following aspects represent typical high-level cues affected by AIGC editing:
\begin{itemize}[left=12pt, labelsep=0.6em, labelwidth=0pt]
    \item \textbf{Shape} cues describe the overall contour and geometry of objects in an image. AIGC editing may unintentionally distort object proportions—such as making limbs longer or altering the curvature of structures—leading to subtle yet detectable inconsistencies in global shape.
    \item \textbf{Structure} cues capture the spatial arrangement and connectivity among image components. Manipulations often disrupt these relationships, e.g., misaligned building elements or inconsistent object layouts caused by incomplete structural understanding of generative models.
    \item \textbf{Relation} cues focus on semantic and spatial relations among multiple objects. AIGC models sometimes generate objects that violate real-world logic—like inconsistent reflections or lighting directions—revealing manipulation through relational incoherence.
    \item \textbf{Text} cues are often difficult for generative models to render accurately. AIGC-edited images may show misspelled or distorted characters, irregular font spacing, or unnatural alignment, which serve as reliable traces of synthetic content.
    \item \textbf{Pose} cues reflect the geometric configuration of human or animal bodies. Manipulated or generated subjects may exhibit unnatural joint angles or inconsistent body alignment, as AIGC models occasionally fail to preserve plausible 3D pose relationships.
    \item \textbf{Expression} cues relate to fine-grained facial or emotional details. Editing may alter expressions inconsistently across regions—such as mismatched mouth and eye emotions—making emotional coherence an important signal for manipulation detection.
\end{itemize}
\textbf{Low-level judgment cues} focus on pixel-level visual statistics such as color, texture, noise, and lighting. They describe the appearance consistency between manipulated and authentic regions, where AIGC-generated content often exhibits deviations in fine-grained details or photometric properties. The following cues illustrate common low-level inconsistencies in AIGC-edited images:
\begin{itemize}[left=12pt, labelsep=0.6em, labelwidth=0pt]
    \item \textbf{Texture} cues describe surface granularity and local visual patterns. AIGC-edited regions often show over-smoothed or repetitive textures due to limited high-frequency detail synthesis, distinguishing them from authentic image regions.
    \item \textbf{Blur} cues reflect local focus and motion consistency. Manipulated content may exhibit inconsistent blur levels compared with surrounding areas, especially when compositing objects captured under different depth-of-field or motion conditions.
    \item \textbf{Noise} cues reveal sensor-level statistics that are hard for generative models to reproduce. Inconsistencies in noise distribution—such as missing camera noise or abnormal color noise patterns—can expose synthetic or edited areas.
    \item \textbf{Light} cues indicate illumination direction, color, and intensity. AIGC-based edits frequently fail to align lighting conditions between inserted and background regions, producing mismatched shadows or specular reflections.
    \item \textbf{Detail} cues correspond to fine local structures such as hair, texture edges, or reflections. AIGC editing may suppress or exaggerate these details, making local realism inconsistent with the overall image fidelity.
    \item \textbf{Color} cues capture chromatic harmony and tone consistency. Generative editing might produce unrealistic saturation, hue shifts, or mismatched white balance, especially when blending objects from different scenes.
\end{itemize}
In practical annotation, annotators can also select "other" to mark cues they consider not included in the predefined categories. However, upon review by expert groups, these annotations were found to also fit into the categories above.

All candidate annotators are required to undergo a 30-minute training session to become familiar with the annotation task and the marking rules. To evaluate their annotation quality, we have constructed a small image set consisting of 50 manipulated images that cover all types of judgment cues, each meticulously annotated by an expert group. Candidate annotators are required to complete annotations on this set, and only those whose annotations achieved an Intersection over Union (IoU) score above 0.7 with the expert annotations are qualified to participate in the main study. Based on this process, a total of 23 annotators were ultimately employed.

All annotations are conducted using a custom Python-based graphical interface displayed on a professionally calibrated LED monitor with a native resolution of 3840 × 2160. The monitor is color-calibrated to ensure consistent brightness, contrast, and color reproduction across sessions. Each image is presented in a randomized order to prevent potential ordering bias or learning effects during the annotation process. The annotation process is conducted under standardized laboratory conditions, with ambient illumination maintained at approximately 20 lux to minimize glare and reflections. Annotators are seated at a distance of approximately 2 feet from the display to ensure consistent visual perception and comfortable viewing. To mitigate visual fatigue and ensure annotation consistency, the entire process is divided into 15 sessions, each lasting no longer than 30 minutes, with mandatory short breaks between sessions. The order of the image sets assigned to each session is also randomized across annotators to further minimize potential contextual influence.

To ensure that the perceptual change is consistent with most people, we adopt a three-stage annotation protocol that integrates independent labeling, cross-verification, and expert voting.
\begin{itemize}[left=12pt, labelsep=0.6em, labelwidth=0pt]
    \item \textbf{Independent labeling:} Each manipulated image is annotated by two annotators. Each annotator independently labels the manipulated regions and corresponding judgment cues without reference to others’ annotations, ensuring unbiased initial results.
    \item \textbf{Cross-verification:} For samples with IoU below 0.9, conflicting judgment cues or different bounding box number between annotators, a re-annotation is performed after discussion between the two person to ensure consistency and data quality. For all other samples, the final bounding box is obtained by averaging the annotations from the two annotators. 
    \item \textbf{Expert Voting} Finally, all annotations were reviewed by 5 experts, and any items receiving fewer than 4 votes were discarded to ensure alignment with human perception. In this process, 27\% of the annotations were removed.
\end{itemize}
The independent labeling and cross-verification required 1,800 working hours, while expert voting took an additional 650 working hours. This rigorous protocol guarantees that the resulting annotations are reliable, consistent, and perceptually meaningful, forming a high-quality foundation for subsequent model training and evaluation.

\subsection{Explanation annotation}
For explanation annotation, annotators are presented with the manipulated image, the corresponding annotated bounding box, judgment cues, and the textual explanation generated by ChatGPT-4o \cite{chatgpt4o}. Ideal explanations should mention all regions within the annotated bounding boxes, cover accurate judgment cues, be linguistically fluent, logically coherent, and free of irrelevant information. Annotators can either accept the provided text, rewrite or delete some words, or seek a regeneration of the LLM draft.

Each image is reviewed by at least two annotators independently. If both annotators approve the draft, it is retained as the final explanation. If one annotator revise the text, the revised explanation is subsequently reviewed by three additional annotators. This process continued iteratively until at least three annotators agreed on the final version, ensuring consensus and alignment with human judgment.

All candidate annotators were required to undergo a training session to familiar with the annotation guidelines, criteria for judgment cues, and standards for textual explanations. The annotation process is conducted under the same controlled laboratory conditions as the localization annotation to minimize environmental variability. In total, this annotation procedure required approximately 750 hours.

\section{More data analysis of ManipBench}
\begin{figure*}[t]
\centering
  \includegraphics[width=0.85\textwidth]{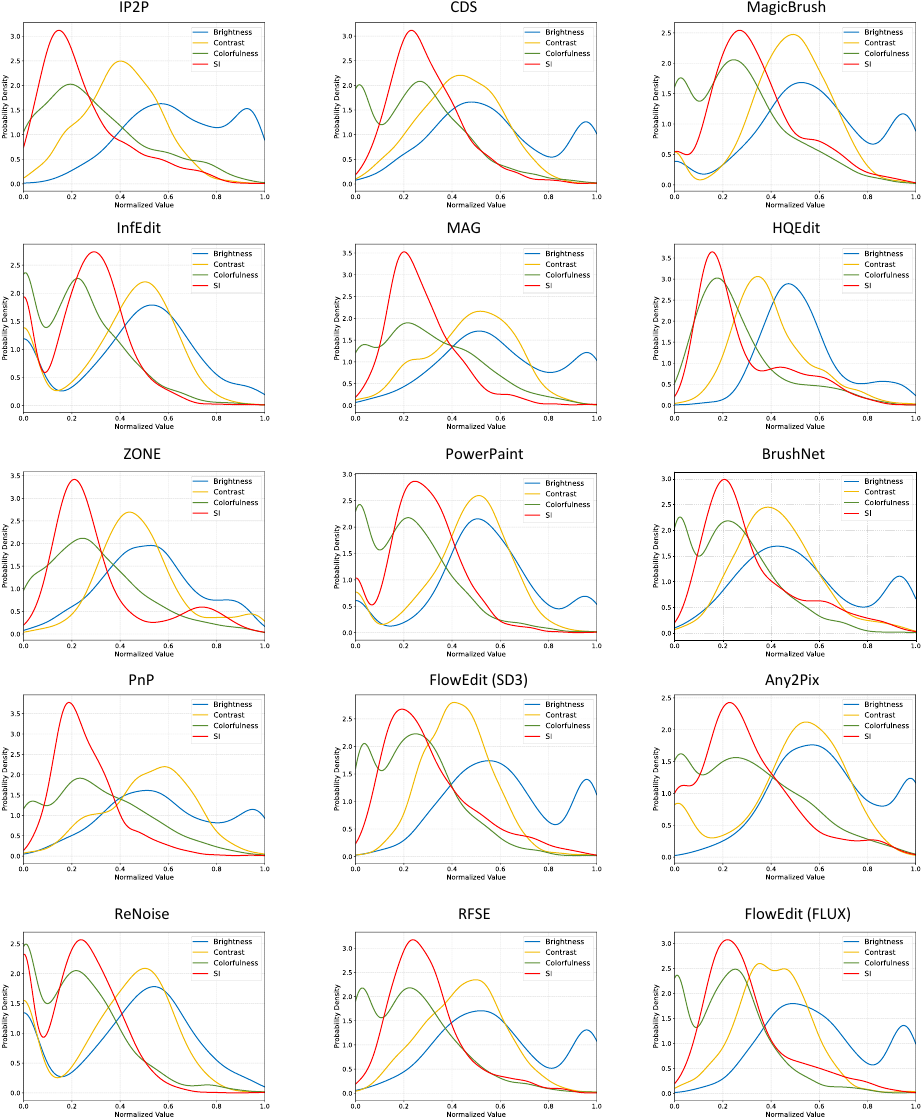}
  \caption{Feature distribution of images generated by different manipulation models.}
  \label{info_all}
\end{figure*}

\begin{figure*}[t]
\centering
    \includegraphics[width=0.85\textwidth]{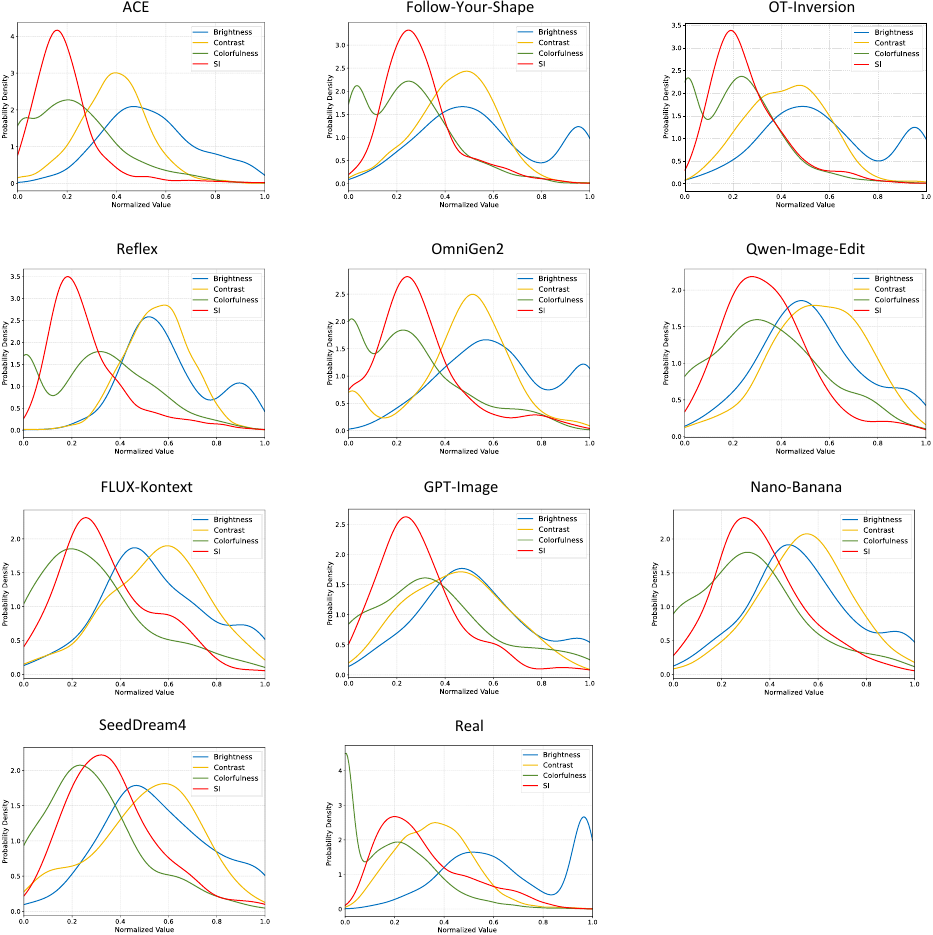}
  \caption{Feature distribution of images generated by different manipulation models and real images.}
  \label{info_all2}
\end{figure*}

As shown in Figure~\ref{info_all} and \ref{info_all2}, we visualize the distributions of brightness, contrast, colorfulness, and spatial information (SI) of images generated by different editing models and real images to assess their visual characteristics and consistency. It is noteworthy that some source images inherently exhibit high brightness or low colorfulness, as certain text-related images are originally black-and-white, which may partially influence the overall distribution of the generated results. Nevertheless, several consistent trends can still be observed across different models.

Compared with real images, models such as IP2P \cite{ip2p}, CDS \cite{CDS}, and MagicBrush \cite{Magicbrush} tend to generate outputs with brightness levels slightly higher than those of other generation models but still below those of real images. This suggests that these models more closely approximate the illumination characteristics of natural scenes, producing visually balanced and realistic outputs. In contrast, Qwen-Image-Edit \cite{qwenedit} and several closed-source models yield smoother and more uniformly distributed brightness, reflecting enhanced tonal and chromatic consistency, albeit with reduced variability.

In terms of contrast, closed-source models generally achieve higher and more stable contrast, likely due to more advanced feature normalization or enhancement mechanisms. However, while excessive contrast can exaggerate textures and boundaries, moderate contrast contributes to more natural and perceptually pleasing results.

Regarding colorfulness, although some input images are intrinsically unsaturated, all models tend to produce more saturated outputs than real images. Open-source models such as PowerPaint \cite{PPT}, ReNoise \cite{renoise}, and InfEdit \cite{InfEdit} exhibit larger variance, indicating less stable color calibration. In contrast, closed-source models generate higher yet smoother and more perceptually consistent colorfulness, suggesting better chromatic adjustment and adaptation to diverse input styles. Notably, the overall increase in colorfulness across generated images may serve as a potential statistical cue for distinguishing real from manipulated images.

Finally, spatial information (SI), which reflects the richness of structural and textural details, further differentiates model behaviors. Models such as ReNoise and InfEdit often produce smoother outputs with relatively low SI, whereas advanced approaches like Qwen-Image-Edit \cite{qwenedit}, FLUX-Kontext \cite{fluxkontext}, and SeedDream4 \cite{seedream4} exhibit consistently higher SI values—sometimes even exceeding those of real images, indicating stronger preservation of fine details and local texture fidelity.

\section{Details of training process}
\label{training_process}
\subsection{Details of hyperparameters}
This configuration includes several key hyperparameters that substantially influence the model’s performance during both training and evaluation. The model is trained in two stages. In the first stage, the vision encoder is trained using Contrastive LoRA Fine-tuning to enhance its ability to capture manipulation-related visual semantics while maintaining parameter efficiency. In the second stage, the vision encoder is frozen, and the large language model is fine-tuned together with the specified decoders and LoRA modules to align visual features with textual understanding and reasoning. The hyperparameters for both stages are summarized as follows:
\begin{itemize}[left=12pt, labelsep=0.6em, labelwidth=0pt]
    \item \textbf{Learning Rate (5e-6/1e-5):} The learning rate determines the step size at each iteration while moving toward the minimum of the loss function. A smaller learning rate allows the model to converge more slowly but with higher precision, whereas a larger learning rate can speed up convergence at the risk of overshooting the optimal solution. In this work, the learning rate is set to 1e-5, which is relatively small and suitable for fine-tuning pre-trained models.

    \item \textbf{Batch Size (32/16):} This hyperparameter defines how many samples are processed together in one forward and backward pass. A batch size of 1 means that each sample is processed individually, which trades off speed for lower memory usage and ensures stable training on GPUs with limited memory capacity.

    \item \textbf{Warm-up Ratio (0.05):} This parameter specifies the proportion of the total training steps during which the learning rate linearly increases from 0 to the initial value. It helps stabilize the optimization process, particularly during the early stages of training.

    \item \textbf{Weight Decay (0.1):} Weight decay is a regularization technique that prevents overfitting by penalizing large parameter magnitudes. A value of 0.1 indicates that model weights are reduced by 10\% during each update step, improving generalization on unseen data.

    \item \textbf{LoRA Rank (16):} The rank determines the dimension of the low-rank adaptation matrices in the LoRA module. A higher rank allows greater adaptation flexibility, while a lower rank reduces parameter overhead. A rank of 16 offers a balanced tradeoff between expressiveness and efficiency.

    \item \textbf{LoRA Alpha (32):} This scaling factor controls the update strength of the LoRA parameters relative to the base model weights. Setting LoRA alpha to 32 amplifies the contribution of the low-rank updates, enabling more effective fine-tuning without increasing the total number of trainable parameters significantly.
\end{itemize}
\subsection{Details of layer discrimination selection}
\label{LDS}
\begin{figure*}
    \centering
    \includegraphics[width=0.9\linewidth]{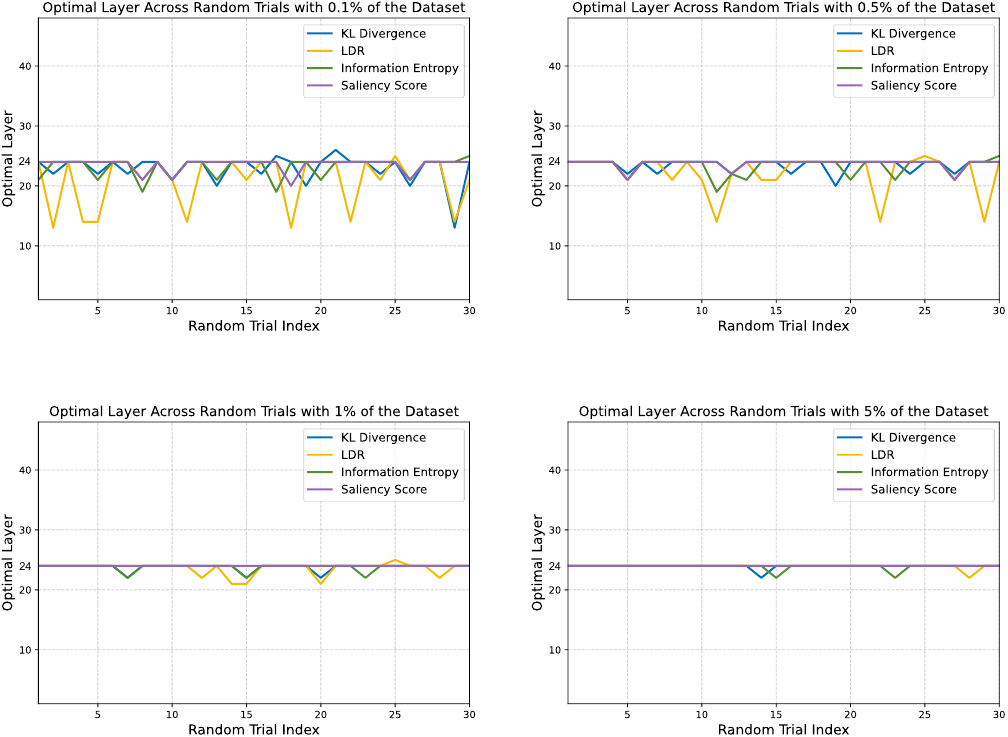}
    \caption{Optimal layer variation across random trials for KL divergence, LDR, information entropy, and saliency score under different sample scales.}
    \label{LDS_more}
\end{figure*}

We combine KL divergence, LDR, and information entropy to select the optimal layer. Figure~\ref{LDS_more} shows that the optimal layer determined by individual metrics fluctuates when the sample size is low. With 0.1\% and 0.5\% of the dataset, both the individual metrics and the combined saliency score vary considerably, indicating that such few samples are insufficient for reliable layer selection. Using 1\% of the dataset (around 4K samples), the saliency score stabilizes, although the individual metrics still fluctuate. With 5\% of the dataset, all metrics are nearly stable, but at a higher computational cost. Therefore, combining the three metrics into a saliency score allows reliable identification of the optimal layer with fewer samples.
\subsection{Details of loss functions}
The training of our model is organized into two sequential stages to gradually build discriminative and task-aware representations. In the first stage, the focus is on learning robust and generalizable visual representations that can effectively distinguish real images from manipulated ones. In the second stage, the model is fine-tuned for multi-task objectives, including binary detection, bounding box regression, and judgment cue classification, thereby enhancing both global and local reasoning capabilities.

\paragraph{Stage 1: contrastive learning}  
In the first training stage, a contrastive loss is employed to enforce the model to separate real and manipulated samples in the feature space. Let the feature embeddings of a positive pair $(i, j)$ be denoted as $f_i$ and $f_j$, where each pair consists of a real image and its manipulated counterpart. The embeddings of all other images within the same batch are denoted as $f_k$ ($k \neq i,j$) and serve as negative samples. The contrastive loss is formulated as:
\begin{equation}
\mathcal{L}_{\mathrm{contrastive}} = - \frac{1}{N} \sum_{i=1}^{N} 
\log \frac{\exp(\mathrm{sim}(f_i, f_j)/\tau)}
{\sum_{k \neq i, j} \exp(\mathrm{sim}(f_i, f_k)/\tau)},
\end{equation}
where $\mathrm{sim}(\cdot)$ denotes the cosine similarity function, $\tau$ is a temperature hyperparameter controlling the concentration level of the distribution, and $N$ is the batch size. All feature embeddings are $\ell_2$-normalized before computing similarity to ensure numerical stability.  
This stage aims to cluster semantically similar features (i.e., real–manipulated pairs) closer in the latent space while pushing dissimilar pairs apart, encouraging the model to capture manipulation-sensitive patterns that are invariant to irrelevant appearance factors such as lighting or texture.

\paragraph{Stage 2: Multi-Task fine-Tuning}  
After the representation space is pre-trained with contrastive supervision, the model is jointly optimized for three downstream objectives:  
(1) to determine whether an image is real or manipulated (binary detection),  
(2) to localize manipulated regions through bounding box regression, and  
(3) to identify the manipulation-related judgment cue category (e.g., light, color, texture).  

The first task is formulated as a binary classification problem optimized using the Focal loss, which focuses training on hard or misclassified examples and mitigates class imbalance between real and manipulated samples. The second task is a regression problem optimized using the mean squared error (MSE) loss to ensure accurate localization of manipulated areas. The third task employs a standard cross-entropy loss to supervise the judgment cue classification. The overall multi-task loss function is defined as:
\begin{equation}
\mathcal{L}_{total} = \lambda_1 \mathcal{L}_{binary} + \lambda_2 \mathcal{L}_{bbox} + \lambda_3 \mathcal{L}_{cue},
\end{equation}
where $\mathcal{L}_{binary}$, $\mathcal{L}_{bbox}$, and $\mathcal{L}_{cue}$ denote the binary judgment loss, bounding box regression loss, and cue classification loss, respectively. 
The weighting factors $\lambda_1$, $\lambda_2$, and $\lambda_3$ are empirically determined to balance the gradient magnitudes of different objectives.

The detailed formulations of these losses are as follows:
\begin{equation}
\mathcal{L}_{binary} = -\alpha (1 - p_t)^{\gamma} \log(p_t),
\end{equation}
\begin{equation}
\mathcal{L}_{bbox} = \frac{1}{N} \sum_{i=1}^{N} (B_i^{pred} - B_i^{gt})^2,
\end{equation}
\begin{equation}
\mathcal{L}_{cue} = - \frac{1}{C} \sum_{c=1}^{C} y_c \log(p_c),
\end{equation}
where $p_t$ represents the predicted probability of the ground-truth class in binary detection, and $\alpha$ and $\gamma$ are the balance and focusing parameters of the Focal loss. 
$B_i^{pred}$ and $B_i^{gt}$ denote the predicted and ground-truth bounding box coordinates, and $N$ is the number of bounding boxes. 
For cue classification, $C$ is the total number of cue categories, $y_c$ is the one-hot ground-truth label, and $p_c$ is the predicted probability of class $c$.  

This multi-task optimization framework allows the model to learn both coarse-level manipulation detection and fine-grained reasoning cues simultaneously. The binary branch enhances global discrimination capability, the regression branch enforces spatial precision, and the cue classification branch strengthens interpretability by associating detection results with human-understandable manipulation causes.

\section{Model comparison details}\
\label{comparison_more}
\subsection{Details of AI-generated images detection methods}
\begin{itemize}[left=12pt, labelsep=0.6em, labelwidth=0pt]

\item \textbf{Univ \cite{univ}}
introduces a unified representation learning framework for detecting AI-generated images across diverse generative models. It employs multi-scale frequency–spatial fusion and self-supervised domain adaptation, enabling strong generalization to unseen generation architectures (e.g., diffusion, GAN, DiT). Univ achieves robust detection even under post-processing operations such as compression and resizing, though its high complexity can limit real-time deployment.

\item \textbf{AIDE \cite{AIDE}}
 proposes a cross-architecture forensic model based on hybrid vision transformers. By leveraging large-scale datasets of real and synthetic samples, AIDE learns source-invariant traces via contrastive fine-tuning. It provides state-of-the-art detection accuracy for diffusion-based fakes, while maintaining interpretability through attention-based saliency maps. However, its dependency on large-scale pretraining may hinder adaptation to niche domains.

\item \textbf{CNNSpot \cite{cnnspot}}
is a CNN-based forensic detector that focuses on low-level texture and frequency inconsistencies. It is lightweight and efficient, making it suitable for fast fake image screening. It introduces frequency-aware convolution kernels and adaptive pooling, significantly improving robustness to image compression. Despite high sensitivity to pixel artifacts, CNNSpot remains vulnerable to adversarial perturbations and unseen generation pipelines.

\item \textbf{Lagrad \cite{lagrad}}
is a gradient-based forensic model designed to capture subtle distributional shifts between real and AI-generated images. It employs layer-wise gradient regularization and feature contrast alignment, enhancing robustness across generative families. It integrates multi-domain calibration, providing better resilience under cross-model evaluation. Nonetheless, gradient dependence can amplify noise in low-quality or heavily compressed images.

\end{itemize}

\subsection{Details of image manipulation detection methods}
\begin{itemize}[left=12pt, labelsep=0.6em, labelwidth=0pt]

\item \textbf{HiFiNet \cite{hifi}}
introduces a high-fidelity manipulation detection network designed to identify subtle tampering traces in high-resolution images. It leverages hierarchical feature fusion between shallow forensic features and deep semantic representations, enabling accurate localization of manipulated regions. HiFiNet excels at detecting inpainting, splicing, and copy–move edits while maintaining low false-positive rates. However, its high-resolution backbone makes inference computationally expensive for large-scale applications.

\item \textbf{FakeShield \cite{fakeshield}}
proposes a lightweight, transformer-enhanced forensic framework that integrates spatial and frequency cues to detect and localize image manipulations. It adopts a dual-branch architecture combining RGB-domain residual analysis and frequency-domain correlation learning, enabling effective generalization to various editing operations. While robust and efficient, FakeShield’s performance may degrade under severe image compression or adversarial noise.

\item \textbf{MVSSNet \cite{mvss}}
extends the original MVSSNet by incorporating multi-scale attention and structure-aware supervision. It models tampering traces as multi-view correlations, combining pixel-level cues with contextual priors. It introduces an adaptive refinement head that enhances boundary detection of small edited regions. MVSSNet achieves state-of-the-art manipulation localization performance but requires extensive training data for optimal generalization.

\item \textbf{PSCCNet \cite{pscc}}
is a transformer-based manipulation detector that maintains consistency between pixel-level noise patterns and semantic features. It incorporates a correlation-consistency loss to enforce coherence across spatial scales, improving accuracy in detecting subtle photo composites and illumination edits. Despite its precision, PSCCNet demands high computational cost and memory during training.

\end{itemize}

\begin{table*}[t]
\belowrulesep=0pt
\aboverulesep=0pt
\centering
\setlength{\tabcolsep}{8pt} 
\caption{Comparison results on different manipulation category subset. $\spadesuit$ standard CNN/Transformer baselines, $\heartsuit$ image manipulation detection methods, $\clubsuit$ AI-generated image detection methods, $\diamondsuit$ multimodal large language models. The fine-tuned results are marked with \raisebox{0.25ex}{\tiny \ding{91}}. The best results are highlighted in \mredbf{red}, and the second-best results are highlighted in \mbluebf{blue}.}
\resizebox{1\textwidth}{!}
{\begin{tabular}{l||ccccccccccccc}
\hline
\toprule
\noalign{\vspace{1pt}}
Testing Subset& \multicolumn{2}{c}{Addition} & \multicolumn{2}{c}{Removal} & \multicolumn{2}{c}{Replacement} & \multicolumn{2}{c}{Color Modification} & \multicolumn{2}{c}{Action Change} & \multicolumn{2}{c}{Background} \\
\cmidrule(lr){2-3}
\cmidrule(lr){4-5}
\cmidrule(lr){6-7}
\cmidrule(lr){8-9}
\cmidrule(lr){10-11}
\cmidrule(lr){12-13}
\noalign{\vspace{1pt}}
 Model/Metric&Acc$\uparrow$& F1$\uparrow$ &Acc$\uparrow$& F1$\uparrow$ &Acc$\uparrow$& F1$\uparrow$ &Acc$\uparrow$& F1$\uparrow$ &Acc$\uparrow$& F1$\uparrow$ &Acc$\uparrow$& F1$\uparrow$ \\
  \midrule
  \noalign{\vspace{0.5pt}}
  $\heartsuit$HifiNet \cite{hifi} & 53.88 & 35.86 & 49.30 & 40.46 & 52.52 & 42.05 & 53.30 & 44.17 & 55.87 & 56.68 & 49.89 & 35.76 \\
$\heartsuit$FakeShield \cite{fakeshield}& 55.24 & 40.28 & 54.34 & 46.91 & 49.86 & 44.58 & 56.76 & 50.00 & 59.86 & 62.75 & 50.79 & 39.89 \\
\hdashline
\noalign{\vspace{0.5pt}}
$\clubsuit$Univ \cite{univ}& 54.40 & 39.33 & 50.56 & 44.41 & 46.22 & 42.34 & 54.80 & 48.37 & 57.98 & 61.17 & 49.89 & 38.95 \\
$\clubsuit$AIDE \cite{AIDE}& 57.97 & 42.47 & 56.72 & 48.93 & 54.06 & 47.44 & 58.11 & 51.48 & 62.44 & 64.91 & 56.46 & 43.53 \\
\hdashline
\noalign{\vspace{0.5pt}}
$\diamondsuit$LLama3.2-Vision (11B) \cite{llama}& 51.42 & 36.39 & 48.37 & 41.90 & 42.53 & 39.31 & 54.08 & 46.52 & 55.28 & 58.42 & 48.63 & 36.93 \\
$\diamondsuit$LLaVA-1.6 (7B) \cite{llava}& 19.47 & 29.98 & 24.18 & 37.86 & 24.73 & 38.03 & 26.97 & 40.43 & 40.60 & 56.76 & 19.87 & 31.78 \\
$\diamondsuit$MiniCPM-V2.6 (8B) \cite{minicpm}& 41.48 & 34.80 & 37.36 & 40.05 & 34.51 & 38.99 & 44.75 & 44.83 & 47.02 & 57.14 & 35.35 & 34.34 \\
$\diamondsuit$mPLUG-Owl3 (7B) \cite{mplug}& 46.55 & 29.64 & 43.21 & 34.69 & 44.57 & 35.24 & 48.98 & 38.81 & 46.10 & 48.58 & 42.81 & 29.88 \\
$\diamondsuit$CogVLM (17B) \cite{cogvlm}& 80.93 & 28.24 & 76.49 & 29.96 & 74.59 & 28.35 & 74.64 & 29.84 & 63.76 & 31.90 & 81.01 & 29.96 \\
$\diamondsuit$Ovis2.5 (9B) \cite{ovis25}& 44.42 & 38.57 & 41.03 & 44.22 & 39.13 & 43.43 & 48.10 & 49.14 & 59.86 & 66.28 & 43.25 & 39.95 \\
$\diamondsuit$DeepSeekVL2 (small) \cite{deepseekv2} & 19.57 & 30.38 & 25.14 & 38.57 & 24.05 & 38.23 & 26.82 & 40.80 & 41.51 & 57.57 & 19.98 & 32.19 \\
$\diamondsuit$InternVL3 (8B) \cite{internvl3}& 63.49 & 35.71 & 58.29 & 39.45 & 44.57 & 32.89 & 55.69 & 39.68 & 52.75 & 49.26 & 53.79 & 32.21 \\
$\diamondsuit$InternVL3.5 (8B) \cite{internvl3_5}& 77.08 & 43.78 & 64.40 & 40.18 & 55.57 & 34.99 & 68.08 & 44.56 & 63.07 & 52.23 & 69.26 & 38.60 \\
$\diamondsuit$Qwen2.5-VL (8B) \cite{qwenvl2}& 65.82 & 44.30 & 50.27 & 42.27 & 46.33 & 40.42 & 60.64 & 49.81 & 57.80 & 59.29 & 51.04 & 37.54 \\
$\diamondsuit$Qwen3-VL (8B) \cite{qwen3}& 64.88 & 44.99 & 51.40 & 44.12 & 49.30 & 43.08 & 60.66 & 51.12 & 59.86 & 61.57 & 52.15 & 39.37 \\
$\diamondsuit$Gemini2.5-Pro \cite{nanobanana}& 67.61 & 41.59 & 63.03 & 45.45 & 49.30 & 37.80 & 61.56 & 46.22 & 58.22 & 55.28 & 58.05 & 37.29 \\
$\diamondsuit$ChatGPT-4o \cite{chatgpt4o}& 64.47 & 44.70 & 61.62 & 50.00 & 62.32 & 50.46 & 66.37 & 55.02 & 64.55 & 64.47 & 63.72 & 46.13 \\
\hline
\noalign{\vspace{0.5pt}}
$\spadesuit$ResNet50\raisebox{0.25ex}{\tiny \ding{91}} \cite{resnet}& 90.14 & 94.68 & 89.86 & 94.47 & 91.08 & 95.14 & 90.96 & 95.13 & 90.38 & 94.74 & 88.73 & 93.79 \\
$\spadesuit$Swin-T\raisebox{0.25ex}{\tiny \ding{91}} \cite{swin}& 89.90 & 94.49 & 89.86 & 94.43 & 90.38 & 94.74 & 91.35 & 95.31 & 90.77 & 94.96 & 89.26 & 94.09 \\
\hdashline
\noalign{\vspace{0.5pt}}
$\heartsuit$MVSSNet\raisebox{0.25ex}{\tiny \ding{91}} \cite{mvss}& 89.54 & 94.27 & 90.03 & 94.58 & 87.24 & 92.91 & 92.12 & 95.78 & 86.54 & 92.54 & 85.68 & 91.95 \\
$\heartsuit$PSCCNet\raisebox{0.25ex}{\tiny \ding{91}} \cite{pscc}& 90.26 & 94.69 & 90.56 & 94.89 & 89.51 & 94.23 & 91.92 & 95.62 & 88.85 & 93.84 & 87.53 & 93.09 \\
$\heartsuit$HifiNet\raisebox{0.25ex}{\tiny \ding{91}} \cite{hifi} & 90.38 & 94.76 & 90.03 & 94.55 & 88.99 & 93.92 & 90.38 & 94.75 & 89.62 & 94.29 & 88.99 & 93.93 \\
$\heartsuit$FakeShield\raisebox{0.25ex}{\tiny \ding{91}} \cite{fakeshield}& 90.02 & 94.53 & 92.13 & 95.73 & 88.46 & 93.62 & 92.50 & 95.95 & 90.77 & 94.96 & 90.85 & 95.00 \\
\hdashline
\noalign{\vspace{0.5pt}}
$\clubsuit$CNNSpot\raisebox{0.25ex}{\tiny \ding{91}} \cite{cnnspot}& 89.78 & 94.40 & 88.46 & 93.63 & 90.56 & 94.84 & 84.23 & 91.07 & 86.54 & 92.47 & 89.79 & 94.40 \\
$\clubsuit$Lagrad\raisebox{0.25ex}{\tiny \ding{91}} \cite{lagrad}& 90.26 & 94.67 & 91.26 & 95.25 & 91.96 & \mbluebf{95.64} & \mbluebf{92.88} & \mbluebf{96.17} & 91.92 & 95.62 & \mbluebf{92.84} & \mbluebf{96.13} \\
$\clubsuit$Univ\raisebox{0.25ex}{\tiny \ding{91}} \cite{univ}& 91.47 & 95.36 & 91.61 & 95.45 & 90.73 & 94.94 & 91.92 & 95.62 & 90.38 & 94.74 & 91.51 & 95.39 \\
$\clubsuit$AIDE\raisebox{0.25ex}{\tiny \ding{91}} \cite{AIDE}& \mbluebf{92.55} & \mbluebf{95.97} & \mbluebf{93.36} & \mbluebf{96.42} & 91.43 & 95.34 & 91.35 & 95.29 & \mbluebf{92.69} & \mbluebf{96.05} & 92.71 & 96.06 \\
\hdashline
\noalign{\vspace{0.5pt}}
$\diamondsuit$DeepSeekVL2 (small){\tiny \ding{91}} \cite{deepseekv2} & 89.90 & 94.46 & 90.91 & 95.04 & \mbluebf{92.78} & 94.36 & 92.31 & 95.83 & 91.92 & 95.62 & 92.71 & 96.06 \\
$\diamondsuit$InternVL3 (8B){\tiny \ding{91}} \cite{internvl3}& 90.50 & 94.81 & 91.08 & 95.14 & 92.60 & 95.27 & 90.77 & 94.96 & 89.23 & 94.07 & 89.12 & 94.01 \\
$\diamondsuit$Qwen3-VL (8B){\tiny \ding{91}} \cite{qwen3}& 91.35 & 95.29 & 91.08 & 95.14 & 92.43 & 95.18 & 91.35 & 95.29 & 88.85 & 93.84 & 90.98 & 95.08 \\

\hdashline
\noalign{\vspace{0.5pt}}
\rowcolor{gray!20}  
ManipShield (Ours)\raisebox{0.25ex}{\tiny \ding{91}} & \mredbf{92.67} & \mredbf{96.04} & \mredbf{94.41} & \mredbf{97.00} & \mredbf{93.71} & \mredbf{96.62} & \mredbf{94.04} & \mredbf{96.81} & \mredbf{95.38} & \mredbf{97.54} & \mredbf{94.16} & \mredbf{96.87} \\
 \noalign{\vspace{-0.5pt}}
\bottomrule
\end{tabular}}

\vspace{0.5em}
\resizebox{1\textwidth}{!}
{\begin{tabular}{l||ccccccccccccc}
\toprule
\noalign{\vspace{1pt}}
Testing Subset & \multicolumn{2}{c}{Material Change} & \multicolumn{2}{c}{Expression} & \multicolumn{2}{c}{Resizing} & \multicolumn{2}{c}{Scene Text} & \multicolumn{2}{c}{Handwritten Text} & \multicolumn{2}{c}{Scanned Text} \\
\cmidrule(lr){2-3}
\cmidrule(lr){4-5}
\cmidrule(lr){6-7}
\cmidrule(lr){8-9}
\cmidrule(lr){10-11}
\cmidrule(lr){12-13}
\noalign{\vspace{1pt}}
Model/Metric &Acc$\uparrow$& F1$\uparrow$ &Acc$\uparrow$& F1$\uparrow$ &Acc$\uparrow$& F1$\uparrow$ &Acc$\uparrow$& F1$\uparrow$ &Acc$\uparrow$& F1$\uparrow$ &Acc$\uparrow$& F1$\uparrow$ \\
   \midrule
   \noalign{\vspace{0.5pt}}
   $\heartsuit$HifiNet \cite{hifi} & 54.58 & 63.90 & 62.09 & 67.96 & 52.11 & 52.01 & 78.25 & 76.16 & 85.14 & 76.88 & 68.16 & 65.78 \\
$\heartsuit$FakeShield \cite{fakeshield}& 64.38 & 72.54 & 75.49 & 79.34 & 54.01 & 56.92 & 83.62 & 83.24 & 88.55 & 83.48 & 73.13 & 72.73 \\
\hdashline
\noalign{\vspace{0.5pt}}
$\clubsuit$Univ \cite{univ}& 62.09 & 70.85 & 75.82 & 79.21 & 50.21 & 54.44 & 85.59 & 84.68 & 89.96 & 84.94 & 73.88 & 72.87 \\
$\clubsuit$AIDE \cite{AIDE}& 68.63 & 75.51 & 75.82 & 80.00 & 56.75 & 59.08 & 81.36 & 81.77 & 81.12 & 75.90 & 70.90 & 71.67 \\
\hdashline
\noalign{\vspace{0.5pt}}
$\diamondsuit$LLama3.2-Vision (11B) \cite{llama}& 60.13 & 68.84 & 73.31 & 76.75 & 48.35 & 52.19 & 85.87 & 84.31 & 90.41 & 84.83 & 71.05 & 69.72 \\
$\diamondsuit$LLaVA-1.6 (7B) \cite{llava}& 54.66 & 70.69 & 57.23 & 71.88 & 35.80 & 52.15 & 66.48 & 73.75 & 89.43 & 86.29 & 45.26 & 60.18 \\
$\diamondsuit$MiniCPM-V2.6 (8B) \cite{minicpm}& 57.88 & 70.16 & 60.13 & 71.30 & 44.03 & 53.10 & 73.13 & 76.05 & 81.21 & 76.24 & 62.04 & 66.38 \\
$\diamondsuit$mPLUG-Owl3 (7B) \cite{mplug}& 47.27 & 57.51 & 57.23 & 62.54 & 44.65 & 45.21 & 77.29 & 73.03 & 84.54 & 73.75 & 66.67 & 61.84 \\
$\diamondsuit$CogVLM (17B) \cite{cogvlm}& 49.52 & 32.03 & 51.13 & 32.74 & 66.46 & 31.22 & 58.45 & 33.04 & 70.84 & 33.18 & 63.50 & 33.04 \\
$\diamondsuit$Ovis2.5 (9B) \cite{ovis25}& 63.02 & 74.95 & 71.70 & 79.63 & 48.56 & 57.91 & 83.38 & 85.15 & 82.19 & 79.08 & 56.69 & 65.90 \\
$\diamondsuit$DeepSeekVL2 (small) \cite{deepseekv2} & 55.95 & 71.64 & 62.06 & 74.57 & 36.42 & 52.82 & 78.12 & 81.41 & 83.95 & 80.84 & 46.72 & 61.24 \\
$\diamondsuit$InternVL3 (8B) \cite{internvl3}& 49.52 & 56.02 & 65.27 & 64.94 & 43.42 & 42.11 & 76.18 & 69.93 & 83.17 & 69.93 & 77.86 & 68.73 \\
$\diamondsuit$InternVL3.5 (8B) \cite{internvl3_5}& 52.41 & 54.32 & 67.20 & 63.31 & 53.29 & 43.67 & 72.85 & 64.23 & 80.82 & 64.23 & 60.10 & 51.76 \\
Qwen2.5-VL (8B) \cite{qwenvl2}& 56.59 & 66.50 & 72.03 & 75.49 & 50.82 & 52.86 & 85.32 & 83.49 & 89.82 & 83.75 & 67.64 & 66.83 \\
$\diamondsuit$Qwen3-VL (8B) \cite{qwen3}& 58.82 & 68.50 & 72.55 & 76.54 & 51.90 & 54.58 & 83.33 & 82.28 & 87.35 & 81.31 & 68.16 & 68.16 \\
$\diamondsuit$Gemini2.5-Pro \cite{nanobanana}& 55.56 & 61.80 & 68.63 & 69.62 & 50.84 & 48.57 & 78.53 & 74.32 & 84.74 & 74.32 & 80.10 & 73.33 \\
$\diamondsuit$ChatGPT-4o \cite{chatgpt4o}& 66.67 & 72.87 & 70.26 & 75.07 & 60.55 & 59.44 & 83.05 & 82.04 & 88.55 & 82.78 & 77.11 & 74.86 \\
\hline
\noalign{\vspace{0.5pt}}
$\spadesuit$ResNet50\raisebox{0.25ex}{\tiny \ding{91}} \cite{resnet}& 91.54 & 95.40 & 83.08 & 90.60 & 88.14 & 93.43 & 92.31 & 95.95 & 87.28 & 92.92 & 94.87 & 97.37 \\
$\spadesuit$Swin-T\raisebox{0.25ex}{\tiny \ding{91}} \cite{swin}& 88.46 & 93.62 & 85.38 & 92.05 & 89.10 & 93.99 & 92.86 & 96.30 & 95.15 & 97.76 & 94.02 & 96.92 \\
\hdashline
\noalign{\vspace{0.5pt}}
$\heartsuit$MVSSNet\raisebox{0.25ex}{\tiny \ding{91}} \cite{mvss}& 89.23 & 94.07 & 88.46 & 93.83 & 85.90 & 92.09 & 92.86 & 96.30 & 93.20 & 96.43 & 90.17 & 94.74 \\
$\heartsuit$PSCCNet\raisebox{0.25ex}{\tiny \ding{91}} \cite{pscc}& 89.23 & 94.12 & 89.23 & 94.31 & 89.74 & 94.39 & 92.86 & 96.30 & 91.42 & 95.33 & 88.03 & 93.64 \\
$\heartsuit$HifiNet\raisebox{0.25ex}{\tiny \ding{91}} \cite{hifi} & 92.31 & 95.83 & 91.54 & 95.55 & 91.35 & 95.29 & 92.31 & 96.00 & 91.42 & 95.33 & 92.31 & 96.00 \\

$\heartsuit$FakeShield\raisebox{0.25ex}{\tiny \ding{91}} \cite{fakeshield}& 93.85 & 96.69 & 90.00 & 94.74 & \mbluebf{92.95} & \mbluebf{96.19} & 91.76 & 95.70 & 90.24 & 94.72 & 89.32 & 94.33 \\
\hdashline
\noalign{\vspace{0.5pt}}
$\clubsuit$CNNSpot\raisebox{0.25ex}{\tiny \ding{91}} \cite{cnnspot}& 92.31 & 95.87 & 80.00 & 88.79 & 90.71 & 94.92 & 86.26 & 92.58 & 34.91 & 48.84 & 51.28 & 66.27 \\
$\clubsuit$Lagrad\raisebox{0.25ex}{\tiny \ding{91}} \cite{lagrad}& \mbluebf{93.85} & \mbluebf{96.69} & 87.69 & 93.39 & 90.38 & 94.74 & 92.31 & 95.83 & 93.79 & 96.66 & 90.17 & 94.76 \\
$\clubsuit$Univ\raisebox{0.25ex}{\tiny \ding{91}} \cite{univ}& 90.00 & 94.51 & 82.31 & 90.21 & 90.38 & 94.74 & 74.73 & 84.87 & 92.90 & 96.17 & 85.47 & 92.06 \\
$\clubsuit$AIDE\raisebox{0.25ex}{\tiny \ding{91}} \cite{AIDE}& 90.77 & 94.96 & 83.85 & 91.14 & 91.99 & 95.65 & \mbluebf{92.86} & \mbluebf{96.30} & \mbluebf{95.86} & \mbluebf{97.80} & 91.45 & 95.54 \\
\hdashline
\noalign{\vspace{0.5pt}}
$\diamondsuit$DeepSeekVL2 (small){\tiny \ding{91}} \cite{deepseekv2} & 93.08 & 96.27 & 90.08 & 96.27 & 90.38 & 94.74 & 92.86 & 96.14 & 92.90 & 96.17 & 92.74 & 95.07 \\
$\diamondsuit$InternVL3 (8B){\tiny \ding{91}} \cite{internvl3}& 93.08 & 96.27 & \mbluebf{92.15} & \mbluebf{95.96} & 90.06 & 94.55 & 92.15 & 95.96 & 91.42 & 95.33 & \mbluebf{94.97} & \mbluebf{97.46} \\
$\diamondsuit$Qwen3-VL (8B){\tiny \ding{91}} \cite{qwen3}& 91.54 & 95.40 & 91.54 & 95.40 & 91.35 & 95.29 & 92.60 & 95.66 & 95.56 & 97.64 & 93.59 & 96.55 \\
\hdashline
\noalign{\vspace{0.5pt}}
\rowcolor{gray!20}  
ManipShield (Ours)\raisebox{0.25ex}{\tiny \ding{91}} & \mredbf{94.62} & \mredbf{97.12} & \mredbf{92.31} & \mredbf{96.00} & \mredbf{93.59} & \mredbf{96.55} & \mredbf{93.41} & \mredbf{96.59} & \mredbf{97.04} & \mredbf{98.47} & \mredbf{95.30} & \mredbf{97.59} \\
 \noalign{\vspace{-0.5pt}}
\bottomrule
\end{tabular}
}
\label{cat}
\end{table*}

\subsection{Details of multimodal large language models}

\begin{itemize}[left=12pt, labelsep=0.6em, labelwidth=0pt]

\item \textbf{LLaMA-3.2-Vision (11B) \cite{llama}}
is Meta’s open multimodal variant of the LLaMA-3.2 family, integrating a visual encoder with the language backbone for image-text reasoning. It supports high-resolution visual grounding, object reasoning, and instruction following. In the context of fake image identification, its large-scale pretraining on paired vision-language data enables robust detection of subtle editing cues, though its performance can degrade under domain shifts or adversarial manipulations.

\item \textbf{LLaVA-1.6 (7B) \cite{llava}}
extends the LLaMA-2 architecture with a CLIP-based visual encoder aligned via instruction tuning. It achieves strong multimodal understanding, especially for localized reasoning such as identifying inconsistent shadows or textures in edited images. However, its reliance on pre-encoded CLIP embeddings limits fine-grained detection of low-level manipulations.

\item \textbf{MiniCPM-V2.6 (8B) \cite{minicpm}}
is a lightweight multimodal model from OpenBMB, designed for efficient visual-text reasoning with strong low-latency inference. It balances compactness and multimodal fidelity, making it suitable for deployment in real-time image authenticity detection. While efficient, its reduced parameter count can hinder nuanced detection of subtle forgeries.

\item \textbf{mPLUG-Owl3 (7B) \cite{mplug}}
developed by Alibaba’s X-PLUG team, introduces unified vision-language alignment using adaptive visual adapters. It excels in complex cross-modal question answering and reasoning about compositional visual semantics, which benefits fake image detection where multimodal contradictions exist. However, it may struggle with fine-scale pixel-level inconsistencies due to its high-level reasoning focus.

\item \textbf{CogVLM (17B) \cite{cogvlm}}
is a large multimodal transformer that integrates visual tokens into a unified autoregressive architecture. Its massive capacity allows rich contextual reasoning about editing traces (e.g., lighting mismatch, semantic incongruence). Despite superior understanding, inference cost is high, and over-reliance on text priors may reduce pixel-level sensitivity.

\item \textbf{Ovis 2.5 (9B) \cite{ovis25}}
is a vision-language model emphasizing efficient architecture and high-resolution image understanding. It employs hierarchical cross-attention to enhance visual grounding, offering balanced accuracy and speed in identifying tampered or composited regions. Yet, its mid-sized configuration can miss micro-level editing artifacts.

\item \textbf{DeepSeek-VL2 \cite{deepseekvl}}
builds on the DeepSeek LLM series with a multi-stage visual alignment strategy, incorporating both coarse and fine-grained perception. It demonstrates superior generalization across diverse visual domains, performing well in fake image detection involving semantic or stylistic inconsistencies. However, closed-source training data may introduce unknown bias.

\item \textbf{InternVL3 / 3.5 \cite{internvl3,internvl3_5}}
developed by OpenGVLab, advances unified vision-language representation with scalable vision encoders and deep fusion layers. It supports multi-image reasoning, beneficial for comparative fake detection (e.g., before-after editing analysis). Its performance is state-of-the-art, though model size imposes heavy GPU requirements.

\item \textbf{Qwen 2.5-VL / 3-VL \cite{qwenvl2,qwen3}}
is part of Alibaba’s Qwen multimodal series integrating DiT-style vision backbones with instruction-tuned LLMs. It provides strong text-grounded reasoning and fine-grained scene understanding, crucial for identifying forged object insertions. While highly accurate, hallucination from language priors may reduce precision in subtle artifact detection.

\item \textbf{Gemini 2.5-Pro \cite{nanobanana}}
is Google’s closed multimodal foundation model with integrated visual reasoning, OCR, and image understanding capabilities. It shows excellent zero-shot detection of composited or generated content due to broad multimodal pretraining. However, its proprietary nature prevents fine-tuning for domain-specific forensic tasks.

\item \textbf{ChatGPT-4o \cite{chatgpt4o}} integrates high-resolution vision and language reasoning in a unified architecture, supporting both spatial and semantic analysis of manipulated images. Its strong visual understanding enables general detection of synthetic artifacts, though closed weights and lack of forensic specialization limit interpretability and reproducibility.

\end{itemize}
%
\subsection{Details of the human evaluation procedure for explanations}
To evaluate the quality of the generated explanation texts, we conducted a controlled human study. We firstly collect the 16K output explanations from 7 MLLMs and our ManipShield based on testing set ManipBench. For each sample, the evaluators were provided with the original image, the manipulated image, the corresponding editing instruction, and the model-generated explanation, including both the detected manipulated region and the textual reasoning. The evaluators were asked to assess whether the explanation accurately identified the manipulated area, whether the reasoning was logically consistent with the visual evidence, and whether the description was clear and informative.

We recruited five domain experts with prior experience in image analysis and visual content interpretation. All evaluations were performed independently to avoid mutual influence. Each explanation was rated on a 0–5 Likert scale, where higher scores indicate better explanation quality in terms of correctness, consistency, and clarity.

To mitigate individual scoring bias and differences in scoring tendencies among evaluators, we first standardized the scores for each evaluator independently. This step ensures that variations caused by personal scoring scales do not disproportionately affect the aggregated results. After standardization, we performed outlier removal to reduce the influence of accidental misjudgments or annotation noise. For each sample, scores that deviated by more than two standard deviations from the mean of the standardized ratings were regarded as outliers and discarded.

Finally, the retained scores were linearly rescaled to a 0–100 range for better interpretability and reporting consistency. The final performance of the model is reported as the mean opinion score averaged across all evaluated samples. All samples were presented in randomized order during evaluation to prevent order effects. The whole evaluation cost 30 working hours.

\subsection{More comparison results}
In this section, we present additional comparison results and analyses to further demonstrate the performance of different models, as well as key insights into image manipulation detection and localization.

Table~\ref{cat} demonstrates the detection performance of various models across different manipulation tasks. We observe that general detection methods struggle with regional manipulations, such as addition, removal, and replacement, without fine-tuning, likely due to their limited ability to capture subtle local cues. In contrast, these methods perform relatively well on expression changes, as previous training datasets contain abundant facial content. They also achieve higher accuracy on text-related subsets, probably because AIGC-based editing models often generate unrealistic text manipulations. After fine-tuning, performance improves across all categories, while our method consistently surpasses all baseline approaches.

Table~\ref{comparison_model} shows the detection performance of various models across different manipulation methods. We observe that manipulation models based on stable diffusion backbones are generally easier to detect than those based on FLUX or DiT backbones, likely because FLUX- and DiT-based models tend to generate more realistic images. However, some diffusion-based models are also challenging to detect, indicating that detection difficulty depends not only on the backbone but also on the characteristics of the individual model itself. This observation highlights the importance of including a diverse set of manipulation models when constructing evaluation datasets.

Table~\ref{diff_backbone_all} presents the detection accuracy of various models when trained on a subset of manipulation models and evaluated on different subsets. The results indicate that all detection models experience a notable drop in performance when tested on subsets containing backbones unseen during training. This decrease is likely due to the diverse characteristics of images generated by different manipulation models, which makes generalization challenging. In contrast, when trained on the full ManipBench dataset, model performance significantly improves even on previously unseen models such as Flux-Kontext, NanoBanana, GPT-Image, and SeedDream4. These findings highlight the critical importance of including a wide variety of manipulation models in the training dataset to capture diverse visual features and enhance generalization. Furthermore, our model consistently demonstrates the strongest generalization ability among all compared detection methods. This superior performance can be attributed to its design, which not only extracts low-level visual features but also effectively captures high-level semantic inconsistencies introduced by manipulations, enabling it to detect subtle and diverse manipulations across unseen models.

\section{Limitations}
\label{limitation}
\subsection{Cost analysis}
\label{cost_analysis}
A limitation of our approach lies in its relatively higher computational cost compared to traditional image manipulation detection methods. Since our model is built on a large language model backbone, it inevitably requires more GPU memory and incurs higher inference cost than lightweight CNN-based methods such as MVSSNet \cite{mvss} and AIDE \cite{AIDE}. As shown in Table~\ref{cost}, these traditional approaches achieve significantly lower resource consumption due to their compact architectures. However, compared to other explanation-oriented methods like FakeShield \cite{fakeshield}, our model is more efficient in terms of computation. In particular, ManipShield reduces both MACs and inference time while maintaining strong performance due to the designed light weight decoders. Overall, while our method introduces higher cost than conventional detectors, it offers a favorable trade-off between interpretability, generalization ability, and computational efficiency among LLM-based and explanation-capable frameworks.

\section{Potential social impact}
\label{socialimpact}
\subsection{Positive impacts}
ManipBench and ManipShield can substantially strengthen the reliability and interpretability of image manipulation detection in the era of advanced generative models. By covering a wide range of state-of-the-art editing techniques and manipulation categories, the proposed benchmark better reflects real-world synthetic image distributions, enabling more robust model development and evaluation. This can help mitigate the spread of AI-generated misinformation, support digital forensics and media verification in journalism, and enhance the content moderation capabilities of online platforms. Moreover, the inclusion of localized evidence and textual explanations encourages the development of more transparent and human-understandable detection systems, which can improve user trust and facilitate responsible deployment of AI in safety-critical visual media applications.

\subsection{Negative impacts}
At the same time, the release of ManipBench and ManipShield may introduce certain risks associated with dual-use scenarios. Although designed for detection, the benchmark itself reveals detailed manipulation categories and model behaviors, which could potentially be exploited by malicious actors to study weaknesses in current detection systems and create more convincing forged images that evade detection. There is also a risk that downstream users or platforms may place excessive trust in automated predictions, leading to over-reliance on imperfect models and false confidence in content authenticity. In addition, despite its scale, the benchmark may not fully capture all emerging manipulation techniques in the rapidly evolving generative landscape, which could result in biased or incomplete generalization when deployed in uncontrolled real-world environments.

\subsection{Safeguards}
\label{safeguards}
To mitigate these risks, careful release and usage governance are necessary. The dataset and models should ideally be distributed under a research-only license with clear usage restrictions and ethical guidelines to discourage malicious applications. It is also important to accompany the release with transparent documentation of known limitations and failure cases, enabling users to better understand model uncertainty rather than treating outputs as absolute judgments. In high-risk deployment scenarios, the system should be used as a decision-support tool rather than an autonomous authority, ideally combined with human review and complementary provenance-based techniques such as watermarking or metadata verification. Continuous evaluation against adaptive and adversarial manipulation methods should also be encouraged to improve robustness over time and reduce potential exploitation.

\begin{table*}[t]
\belowrulesep=0pt
\aboverulesep=0pt
\centering
\setlength{\tabcolsep}{8pt} 
\caption{Comparison results on different manipulation method subset. $\spadesuit$ standard CNN/Transformer baselines, $\heartsuit$ image manipulation detection methods, $\clubsuit$ AI-generated image detection methods, $\diamondsuit$ multimodal large language
models. The fine-tuned results are marked with \raisebox{0.25ex}{\tiny \ding{91}}. The best results are highlighted in \mredbf{red}, and the second-best results are highlighted in \mbluebf{blue}.}
\resizebox{1\textwidth}{!}{\begin{tabular}{l||ccccccccccccccc}
\toprule
\noalign{\vspace{0.5pt}}
 & \multicolumn{2}{c}{IP2P \cite{ip2p}} & \multicolumn{2}{c}{CDS \cite{CDS}} & \multicolumn{2}{c}{MagicBrush \cite{Magicbrush}} & \multicolumn{2}{c}{InfEdit \cite{InfEdit}} & \multicolumn{2}{c}{MAG \cite{MAG}} & \multicolumn{2}{c}{HQEdit \cite{HQ}} & \multicolumn{2}{c}{PnP \cite{PnP}} \\
\noalign{\vspace{-1pt}}
\cmidrule(lr){2-3}
\cmidrule(lr){4-5}
\cmidrule(lr){6-7}
\cmidrule(lr){8-9}
\cmidrule(lr){10-11}
\cmidrule(lr){12-13}
\cmidrule(lr){14-15}
\noalign{\vspace{0.5pt}}
 Model&Acc$\uparrow$& F1$\uparrow$ &Acc$\uparrow$& F1$\uparrow$ &Acc$\uparrow$& F1$\uparrow$ &Acc$\uparrow$& F1$\uparrow$ &Acc$\uparrow$& F1$\uparrow$ &Acc$\uparrow$& F1$\uparrow$ &Acc$\uparrow$& F1$\uparrow$ \\
\midrule

$\diamondsuit$LLama3.2-Vision (11B) \cite{llama}& 77.42 & 76.54 & 65.32 & 67.99 & 70.16 & 71.17 & 67.47 & 69.37 & 63.98 & 67.16 & 83.87 & 82.04 & 63.98 & 67.16 \\
$\diamondsuit$LLaVA-1.6 (7B) \cite{llava}& 49.73 & 64.52 & 50.54 & 64.89 & 50.54 & 64.89 & 50.81 & 65.01 & 50.27 & 64.76 & 56.72 & 67.86 & 50.27 & 64.76 \\
$\diamondsuit$MiniCPM-V2.6 (8B) \cite{minicpm}& 68.55 & 72.47 & 50.27 & 62.47 & 58.60 & 66.67 & 55.91 & 65.25 & 59.14 & 66.96 & 73.66 & 75.86 & 62.37 & 68.75 \\
$\diamondsuit$mPLUG-Owl3 (7B) \cite{mplug}& 62.90 & 61.67 & 52.69 & 55.78 & 55.91 & 57.51 & 53.76 & 56.35 & 56.18 & 57.66 & 67.47 & 64.72 & 52.69 & 55.78 \\
$\diamondsuit$CogVLM (17B) \cite{cogvlm}& 59.95 & 33.18 & 59.41 & 32.89 & 59.41 & 32.89 & 56.99 & 31.62 & 58.60 & 32.46 & 59.95 & 33.18 & 58.60 & 32.46 \\
$\diamondsuit$Ovis2.5 (9B) \cite{ovis25}& 79.30 & 81.71 & 56.45 & 67.98 & 65.05 & 72.57 & 58.60 & 69.08 & 68.82 & 74.78 & 89.52 & 89.82 & 67.47 & 73.98 \\
$\diamondsuit$DeepSeekVL2 (small) \cite{deepseekv2} & 55.11 & 67.45 & 51.08 & 65.53 & 52.96 & 66.41 & 50.54 & 65.28 & 51.88 & 65.90 & 56.72 & 68.24 & 51.88 & 65.90 \\
$\diamondsuit$InternVL3 (8B) \cite{internvl3}& 70.70 & 64.72 & 55.91 & 54.95 & 62.90 & 59.17 & 59.95 & 57.31 & 60.48 & 57.64 & 75.00 & 68.26 & 60.48 & 57.64 \\
$\diamondsuit$InternVL3.5 (8B) \cite{internvl3_5}& 76.08 & 75.48 & 60.75 & 65.24 & 69.89 & 70.98 & 66.67 & 68.84 & 64.52 & 67.49 & 81.45 & 79.88 & 65.05 & 67.82 \\
$\diamondsuit$Qwen2.5-VL (8B) \cite{qwenvl2}& 70.97 & 61.97 & 63.98 & 56.77 & 68.28 & 59.86 & 65.32 & 57.70 & 66.13 & 58.28 & 72.85 & 63.54 & 66.13 & 58.28 \\
$\diamondsuit$Qwen3-VL (8B) \cite{qwen3}& 76.61 & 75.49 & 59.14 & 63.81 & 68.28 & 69.43 & 66.13 & 68.02 & 65.05 & 67.34 & 81.99 & 80.00 & 65.05 & 67.34 \\
$\diamondsuit$Gemini2.5-Pro \cite{nanobanana}& 73.66 & 69.18 & 60.48 & 59.95 & 67.47 & 64.52 & 64.78 & 62.68 & 64.52 & 62.50 & 77.69 & 72.61 & 65.59 & 63.22 \\
$\diamondsuit$ChatGPT-4o \cite{chatgpt4o}& 74.73 & 74.46 & 69.89 & 70.98 & 68.55 & 70.08 & 68.01 & 69.72 & 69.89 & 70.98 & 76.34 & 75.69 & 69.35 & 70.62 \\
\hline
\noalign{\vspace{0.5pt}}
$\spadesuit$ResNet50{\tiny \ding{91}} \cite{resnet}& 88.98 & 80.15 & 88.17 & 80.00 & 88.98 & 80.15 & 88.98 & 80.15 & 88.98 & 80.15 & 88.71 & 80.10 & 88.98 & 80.15 \\
$\spadesuit$Swin-T{\tiny \ding{91}} \cite{swin}& 90.59 & 82.80 & 90.59 & 82.80 & 90.59 & 82.80 & 90.59 & 82.80 & 90.59 & 82.80 & 90.59 & 82.80 & 90.59 & 82.80 \\
\hdashline
\noalign{\vspace{0.5pt}}
$\heartsuit$MVSSNet{\tiny \ding{91}} \cite{mvss}& 84.68 & 77.78 & 87.90 & 78.42 & 85.75 & 78.00 & 87.90 & 78.42 & 87.90 & 78.42 & 86.83 & 78.21 & 87.90 & 78.42 \\
$\heartsuit$PSCCNet{\tiny \ding{91}} \cite{pscc}& 90.59 & 83.62 & 91.13 & 83.70 & 91.13 & 83.70 & 91.13 & 83.70 & 91.13 & 83.70 & 90.86 & 83.66 & 91.13 & 83.70 \\
$\heartsuit$HifiNet{\tiny \ding{91}} \cite{hifi} & 90.59 & 87.08 & 93.28 & 87.41 & 91.40 & 87.18 & 93.28 & 87.41 & 93.01 & 87.37 & 92.47 & 87.31 & 93.01 & 87.37 \\
$\heartsuit$FakeShield{\tiny \ding{91}} \cite{fakeshield}& 92.74 & 87.34 & 93.28 & 87.41 & 92.47 & 87.31 & 93.28 & 87.41 & 93.28 & 87.41 & 92.74 & 87.34 & 93.28 & 87.41 \\
\hline
\noalign{\vspace{0.5pt}}
$\clubsuit$CNNSpot{\tiny \ding{91}} \cite{cnnspot}& 90.59 & 91.83 & 90.59 & 91.83 & 89.52 & 91.74 & 93.28 & 92.04 & 91.13 & 91.87 & 95.16 & 92.19 & 91.13 & 91.87 \\
$\clubsuit$Lagrad{\tiny \ding{91}} \cite{lagrad}& 94.89 & 90.28 & 94.89 & 90.28 & 94.89 & 90.28 & 94.35 & 90.23 & 94.89 & 90.28 & 94.62 & 90.26 & 94.89 & 90.28 \\
$\clubsuit$Univ{\tiny \ding{91}} \cite{univ}& 92.20 & 87.72 & 93.55 & 87.88 & 93.01 & 87.82 & 93.55 & 87.88 & 93.28 & 87.85 & 93.01 & 87.82 & 93.28 & 87.85 \\
$\clubsuit$AIDE{\tiny \ding{91}} \cite{AIDE}& \mbluebf{95.70} & \mbluebf{94.18} & \mbluebf{95.43} & \mbluebf{94.16} & \mbluebf{96.24} & \mbluebf{94.21} & \mbluebf{95.43} & \mbluebf{94.16} & \mbluebf{96.51} & \mbluebf{94.23} & \mbluebf{96.51} & \mbluebf{94.23} & \mbluebf{96.51} & \mbluebf{94.23} \\
\hdashline
\noalign{\vspace{0.5pt}}
 
\rowcolor{gray!20}  
ManipShield (Ours){\tiny \ding{91}} & \mredbf{97.31} & \mredbf{95.26} & \mredbf{97.04} & \mredbf{95.25} & \mredbf{97.31} & \mredbf{95.26} & \mredbf{97.04} & \mredbf{95.25} & \mredbf{97.58} & \mredbf{95.28} & \mredbf{97.31} & \mredbf{95.26} & \mredbf{97.58} & \mredbf{95.28} \\
\bottomrule
\end{tabular}}
\vspace{0.5em}
\resizebox{1\textwidth}{!}{\begin{tabular}{l||ccccccccccccccc}
\toprule
\noalign{\vspace{0.5pt}}
 & \multicolumn{2}{c}{ZONE \cite{zone}} & \multicolumn{2}{c}{PowerPaint \cite{PPT}} & \multicolumn{2}{c}{BrushNet \cite{brushnet}} & \multicolumn{2}{c}{Any2Pix \cite{instructany2pix}} & \multicolumn{2}{c}{ReNoise \cite{renoise}} & \multicolumn{2}{c}{FlowEdit(SD3) \cite{Flowedit}} & \multicolumn{2}{c}{RFSE \cite{RFSE}} \\
\noalign{\vspace{-1pt}}
\cmidrule(lr){2-3}
\cmidrule(lr){4-5}
\cmidrule(lr){6-7}
\cmidrule(lr){8-9}
\cmidrule(lr){10-11}
\cmidrule(lr){12-13}
\cmidrule(lr){14-15}
\noalign{\vspace{0.5pt}}
Model &Acc$\uparrow$& F1$\uparrow$ &Acc$\uparrow$& F1$\uparrow$ &Acc$\uparrow$& F1$\uparrow$ &Acc$\uparrow$& F1$\uparrow$ &Acc$\uparrow$& F1$\uparrow$ &Acc$\uparrow$& F1$\uparrow$ &Acc$\uparrow$& F1$\uparrow$ \\
 \midrule

$\diamondsuit$LLama3.2-Vision (11B) \cite{llama}& 68.55 & 70.08 & 63.44 & 66.83 & 63.44 & 66.83 & 61.83 & 65.87 & 65.32 & 67.99 & 56.18 & 62.70 & 54.03 & 61.57 \\
$\diamondsuit$LLaVA-1.6 (7B) \cite{llava}& 50.54 & 64.89 & 52.69 & 65.89 & 50.81 & 65.01 & 49.73 & 64.52 & 55.91 & 67.46 & 49.73 & 64.52 & 50.00 & 64.64 \\
$\diamondsuit$MiniCPM-V2.6 (8B) \cite{minicpm}& 59.41 & 67.10 & 56.45 & 65.53 & 55.38 & 64.98 & 59.41 & 67.10 & 59.95 & 67.40 & 52.42 & 63.51 & 52.96 & 63.77 \\
$\diamondsuit$mPLUG-Owl3 (7B) \cite{mplug}& 53.23 & 56.06 & 56.72 & 57.96 & 54.84 & 56.92 & 53.23 & 56.06 & 57.53 & 58.42 & 51.88 & 55.36 & 51.34 & 55.09 \\
$\diamondsuit$CogVLM (17B) \cite{cogvlm}& 59.68 & 33.04 & 59.41 & 32.89 & 58.87 & 32.60 & 57.53 & 31.90 & 59.68 & 33.04 & 59.41 & 32.89 & 57.80 & 32.03 \\
$\diamondsuit$Ovis2.5 (9B) \cite{ovis25}& 65.32 & 72.73 & 58.33 & 68.94 & 66.67 & 73.50 & 72.31 & 76.96 & 63.17 & 71.52 & 55.11 & 67.32 & 59.95 & 69.78 \\
$\diamondsuit$DeepSeekVL2 (small) \cite{deepseekv2} & 52.69 & 66.28 & 51.34 & 65.65 & 51.34 & 65.65 & 51.61 & 65.78 & 52.15 & 66.03 & 52.96 & 66.41 & 52.15 & 66.03 \\
$\diamondsuit$InternVL3 (8B) \cite{internvl3}& 61.56 & 58.31 & 55.38 & 54.64 & 59.68 & 57.14 & 62.37 & 58.82 & 61.83 & 58.48 & 53.49 & 53.62 & 48.66 & 51.15 \\
$\diamondsuit$InternVL3.5 (8B) \cite{internvl3_5}& 67.20 & 69.19 & 62.63 & 66.34 & 64.25 & 67.32 & 68.55 & 70.08 & 65.05 & 67.82 & 57.53 & 63.43 & 59.68 & 64.62 \\
$\diamondsuit$Qwen2.5-VL (8B) \cite{qwenvl2}& 64.78 & 57.33 & 63.17 & 56.23 & 64.25 & 56.96 & 62.10 & 55.52 & 66.13 & 58.28 & 59.68 & 53.99 & 59.14 & 53.66 \\
$\diamondsuit$Qwen3-VL (8B) \cite{qwen3}& 66.94 & 68.54 & 62.90 & 66.01 & 64.25 & 66.83 & 68.28 & 69.43 & 66.13 & 68.02 & 56.18 & 62.18 & 57.26 & 62.76 \\
$\diamondsuit$Gemini2.5-Pro \cite{nanobanana}& 65.86 & 63.40 & 61.29 & 60.44 & 64.52 & 62.50 & 66.40 & 63.77 & 65.05 & 62.86 & 58.33 & 58.67 & 55.65 & 57.14 \\
$\diamondsuit$ChatGPT-4o \cite{chatgpt4o}& 70.97 & 71.73 & 72.04 & 72.49 & 72.04 & 72.49 & 67.74 & 69.54 & 69.89 & 70.98 & 70.16 & 71.17 & 68.55 & 70.08 \\
\hline
\noalign{\vspace{0.5pt}}
$\spadesuit$ResNet50{\tiny \ding{91}} \cite{resnet}& 88.98 & 80.15 & 88.98 & 80.15 & 88.98 & 80.15 & 88.71 & 80.10 & 88.98 & 80.15 & 88.71 & 80.10 & 87.63 & 79.90 \\
$\spadesuit$Swin-T{\tiny \ding{91}} \cite{swin}& 90.59 & 82.80 & 90.59 & 82.80 & 90.32 & 82.76 & 90.59 & 82.80 & 90.59 & 82.80 & 89.78 & 82.67 & 90.32 & 82.76 \\
\hdashline
\noalign{\vspace{0.5pt}}
$\heartsuit$MVSSNet{\tiny \ding{91}} \cite{mvss}& 87.90 & 78.42 & 87.90 & 78.42 & 87.10 & 78.26 & 87.63 & 78.37 & 87.90 & 78.42 & 86.02 & 78.05 & 87.90 & 78.42 \\
 $\heartsuit$PSCCNet{\tiny \ding{91}} \cite{pscc}& 91.13 & 83.70 & 91.13 & 83.70 & 90.86 & 83.66 & 90.86 & 83.66 & 91.13 & 83.70 & 91.13 & 83.70 & 91.13 & 83.70 \\
 $\heartsuit$HifiNet{\tiny \ding{91}} \cite{hifi} & 93.28 & 87.41 & 93.01 & 87.37 & 93.01 & 87.37 & 92.20 & 87.28 & 93.28 & 87.41 & 92.20 & 87.28 & 93.01 & 87.37 \\
 $\heartsuit$FakeShield{\tiny \ding{91}} \cite{fakeshield}& 93.01 & 87.37 & 93.01 & 87.37 & 92.74 & 87.34 & 92.47 & 87.31 & 93.28 & 87.41 & 92.74 & 87.34 & 93.01 & 87.37 \\
\hline
\noalign{\vspace{0.5pt}}
$\clubsuit$CNNSpot{\tiny \ding{91}} \cite{cnnspot}& 91.94 & 91.94 & 91.13 & 91.87 & 89.25 & 91.71 & 90.32 & 91.80 & 93.82 & 92.08 & 90.05 & 91.78 & 90.59 & 91.83 \\
$\clubsuit$Lagrad{\tiny \ding{91}} \cite{lagrad}& 94.62 & 90.26 & 94.89 & 90.28 & 94.89 & 90.28 & 94.62 & 90.26 & 94.89 & 90.28 & 94.89 & 90.28 & 94.62 & 90.26 \\
$\clubsuit$Univ{\tiny \ding{91}} \cite{univ}& 93.55 & 87.88 & 93.28 & 87.85 & 93.28 & 87.85 & 93.28 & 87.85 & 93.55 & 87.88 & 93.01 & 87.82 & 93.28 & 87.85 \\
$\clubsuit$AIDE{\tiny \ding{91}} \cite{AIDE}& \mbluebf{95.43} & \mbluebf{94.16} & \mbluebf{96.51} & \mbluebf{94.23} & \mbluebf{96.24} & \mbluebf{94.21} & \mbluebf{96.51} & \mbluebf{94.23} & \mbluebf{95.97} & \mbluebf{94.20} & \mbluebf{94.62} & \mbluebf{94.12} & \mbluebf{96.51} & \mbluebf{94.23} \\
\hdashline
\noalign{\vspace{0.5pt}}
 
\rowcolor{gray!20}  
ManipShield (Ours){\tiny \ding{91}} & \mredbf{97.58} & \mredbf{95.28} & \mredbf{97.58} & \mredbf{95.28} & \mredbf{97.31} & \mredbf{95.26} & \mredbf{97.31} & \mredbf{95.26} & \mredbf{97.04} & \mredbf{95.25} & \mredbf{97.31} & \mredbf{95.26} & \mredbf{97.31} & \mredbf{95.26} \\
\bottomrule
\end{tabular}}
\vspace{0.5em}
\resizebox{1\textwidth}{!}
{\begin{tabular}{l||ccccccccccccccc}
\toprule
\noalign{\vspace{0.5pt}}
& \multicolumn{2}{c}{FlowEdit(Flux) \cite{Flowedit}} & \multicolumn{2}{c}{ACE \cite{ACE}} & \multicolumn{2}{c}{Reflex \cite{reflex}} & \multicolumn{2}{c}{OT \cite{OT}} & \multicolumn{2}{c}{Follow-Your-Shape \cite{followyourshape}} & \multicolumn{2}{c}{OmniGen2 \cite{omnigen2}} & \multicolumn{2}{c}{Qwen-Image-Edit \cite{qwenedit}} \\
\noalign{\vspace{-1pt}}
\cmidrule(lr){2-3}
\cmidrule(lr){4-5}
\cmidrule(lr){6-7}
\cmidrule(lr){8-9}
\cmidrule(lr){10-11}
\cmidrule(lr){12-13}
\cmidrule(lr){14-15}
\noalign{\vspace{0.5pt}}
 Model&Acc$\uparrow$& F1$\uparrow$ &Acc$\uparrow$& F1$\uparrow$ &Acc$\uparrow$& F1$\uparrow$ &Acc$\uparrow$& F1$\uparrow$ &Acc$\uparrow$& F1$\uparrow$ &Acc$\uparrow$& F1$\uparrow$ &Acc$\uparrow$& F1$\uparrow$ \\
  \midrule

$\diamondsuit$LLama3.2-Vision (11B) \cite{llama}& 50.54 & 59.83 & 66.13 & 68.50 & 62.90 & 66.50 & 55.91 & 62.56 & 54.57 & 61.85 & 52.42 & 60.75 & 59.14 & 64.32 \\
$\diamondsuit$LLaVA-1.6 (7B) \cite{llava}& 48.92 & 64.15 & 51.61 & 65.38 & 50.54 & 64.89 & 49.73 & 64.52 & 50.27 & 64.76 & 49.19 & 64.27 & 49.73 & 64.52 \\
$\diamondsuit$MiniCPM-V2.6 (8B) \cite{minicpm}& 49.73 & 62.22 & 57.80 & 66.24 & 59.95 & 67.40 & 54.84 & 64.71 & 52.96 & 63.77 & 52.15 & 63.37 & 58.33 & 66.52 \\
$\diamondsuit$mPLUG-Owl3 (7B) \cite{mplug}& 48.92 & 53.88 & 56.45 & 57.81 & 53.76 & 56.35 & 50.81 & 54.81 & 49.46 & 54.15 & 48.66 & 53.75 & 54.30 & 56.63 \\
$\diamondsuit$CogVLM (17B) \cite{cogvlm}& 58.33 & 32.31 & 58.87 & 32.60 & 59.41 & 32.89 & 59.41 & 32.89 & 58.87 & 32.60 & 57.53 & 31.90 & 55.65 & 30.96 \\
$\diamondsuit$Ovis2.5 (9B) \cite{ovis25}& 52.15 & 65.90 & 64.52 & 72.27 & 71.77 & 76.61 & 62.37 & 71.07 & 56.18 & 67.85 & 56.45 & 67.98 & 61.83 & 70.78 \\
$\diamondsuit$DeepSeekVL2 (small) \cite{deepseekv2} & 50.81 & 65.41 & 54.30 & 67.05 & 52.15 & 66.03 & 50.81 & 65.41 & 52.15 & 66.03 & 48.92 & 64.55 & 51.61 & 65.78 \\
$\diamondsuit$InternVL3 (8B) \cite{internvl3}& 46.77 & 50.25 & 60.75 & 57.80 & 55.91 & 54.95 & 50.00 & 51.81 & 50.81 & 52.22 & 55.38 & 54.64 & 61.02 & 57.97 \\
$\diamondsuit$InternVL3.5 (8B) \cite{internvl3_5}& 53.23 & 61.16 & 63.98 & 67.16 & 67.47 & 69.37 & 59.95 & 64.78 & 56.18 & 62.70 & 62.37 & 66.18 & 72.85 & 73.07 \\
$\diamondsuit$Qwen2.5-VL (8B) \cite{qwenvl2}& 54.30 & 50.87 & 66.40 & 58.47 & 63.98 & 56.77 & 60.22 & 54.32 & 57.26 & 52.54 & 56.99 & 52.38 & 53.49 & 50.43 \\
$\diamondsuit$Qwen3-VL (8B) \cite{qwen3}& 51.08 & 59.56 & 65.05 & 67.34 & 66.67 & 68.37 & 58.33 & 63.36 & 54.57 & 61.33 & 62.10 & 65.53 & 74.46 & 73.83 \\
$\diamondsuit$Gemini2.5-Pro \cite{nanobanana}& 52.15 & 55.28 & 65.32 & 63.04 & 60.22 & 59.78 & 56.18 & 57.44 & 55.91 & 57.29 & 59.41 & 59.30 & 64.52 & 62.50 \\
$\diamondsuit$ChatGPT-4o \cite{chatgpt4o}& 66.13 & 68.50 & 69.09 & 70.44 & 66.94 & 69.02 & 68.28 & 69.90 & 65.32 & 67.99 & 62.90 & 66.50 & 70.43 & 71.35 \\
\hline
\noalign{\vspace{0.5pt}}
$\spadesuit$ResNet50{\tiny \ding{91}} \cite{resnet}& 88.71 & 80.10 & 88.98 & 80.15 & 88.71 & 80.10 & 88.98 & 80.15 & 88.71 & 80.10 & 87.37 & 79.85 & 85.75 & 79.55 \\
$\spadesuit$Swin-T{\tiny \ding{91}} \cite{swin}& 90.59 & 82.80 & 90.59 & 82.80 & 89.25 & 82.59 & 90.59 & 82.80 & 89.25 & 82.59 & 90.05 & 82.72 & 88.98 & 82.54 \\
\hdashline
\noalign{\vspace{0.5pt}}
 $\heartsuit$MVSSNet{\tiny \ding{91}} \cite{mvss}& 87.63 & 78.37 & 87.90 & 78.42 & 87.90 & 78.42 & 87.63 & 78.37 & 87.10 & 78.26 & 81.72 & 77.16 & 83.87 & 77.61 \\
 $\heartsuit$PSCCNet{\tiny \ding{91}} \cite{pscc}& 91.13 & 83.70 & 91.13 & 83.70 & 91.13 & 83.70 & 91.13 & 83.70 & 90.86 & 83.66 & 90.05 & 83.54 & 88.17 & 83.25 \\
 $\heartsuit$HifiNet{\tiny \ding{91}} \cite{hifi} & 93.01 & 87.37 & 93.28 & 87.41 & 93.01 & 87.37 & 93.01 & 87.37 & 91.94 & 87.24 & 89.52 & 86.95 & 89.78 & 86.98 \\
 $\heartsuit$FakeShield{\tiny \ding{91}} \cite{fakeshield}& 93.28 & 87.41 & 93.28 & 87.41 & 93.28 & 87.41 & 93.01 & 87.37 & 92.74 & 87.34 & 88.98 & 86.88 & 89.78 & 86.98 \\
\hline
\noalign{\vspace{0.5pt}}
$\clubsuit$CNNSpot{\tiny \ding{91}} \cite{cnnspot}& 90.86 & 91.85 & 95.16 & 92.19 & 91.67 & 91.91 & 89.52 & 91.74 & 89.78 & 91.76 & 86.29 & 91.45 & 85.75 & 91.40 \\
$\clubsuit$Lagrad{\tiny \ding{91}} \cite{lagrad}& 94.89 & 90.28 & 94.89 & 90.28 & 94.89 & 90.28 & 94.89 & 90.28 & 94.89 & 90.28 & 92.20 & 90.03 & 93.01 & 90.10 \\
$\clubsuit$Univ{\tiny \ding{91}} \cite{univ}& 93.55 & 87.88 & 93.55 & 87.88 & 93.55 & 87.88 & 93.55 & 87.88 & 93.55 & 87.88 & 92.47 & 87.76 & 91.94 & 87.69 \\
$\clubsuit$AIDE{\tiny \ding{91}} \cite{AIDE}& \mbluebf{95.70} & \mbluebf{94.18} & \mbluebf{96.24} & \mbluebf{94.21} & \mbluebf{96.77} & \mbluebf{94.24} & \mbluebf{95.97} & \mbluebf{94.20} & \mbluebf{94.89} & \mbluebf{94.13} & \mbluebf{95.16} & \mbluebf{94.15} & \mbluebf{92.47} & \mbluebf{93.99} \\
\hdashline
\noalign{\vspace{0.5pt}}

\rowcolor{gray!20}  
ManipShield (Ours){\tiny \ding{91}} & \mredbf{97.58} & \mredbf{95.28} & \mredbf{97.04} & \mredbf{95.25} & \mredbf{97.58} & \mredbf{95.28} & \mredbf{97.31} & \mredbf{95.26} & \mredbf{97.31} & \mredbf{95.26} & \mredbf{97.04} & \mredbf{95.25} & \mredbf{96.77} & \mredbf{95.24} \\
\bottomrule
\end{tabular}}
\label{comparison_model}
\end{table*}

\begin{table*}[t]
\belowrulesep=0pt
\aboverulesep=0pt
\centering
\caption{ Accuracy results (Acc$\uparrow$) of cross-validation on different training and testing subsets.}
\resizebox{1\textwidth}{!}{\begin{tabular}{l||cccccc||cccc||c}
\toprule
\noalign{\vspace{1pt}}
ResNet50 \cite{resnet} & \multicolumn{10}{c}{Testing Subset} \\
\cmidrule(lr){2-11}
\noalign{\vspace{1.5pt}}
Training Subset& SD1.4 & SD1.5 & SD3 & SDXL & FLUX & DiT & FLUX-Kontext & NanoBanana & GPT-Image &  SeedDream4 &Avg\\
\midrule
\noalign{\vspace{0.5pt}}
SD1.4 & \textbf{93.19} & 85.75 & 72.04 & 86.69 & 68.77 & 62.50 & 54.57 & 58.06 & 60.48 & 62.10 & 70.42 \\
SD1.5 & 88.84 & \textbf{92.11} & 66.40 & 70.16 & 70.39 & 64.25 & 58.60 & 59.95 & 58.06 & 64.52 & 69.33 \\
SD3 & 76.39 & 72.31 & \textbf{91.67} & 71.10 & 62.99 & 61.02 & 60.75 & 56.45 & 58.60 & 63.44 & 67.47 \\
SDXL & 77.24 & 68.28 & 60.22 & \textbf{92.47} & 63.98 & 63.31 & 56.99 & 58.60 & 59.41 & 60.48 & 66.10 \\
FLUX & 71.06 & 74.73 & 70.97 & 77.42 & \textbf{93.41} & 77.15 & 65.05 & 64.78 & 55.91 & 66.40 & 71.69 \\
DiT & 69.80 & 66.13 & 66.13 & 73.66 & 71.33 & \textbf{92.61} & 63.44 & 63.98 & 58.33 & 68.28 & 69.37 \\
\hdashline
\noalign{\vspace{0.5pt}}
Full Dataset & 88.80 & 88.98 & 88.84 & 88.71 & 88.62 &  86.56  & 65.32&58.06&56.99&68.01&84.33  \\
\noalign{\vspace{-0.5pt}}
\bottomrule
\end{tabular}}

\resizebox{1\textwidth}{!}{\begin{tabular}{l||cccccc||cccc||c}
\toprule
\noalign{\vspace{1pt}}
Swin-T \cite{swin} & \multicolumn{10}{c}{Testing Subset} \\
\cmidrule(lr){2-11}
\noalign{\vspace{1.5pt}}
Training Subset& SD1.4 & SD1.5 & SD3 & SDXL & FLUX & DiT & FLUX-Kontext & NanoBanana & GPT-Image &  SeedDream4 &Avg\\
\midrule
\noalign{\vspace{0.5pt}}
SD1.4 & \textbf{90.37} & 87.46 & 73.39 & 81.45 & 63.40 & 65.19 & 57.80 & 57.53 & 61.56 & 66.40 & 70.45 \\
SD1.5 & 80.02 & \textbf{90.95} & 65.59 & 69.49 & 66.08 & 61.16 & 53.76 & 65.05 & 52.42 & 65.05 & 66.96 \\
SD3 & 78.99 & 74.46 & \textbf{92.20} & 70.03 & 64.20 & 57.26 & 56.72 & 57.26 & 60.75 & 63.71 & 67.56 \\
SDXL & 70.47 & 73.92 & 62.37 & \textbf{92.20} & 61.56 & 69.09 & 52.42 & 63.71 & 55.38 & 56.45 & 65.76 \\
FLUX & 74.06 & 79.57 & 68.28 & 81.59 & \textbf{92.43} & 81.18 & 66.13 & 62.10 & 55.38 & 65.05 & 72.58 \\
DiT & 74.10 & 67.74 & 67.20 & 69.76 & 71.42 & \textbf{91.13} & 62.10 & 67.47 & 64.78 & 61.56 & 69.73 \\
\noalign{\vspace{-0.5pt}}
\bottomrule
\end{tabular}}

\resizebox{1\textwidth}{!}{\begin{tabular}{l||cccccc||cccc||c}
\toprule
\noalign{\vspace{1pt}}
CNNSpot \cite{cnnspot} & \multicolumn{10}{c}{Testing Subset} \\
\cmidrule(lr){2-11}
\noalign{\vspace{1.5pt}}
Training Subset& SD1.4 & SD1.5 & SD3 & SDXL & FLUX & DiT & FLUX-Kontext & NanoBanana & GPT-Image &  SeedDream4 &Avg\\
\midrule
\noalign{\vspace{0.5pt}}
SD1.4 & \textbf{95.21} & 77.42 & 67.20 & 78.36 & 61.78 & 64.11 & 53.49 & 57.53 & 51.88 & 58.60 & 66.56 \\
SD1.5 & 85.93 & \textbf{92.20} & 62.10 & 70.03 & 67.03 & 66.53 & 50.00 & 52.15 & 61.29 & 69.62 & 67.69 \\
SD3 & 66.53 & 69.35 & \textbf{95.43} & 70.97 & 69.22 & 60.62 & 61.29 & 52.69 & 61.56 & 67.20 & 67.49 \\
SDXL & 79.35 & 61.11 & 62.37 & \textbf{93.82} & 68.06 & 65.73 & 55.91 & 56.45 & 58.33 & 61.83 & 66.29 \\
FLUX & 62.01 & 67.11 & 72.58 & 71.64 & \textbf{95.83} & 81.72 & 69.09 & 63.17 & 55.65 & 58.60 & 69.74 \\
DiT & 69.40 & 66.31 & 68.28 & 66.13 & 65.77 & \textbf{92.20} & 57.26 & 56.99 & 58.60 & 70.16 & 67.11 \\
\noalign{\vspace{-0.5pt}}
\bottomrule
\end{tabular}}

\resizebox{1\textwidth}{!}{\begin{tabular}{l||cccccc||cccc||c}
\toprule
\noalign{\vspace{1pt}}
AIDE \cite{AIDE} & \multicolumn{10}{c}{Testing Subset} \\
\cmidrule(lr){2-11}
\noalign{\vspace{1.5pt}}
Training Subset& SD1.4 & SD1.5 & SD3 & SDXL & FLUX & DiT & FLUX-Kontext & NanoBanana & GPT-Image &  SeedDream4 &Avg\\
\midrule
\noalign{\vspace{0.5pt}}
SD1.4 & \textbf{96.06} & 88.35 & 82.53 & 87.90 & 70.39 & 66.67 & 63.44 & 64.78 & 61.29 & 68.01 & 74.94 \\
SD1.5 & 84.90 & \textbf{93.37} & 69.09 & 71.51 & 68.59 & 61.83 & 58.87 & 62.90 & 64.78 & 65.59 & 70.14 \\
SD3 & 74.87 & 70.07 & \textbf{95.70} & 71.77 & 70.34 & 64.92 & 61.83 & 59.41 & 59.95 & 69.89 & 69.87 \\
SDXL & 83.60 & 73.21 & 71.77 & \textbf{95.83} & 67.47 & 63.04 & 58.87 & 58.87 & 61.83 & 64.78 & 69.93 \\
FLUX & 78.45 & 74.82 & 70.70 & 78.09 & \textbf{95.61} & 75.81 & 77.15 & 70.16 & 65.05 & 70.16 & 75.60 \\
DiT & 74.46 & 63.17 & 68.28 & 79.30 & 78.00 & \textbf{94.35} & 72.85 & 71.24 & 73.92 & 74.73 & 75.03 \\
\noalign{\vspace{-0.5pt}}
\bottomrule
\end{tabular}}

\resizebox{1\textwidth}{!}{\begin{tabular}{l||cccccc||cccc||c}
\toprule
\noalign{\vspace{1pt}}
MVSSNet \cite{mvss} & \multicolumn{10}{c}{Testing Subset} \\
\cmidrule(lr){2-11}
\noalign{\vspace{1.5pt}}
Training Subset& SD1.4 & SD1.5 & SD3 & SDXL & FLUX & DiT & FLUX-Kontext & NanoBanana & GPT-Image &  SeedDream4 &Avg\\
\midrule
\noalign{\vspace{0.5pt}}
SD1.4 & \textbf{90.68} & 83.69 & 61.56 & 74.87 & 58.87 & 53.90 & 59.41 & 58.33 & 51.88 & 61.83 & 65.50 \\
SD1.5 & 80.60 & \textbf{92.83} & 59.41 & 65.19 & 70.52 & 64.78 & 56.72 & 51.88 & 56.99 & 61.29 & 66.02 \\
SD3 & 77.02 & 62.63 & \textbf{86.83} & 68.95 & 57.17 & 57.26 & 62.10 & 58.33 & 51.08 & 54.84 & 63.62 \\
SDXL & 72.09 & 70.16 & 62.10 & \textbf{86.29} & 54.75 & 59.01 & 56.99 & 50.27 & 59.14 & 54.30 & 62.51 \\
FLUX & 67.92 & 72.94 & 62.37 & 73.52 & \textbf{85.08} & 78.76 & 61.56 & 66.13 & 55.91 & 68.01 & 69.22 \\
DiT & 68.59 & 63.80 & 58.87 & 73.39 & 69.89 & \textbf{89.38} & 60.75 & 56.45 & 68.82 & 67.47 & 67.74 \\
\noalign{\vspace{-0.5pt}}
\bottomrule
\end{tabular}}

\resizebox{1\textwidth}{!}{\begin{tabular}{l||cccccc||cccc||c}
\toprule
\noalign{\vspace{1pt}}
HiFiNet \cite{hifi} & \multicolumn{10}{c}{Testing Subset} \\
\cmidrule(lr){2-11}
\noalign{\vspace{1.5pt}}
Training Subset& SD1.4 & SD1.5 & SD3 & SDXL & FLUX & DiT & FLUX-Kontext & NanoBanana & GPT-Image &  SeedDream4 &Avg\\
\midrule
\noalign{\vspace{0.5pt}}
SD1.4 & \textbf{95.21} & 81.72 & 79.84 & 94.89 & 76.08 & 60.22 & 65.59 & 62.10 & 55.91 & 59.14 & 73.07 \\
SD1.5 & 95.61 & \textbf{91.94} & 67.74 & 70.16 & 70.25 & 68.95 & 61.29 & 56.18 & 55.65 & 66.13 & 70.39 \\
SD3 & 83.87 & 72.31 & \textbf{97.85} & 74.33 & 62.99 & 61.16 & 66.94 & 58.87 & 54.57 & 59.95 & 69.28 \\
SDXL & 83.83 & 67.29 & 64.25 & \textbf{90.19} & 71.15 & 61.83 & 60.22 & 61.83 & 56.18 & 67.20 & 68.40 \\
FLUX & 72.58 & 73.30 & 76.34 & 73.39 & \textbf{94.27} & 75.27 & 68.28 & 67.20 & 58.87 & 69.09 & 72.86 \\
DiT & 72.94 & 73.48 & 61.83 & 73.52 & 75.94 & \textbf{91.53} & 68.28 & 60.22 & 67.47 & 67.74 & 71.29 \\
\noalign{\vspace{-0.5pt}}
\bottomrule
\end{tabular}}

\resizebox{1\textwidth}{!}{\begin{tabular}{l||cccccc||cccc||c}
\toprule
\noalign{\vspace{1pt}}
ManipShield (Ours) & \multicolumn{10}{c}{Testing Subset} \\
\cmidrule(lr){2-11}
\noalign{\vspace{1.5pt}}
Training Subset& SD1.4 & SD1.5 & SD3 & SDXL & FLUX & DiT & FLUX-Kontext & NanoBanana & GPT-Image &  SeedDream4 &Avg\\
\midrule
\noalign{\vspace{0.5pt}}
SD1.4 & \textbf{94.31} & 90.59 & 83.33 & 85.89 & 74.10 & 73.79 & 67.74 & 66.67 & 68.55 & 68.01 & 77.30 \\
SD1.5 & 91.85 & \textbf{92.92} & 69.35 & 69.09 & 79.12 & 70.43 & 66.40 & 63.71 & 62.90 & 64.25 & 73.00 \\
SD3 & 80.33 & 79.21 & \textbf{90.86} & 81.59 & 67.38 & 65.59 & 70.43 & 68.28 & 67.47 & 71.51 & 74.27 \\
SDXL & 80.51 & 70.88 & 78.49 & \textbf{92.47} & 65.50 & 64.92 & 66.94 & 57.80 & 63.17 & 62.90 & 70.36 \\
FLUX & 85.39 & 85.75 & 72.85 & 83.47 & \textbf{92.07} & 79.03 & 77.96 & 71.24 & 70.16 & 76.08 & 79.40 \\
DiT & 76.48 & 70.34 & 75.54 & 76.88 & 81.32 & \textbf{95.70} & 76.61 & 73.39 & 77.69 & 78.49 & 78.24 \\
\noalign{\vspace{-0.5pt}}
\bottomrule
\end{tabular}}
\label{diff_backbone_all}
\end{table*}

\begin{table*}[t]
\centering
 \resizebox{1\textwidth}{!}{\begin{tabular}{lcccc}
        \toprule
Method &GPU Memory& MACs(G) & Trainable Params(M) & Infer Time (s) \\

 \midrule
 
                MVSSNet \cite{mvss} &4GB& 106.20 & 85.75 & 0.468  \\
                                AIDE \cite{AIDE} &9GB& 325.25 & 180.24 & 0.726  \\
        FakeShield \cite{fakeshield} &42GB& 5809.82 & 486.43 & 1.530  \\
          \rowcolor{gray!20}  
        ManipShield &23GB& 2980.08 & 122.26 & 0.825   \\
        \bottomrule
    \end{tabular}}
\caption{Cost analysis.}
\label{cost}
\end{table*}

\begin{figure*}[t]
  \includegraphics[width=0.9\textwidth]{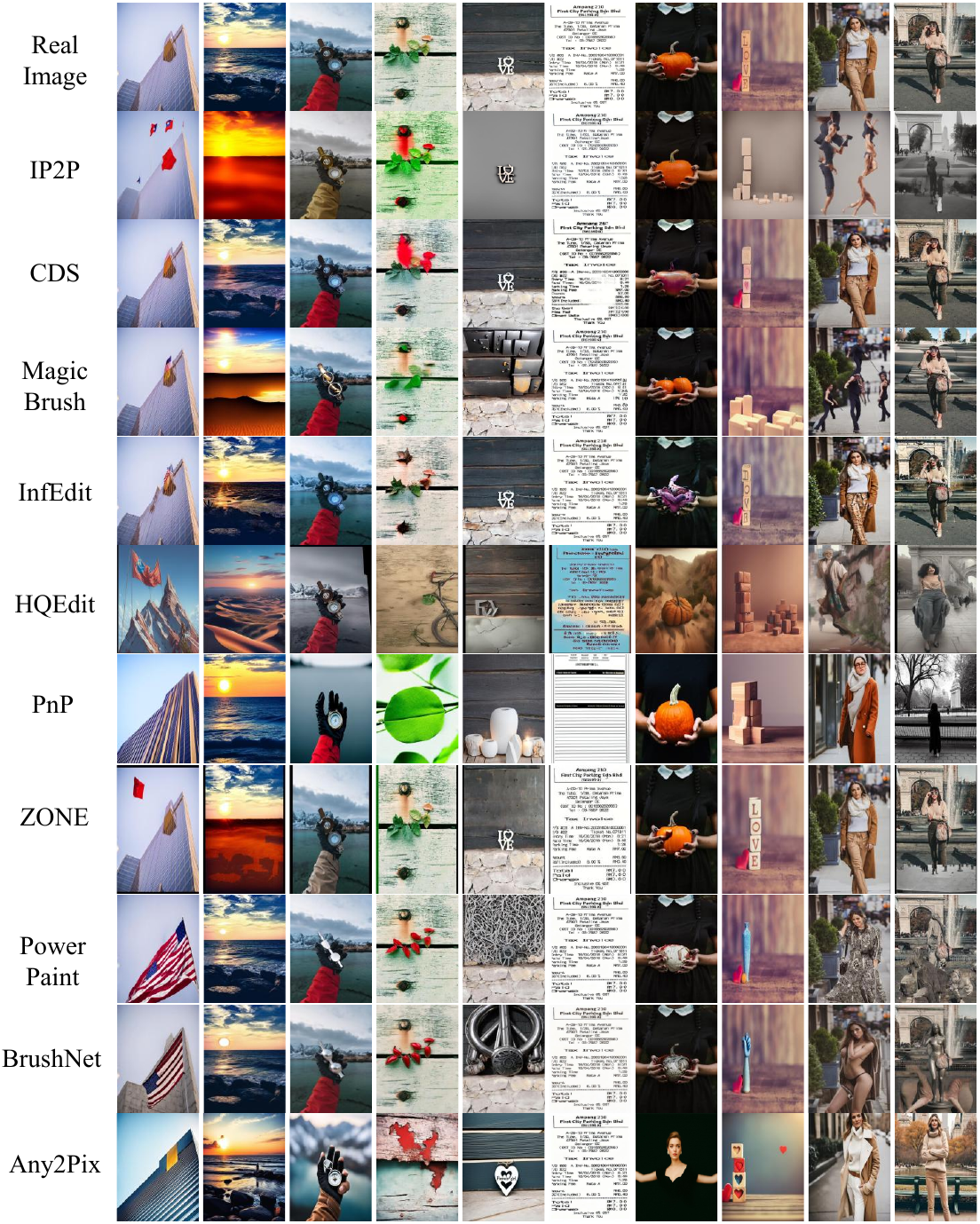}
  \caption{Example images generated by different manipulation models in ManipBench.}
  \label{eg1}
\end{figure*}
\begin{figure*}[t]
    \includegraphics[width=0.9\textwidth]{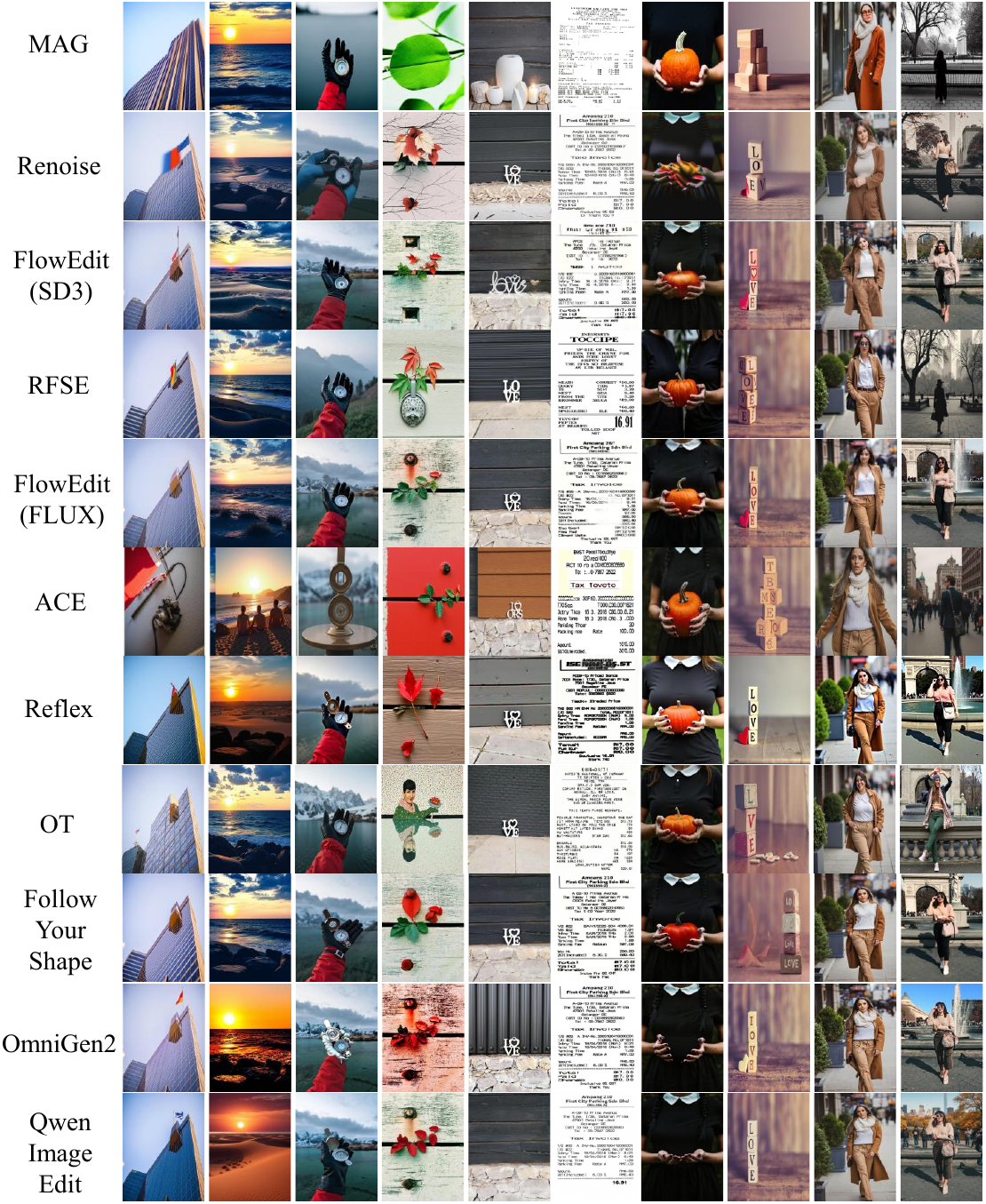}
  \caption{Example images generated by different manipulation models in ManipBench.}
  \label{eg2}
\end{figure*}

\begin{figure*}[t]
    \includegraphics[width=0.9\textwidth]{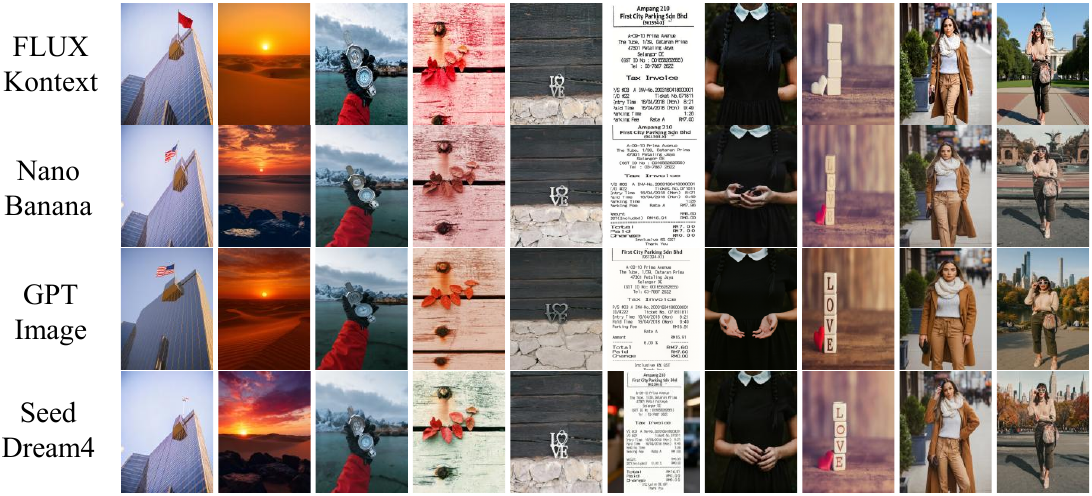}
  \caption{Example images generated by different manipulation models in ManipBench.}
  \label{eg3}
\end{figure*}


\end{document}